\documentclass[journal]{IEEEtran}
\usepackage{bm}
\usepackage{amsmath} 
\usepackage{amsthm}
\usepackage{graphicx}
\usepackage{pdfcomment}

\usepackage{graphicx}
\usepackage{algorithm}
\usepackage[noend]{algpseudocode}

\usepackage{breqn}
\usepackage{multirow}
\usepackage{psfrag}
\usepackage{url}
\usepackage{xcolor}
\usepackage{pdfcomment}
\usepackage{textcomp}
\usepackage{amsfonts}
\usepackage{amssymb}
\usepackage{caption}
\usepackage{subcaption}
\usepackage{paralist}

\newtheorem{theorem}{Theorem}
\newtheorem{lemma}{Lemma}

\newtheorem{assumption}{Assumption}

\usepackage{mathrsfs}
\usepackage{braket}
\usepackage{lscape}
\usepackage{threeparttable}
\usepackage{svg}
\usepackage{multicol}
\usepackage{multirow}
\usepackage{tikz} 
\usepackage{amsmath} 
\usepackage{tabularx}

\begin{document}
\title{
Communication Efficient Adaptive Model-Driven Quantum Federated Learning
}
% \author{
% Dev Gurung and Shiva Raj Pokhrel
% \thanks{
% Authors are associated with School of IT, Deakin University, Australia. 
% Emails: d.gurung@deakin.edu.au, shiva.pokhrel@deakin.edu.au
% }
% }

\author{
Dev Gurung and Shiva Raj Pokhrel
\thanks{Authors are with the School of IT, Deakin University, Australia \\
(e-mails: d.gurung@deakin.edu.au, shiva.pokhrel@deakin.edu.au)}
}
\maketitle
\begin{abstract}
Training with huge datasets and a large number of participating devices leads to bottlenecks in federated learning (FL). 
Furthermore, the challenges of heterogeneity between multiple FL clients affect the overall performance of the system. 
In a quantum federated learning (QFL) context, we address these three main challenges: i) training bottlenecks from massive datasets, ii) the involvement of a substantial number of devices, and iii) non-IID data distributions.
We introduce a model-driven quantum federated learning algorithm (mdQFL) to tackle these challenges. Our proposed approach is efficient and adaptable to various factors, including different numbers of devices. 
To the best of our knowledge, it is the first to explore training and update personalization, as well as test generalization within a QFL setting, which can be applied to other FL scenarios. 
We evaluated the efficiency of the proposed mdQFL framework through extensive experiments under diverse non-IID data heterogeneity conditions using various datasets within the Qiskit environment. 
Our results demonstrate a nearly 50\% decrease in total communication costs while maintaining or, in some cases, exceeding the accuracy of the final model and consistently improving local model training compared to the standard QFL baseline.
Moreover, our experimental evaluation thoroughly explores the QFL and mdQFL algorithms, along with several influencing factors.
In addition, we present a theoretical analysis to clarify the complexities of the proposed algorithm.
The experimental code is available at \footnote{https://github.com/s222416822/mdQFL}.
\end{abstract}
\section{Introduction}
Federated Learning (FL) has emerged as a pivotal technique to address the challenges of privacy and security in distributed machine learning \cite{mcmahanCommunicationEfficientLearningDeep2023, dinhFederatedLearningWireless2021}. However, as data complexity and
device numbers increase, FL suffers from issues related to computational bottlenecks and data heterogeneity
\cite{liFederatedLearningChallenges2020}. In particular, non-IID (non-independent and identically distributed) data significantly affect the performance of federated learning systems.

Quantum Federated Learning (QFL) is proposed to harness the power of quantum computing to solve complex machine learning problems
on decentralized quantum devices \cite{knillQuantumComputing2010}. In QFL setup with a large number of devices, challenges such as
non-IID data distribution, communication overhead, and training complexity due to large datasets are still prevalent. 

\begin{figure}
    \centering
      \includegraphics[width=0.7\columnwidth]{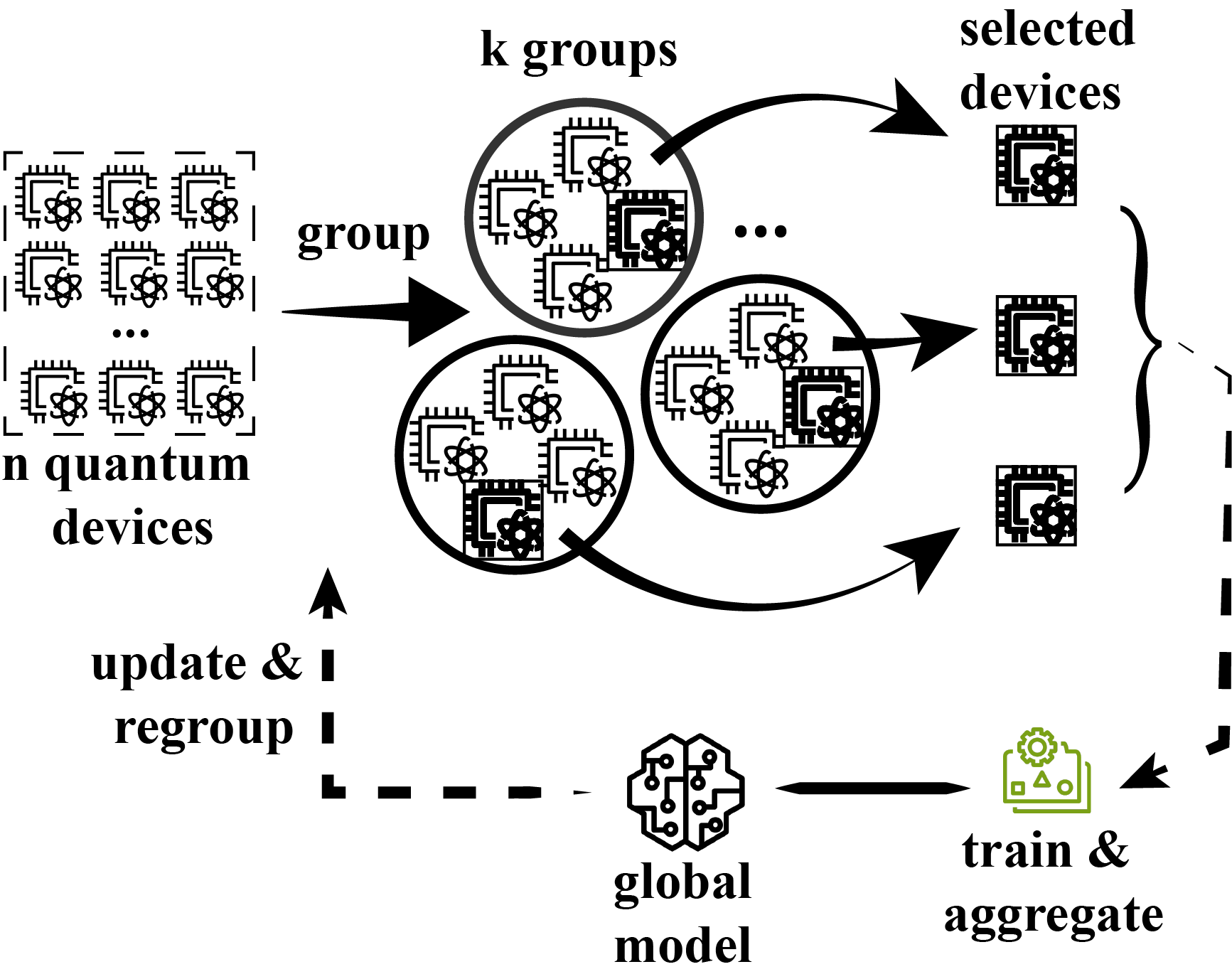}
    \caption{Preview of the proposed mdQFL framework: \textit{Based on training progress of $n$ quantum devices, they are grouped in 
    various clusters; Next, a single device is selected from each group to participate in training which generates representative cluster model it belongs to; Then, updating model to all devices and server follows varying degree of personalization 
    and generalization.
    }}
    \label{fig:mdQFL-preview}
\end{figure}

In this paper, we propose a novel Model-Driven Quantum Federated Learning (mdQFL) framework, which addresses these issues through a
combination of grouping mechanism, optimized device selection, and efficient adaptive model designs. As shown in Fig.~\ref{fig:mdQFL-preview}, we explicitly focus on reducing QFL training
bottlenecks in distributed quantum environments while mandating high scalability and efficiency.

\subsection{Related Work}
Various works have been done in the context of classical FL (CFL), to address heterogeneity. Li et al. \cite{fedProx_2020} introduced FedProx, an extension of FedAvg, which allows for variable workloads among devices and incorporates a proximal term to mitigate the effects of system heterogeneity. Li et al. \cite{fedBN_ICLR2021} proposed applying batch normalization prior to FedAvg, suggesting that this method outperforms both FedAvg and FedProx on non-IID data. Gao et al. \cite{FedDCFederatedLearning2022} suggested FedDC to tackle statistical heterogeneity by introducing adjustments during the local training phase and employing an auxiliary local drift variable to monitor discrepancies between local and global models.
Dai et al. \cite{daiDisPFLCommunicationEfficientPersonalized2022} introduced a personalized FL approach to address the issue of heterogeneous data across clients.
The concept aims to develop specific local models tailored for each individual user.
However, all of these studies are set in classical contexts.
Given the substantial differences in the operation of classical and quantum computers, it remains uncertain whether the same CFL algorithmic strategies can be applied to QFL.
In QFL, the challenges will differ regarding communication, data management, aggregation methods, and other factors.

Among the very few works done to solve the non-IID issues in QFL is by Zhao et al. \cite{zhaoExactDecompositionQuantum2022a}.
They proposed a general QFL framework (qFedInf) for non-IID data with one-shot communication complexity. 
The main idea presented is the decomposition of a global quantum channel into channels trained by each client with the help of local density estimators.
However, the proposed framework includes two steps or phases. 
The first is the training phase, and the second is the inference phase.
One-shot complexity is of great benefit; however, with the involvement of two phases, it is worthwhile to explore if there exists a personalized approach to achieve the same or even better performance.

Other works proposed for better QFL frameworks are the work of Xia et al. \cite{xiaQuantumFedFederatedLearning2021a} who introduced a federated learning framework for quantum training with quantum nodes, while Chehimi et al. \cite{chehimiQuantumFederatedLearning2022} and Huang et al. \cite{huangQuantumFederatedLearning2022} proposed frameworks and algorithms to improve communication efficiency and privacy. Other notable works include
Chen et al. \cite{chenFederatedQuantumMachine2021} presenting a hybrid quantum-classical approach, and Zhang et al. \cite{zhangFederatedLearningQuantum2022} proposing a secure aggregation scheme. 

% \subsection{Client Selection}
In terms of client selection for data heterogeneity, 
Li et al. \cite{liDataHeterogeneityRobustFederated2022} proposed a hierarchical FL framework with edge-end cloud that includes the selection of a subset of devices within natively clustered factory devices for training.
The core idea implemented is to strategically select a small number of devices from each group to form super nodes but with homogeneous data distributions based.
In other work by Ghosh et al. \cite{ghoshEfficientFrameworkClustered2020a}, an efficient framework for federated cluster learning was proposed that alternatively estimates the cluster identities of users and gradient descent is used to optimize the model parameters for the clusters of users.
Our research diverges from previous studies by exploring Federated Learning (FL) within a quantum framework. In addition, we have developed an innovative algorithmic method specifically tailored to QFL.

\subsection{Our Ideas and Contributions}
The ideas and contributions in this work are as follows:
\begin{enumerate}[a)]
    \item \textbf{Device Grouping Mechanism}: 
    We introduce a novel device grouping mechanism that integrates with popular clustering algorithms to organize quantum clients based on their learning progress. 
    Next, adhering to the standard distance methods used in clustering algorithms, 
    the dissimilarity measure \( D(i, j) \) between clients \( i \) and \( j \) can be computed. 
    This measure can be based on differences in their model parameters \( \theta_i \) 
    or performance metrics such as their local loss functions \( f_i(\theta_i) \).
    The clients are then partitioned into \( k \) clusters \( \{ C_1, C_2, \dots, C_K \} \) by solving the following optimization problem
    \[
    \min_{\{ C_k \}} \sum_{k=1}^K \sum_{i, j \in C_k} D(i, j),
    \]
    where, \( D(i, j) = \| \theta_i - \theta_j \|^2 \) is the squared Euclidean distance, Cosine distance or any other method between the model parameters of the clients \( i \) and \( j \). This mechanism minimizes redundant updates by grouping similar clients, thus optimizing communication overhead and computational resources.

    \item \textbf{Device Selection Method}: Using the proposed grouping mechanism, we introduce a device selection method designed specifically for local training purposes. Within each group \( C_k \), we select a representative device \( d_k \) 
    based on a selection criterion \( s(i) \). The selection can be random or follow a greedy data-driven approach aimed at optimizing the performance of the model. Formally, the selected device in cluster \( k \) is determined by:

    \[
    d_k = \arg\min_{i \in C_k} s(i),
    \]

    where, \( s(i) \) could be the latest local loss \( f_i(\theta_i) \), the norm of the gradient \( \| \nabla f_i(\theta_i) \| \), or another performance metric indicative of client \( i \)'s learning progress. 
    This strategy reduces computational redundancy and focuses on training on devices that are most likely to enhance the global model.

   \item \textbf{Degrees of Personalization and Generalization}: We establish mathematically grounded degrees of personalization and generalization by introducing adaptive formulations that allow devices and the server to update, train, and test their models using different combinations of models, such as their own model, the global model, or the mean of other models.
    Specifically, we define the update mechanism for the client or server as
    
    \[
    \theta_i^{(t+1)} = \Phi(\alpha_i \theta_i^{(t)} + \beta_i \theta_{\text{cluster},i}^{(t)} + \gamma_i \theta_{\text{global}}^{(t)}),
    \]
    
    where \( \theta_i^{(t)} \) is the local model parameters of client \( i \) at iteration \( t \); \( 
    \theta_{\text{cluster},i}^{(t)}\) is the mean of the models 
    from clients in the same cluster \( C_i \) as client \( i \) at iteration \( t \) (in our implementation, this is the model learned from the selected device); \( \theta_{\text{global}}^{(t)} = \dfrac{1}{n} \sum_{j=1}^{n} \theta_j^{(t)} \) is the global model parameters in iteration \( t \); and \( \alpha_i, \beta_i, \gamma_i \geq 0 \) are weighting parameters for which model to be used.
    Adjusting \( \alpha_i, \beta_i, \gamma_i \), each client can choose to incorporate its own local model, the cluster model, and the global model in their updates. This flexible framework allows clients and the server to update, train and test their models using different combinations, facilitating effective aggregation and improving performance in various phases.
    
    \item \textbf{Extensive Theoretical and Empirical Analysis}: We present a theoretical analysis to demonstrate the convergence properties and performance guarantees of our proposed methods under non-IID data distributions and quantum noise. We derive convergence bounds showing that our algorithm achieves a convergence rate of \( \mathcal{O}\left( 
    \frac{1}{T} \right) \) for convex loss functions, where \( T \) is the number of communication rounds. Additionally, we perform empirical evaluations on datasets such as MNIST, Genomic etc. using Qiskit framework to validate our theoretical findings, showing 
    significant improvements in communication efficiency and model accuracy compared to baseline methods. 

\end{enumerate}

\subsection{Problem formulation}
Consider a QFL setup with \( n \) quantum clients, each possessing its local quantum dataset. These clients collaboratively aim to develop a global quantum model \(\theta\) by solving the following optimization problem as
\cite{mcmahanCommunicationEfficientLearningDeep2023},

\begin{equation}
\label{eqn:standard_fedavg}
QFL:\min_{\theta \in R^d} F(\theta), \text{where}  
F(\theta) := 
\frac{1}{n} 
\sum_{i=1}^{n} f_i(\theta)  \notag
\end{equation}

Here, \( f_i \) denotes the local loss of the prediction for the client \( i \) in the data sample $(x_i, y_i)$ with the model parameters $\theta$, which is calculated as 

\begin{equation}
    f_i(\theta) = \mathbb{E}_{\xi_i} [f_i(\theta_i, |\psi(x)\rangle_i, y_i ; \xi_i)], \notag
\end{equation}
where, \(\xi_i\) is a random sample from the  \( i \) client's data distribution, \( |\psi(x)\rangle_i \) represents the quantum state of the input, \( y_i \) is the output and \( f_i(\theta_i, |\psi(x)\rangle_i, y_i ; \xi_i) \) is the local loss function.
In addition,
\ref{eqn:standard_fedavg} can be represented as \cite{liFederatedLearningChallenges2020}, 
\[
\min_\theta F(\theta), \text{ where  }  F(\theta) := \sum_{i=1}^{n} p_i F_i(\theta). \notag
\]
Here, \( n \) is the total number of devices,  $p_i$ is the relative weightage of the client's contribution usually a ration of data sample at client against total number of samples across all clients
and 
\( F_i \) is the local objective function for the \( i \)-th device. 
Now, the local objective function can also be expressed in terms of the empirical risk on local data as
\[
F_i(\theta) = \frac{1}{n_i} \sum_{i=1}^{n_j} f_{i}(\theta; |\psi(x_{j})\rangle, y_{j}), \notag
\]
where, \( n_j \) is the number of samples available locally, \( p_i \) specifies the relative impact of the device, 
and \( n_j = \sum_{i} n_i \) is the total number of samples.

\begin{figure}
    \centering
    \includegraphics[width=\columnwidth]{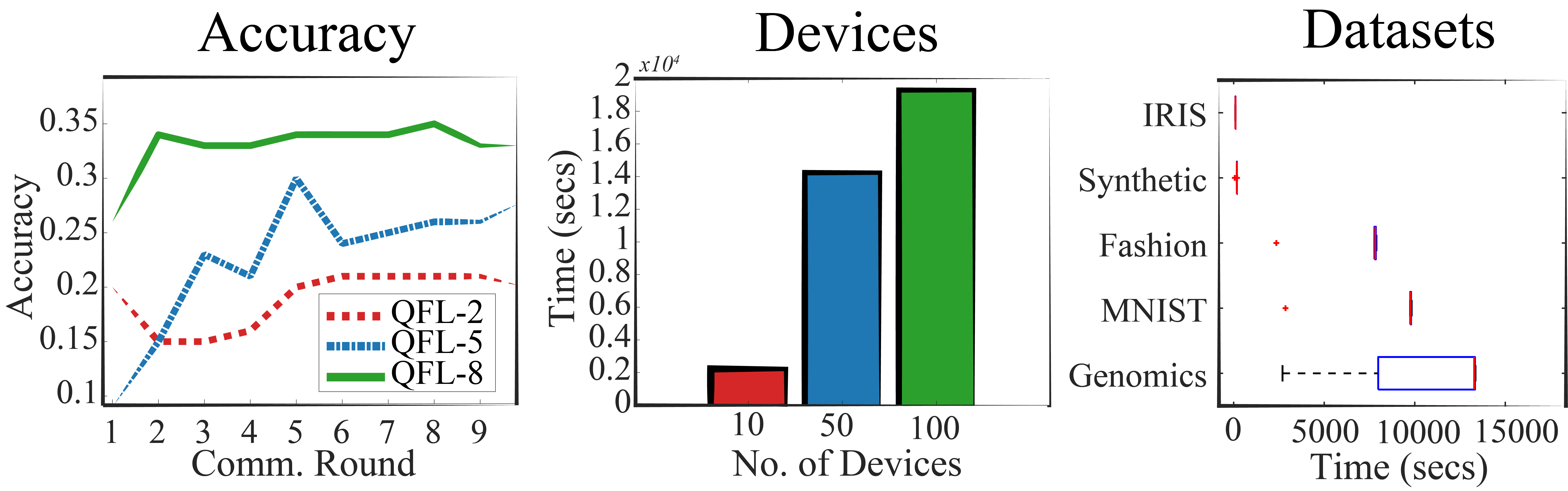}
   \caption{Research Problem: Degrading performance due to increase in degree of non-IID; \textit{QFL-2, QFL-5 and QFL-8 represent the degree of non-IID, with ``-2" representing only 2 unique labels assigned to a device which are not present in other devices or at least just one label overlapping with the next device. The accuracy of the server device which has all labels suffer from the performance due to higher degree of non-IID; Likewise, the influence on communication latency due to the quantity of devices, the type of the dataset, and its size is evident.}}
    \label{fig:impact_of_noniid}
\end{figure}

However, in the presence of heterogeneous data distributions among clients, we present an optimization goal divergent from the general QFL problem. 
For $k$ different clusters, 
we formulate our optimization problem as,
\begin{equation}
\text{mdQFL} : \min_{\theta \in \mathbb{R}} 
\left\{ F(\theta) := \frac{1}{k} 
\sum_{k=1, i \in d_k}^k
f_k^d(\theta_{k}^d) 
\right\}
\end{equation}
where, 
\( f_k^d \) is the loss function for the client \( d_k \) which is selected from each cluster group, \(\theta_k^d\) is the local model, \( k \) is the total number of clusters.

Figure \ref{fig:impact_of_noniid} demonstrates the effect of the non-IID data distribution on QFL clients. It is clear that the global model's accuracy is highly impacted by the degree of non-IID. 
When each client is associated with just two distinct class labels (indicating a higher degree of non-IID), the accuracy deteriorates and hits its lowest point.
In contrast, as the number of class labels per client increases to 5-8, the accuracy improves accordingly.

\subsection{Understanding the Research Challenges}
This work seeks reliable methods to address the following research challenges.

\begin{enumerate}[a)]
    \item \textit{Is it necessary to include all the devices for training, especially when groups of devices train similarly and produce similar machine learning models?}

    In traditional FL, the global model is updated by aggregating the updates from all participating devices or clients.
    However, this approach may not be efficient or necessary when there are groups of devices whose local models converge to similar parameters due to similar data distributions or learning behaviors. Thus, it would be beneficial to identify and group devices that produce similar models to reduce redundant computations and communications.
    \item \textit{Are non-IID, large datasets, and huge number of devices factors closely related? How do we address them together?}
    
    The heterogeneity of the data between devices can introduce significant challenges in aggregating local models, especially when a large number of devices amplifies this impact. Thus, it would be crucial to understand how large datasets increase the variance of local updates, that affect convergence of the global model. Understanding how these factors affect the convergence rate and mitigating those challenges is crucial.

    \item \textit{What is the optimal solution to train large datasets with huge number of devices without causing any bottlenecks in the performance of the system?}

    Minimizing the variance caused by the varying number of devices participating is the key to improving convergence rates. This may involve adjusting parameters such as the size of the datasets and the number of participating devices or clusters. It is crucial to find answers to how we can improve the convergence bound and enhance training efficiency under such conditions.

\end{enumerate}

\subsection{New Insights from FL}
The quantum version of Federated averaging~\cite{zhaoExactDecompositionQuantum2022a} can be modelled as,
\begin{equation}
    \underset{\theta}{\min \in R^d} \quad \frac{1}{n} \sum_{(|\psi(x)\rangle, y) \in D} f(\theta, |\psi(x)\rangle, y) \notag
\end{equation}
where, $n$ is the total number of samples, $\psi$ is the input quantum state that is fed to the quantum channel or circuit, $\theta$ is the variational parameters, $y$ is the corresponding label class and $f$ is the loss function on training dataset $D$.
The main goal is to find and tune the quantum circuit that can take the quantum input and produce the desired output. 
This is done by tuning $\theta$ so that the loss function can be minimized on the training data set.
The optimization problem is solved using various optimizers, whereas for SGD with learning rate $\eta$, we have
\begin{equation}
    \theta_t = \theta_{t-1} - \eta \nabla_\theta \frac{1}{n} \sum_{(|\psi(x)\rangle, y) \in D} f(\theta, |\psi(x)\rangle, y) \notag
\end{equation}

Now, after certain communication rounds, the quantum fed-averaging is performed as, 
\begin{equation}
    \theta_g = \frac{1}{n} \sum_{i=1}^n \theta_i \notag
\end{equation}

In a distributed setting, heterogeneity can be divided into data, communication, and computational heterogeneity.
Variations can arise from differences in these objective functions between clients, 
resulting in divergence in models and general convergence of global models. 
With a significant number of local updates, 
the global model tends to converge towards an average of local minima, 
which might deviate from the true optimum. 
However, with communication heterogeneity,
in a typical FL framework, 
there could be up to \(n\) clients.
These clients
may not always be available for various reasons, such
as power constraints or being in different time zones. 
It is often impractical to expect consistent availability from all clients in a real-world scenario. 
Finally, with computational heterogeneity, 
clients exhibit different computational speeds
and might have various memory constraints.
Not all clients have the same number of local updates, 
which introduces further heterogeneity. 
This computational disparity can increase existing challenges with data and communication heterogeneity. 
Furthermore, clients might employ different hyper-parameters
and different optimization methods,
leading to variations in the computational results. 
Therefore, comprehending the consequences of these computational variations is essential.

\section{Details of the proposed mdQFL}
In this section, we elaborate on our proposed mdQFL framework. The specifics of this framework are outlined in the proposed Algorithms \ref{alg:mdQFL-algorithm} and further details for the setup and processes of the pre-Algorithms are presented in the Appendix \ref{sec:appendix}. Figure \ref{fig:overall_mdQFL} illustrates an overview of the proposed mdQFL framework. 

\begin{figure*}
    \centering
    \includegraphics[width=0.9\textwidth]{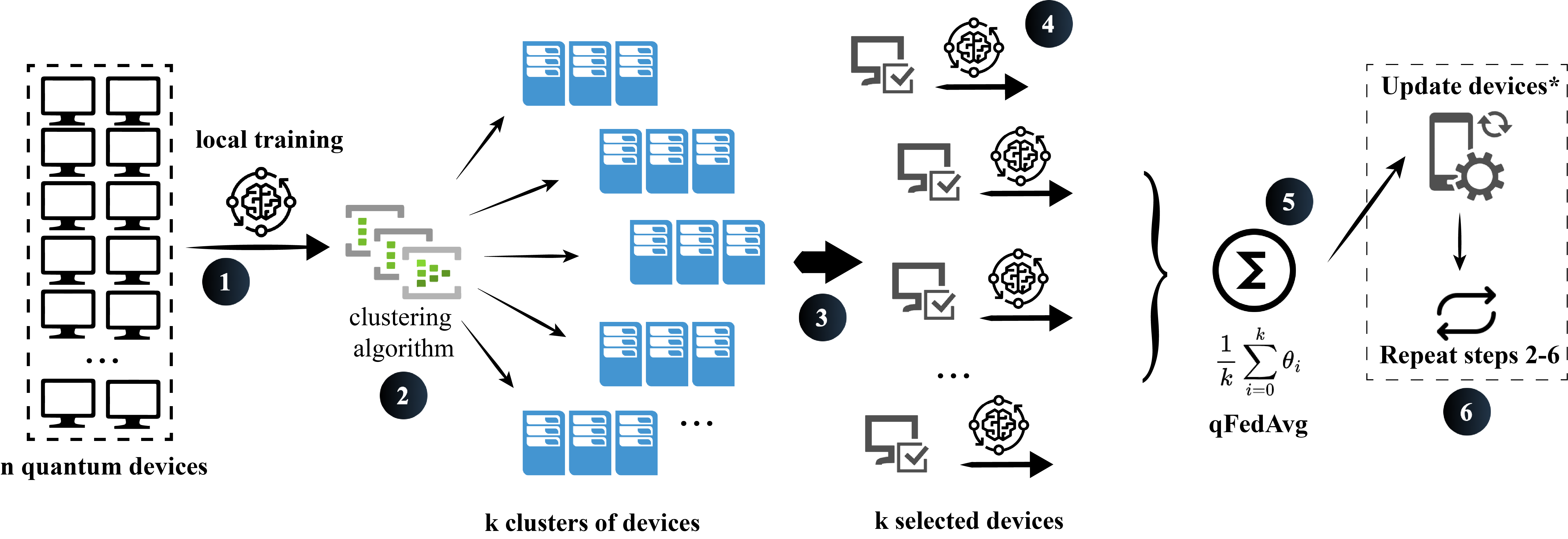}
    \caption{Proposed model driven QFL (mdQFL) framework: 
    \textcircled{1} First all devices initially train with random model parameters; \textcircled{2} Based on the learnt models, devices are grouped into $k$ clusters; \textcircled{3} From each group a device is selected; \textcircled{4} Selected devices perform local training followed by qFedAvg at \textcircled{5}; After that update all local devices models and repeat all steps from \textcircled{2} to \textcircled{6}.
    }
    \label{fig:overall_mdQFL}
\end{figure*}

\begin{algorithm}[!htbh]
\caption{Model-Driven QFL}\label{alg:mdQFL-algorithm}
\begin{algorithmic}[1]
\State \textbf{Input:} Number of devices $n$, List of devices $\{d_1, d_2, \ldots, d_n\}$, Communication rounds $R$, 
Total weights $\theta_{total}$, Average weights $\theta_{avg}$,  Average Cluster Model $\theta_c$,  Server device $S$, Number of clusters $k$, Clustering method $method$.
\State \textbf{Output:} Selected devices $\mathcal{D}_s$, Global model $\theta_{g}$, Cluster Models List  $list_c$
\For{$r = 0$ to $R-1$}
    \If{$r = 0$}
        \For{each device $d_i$ in $\{d_1, d_2, \ldots, d_n\}$}
            \State $\theta_{d_i} \gets \text{Initialize}$
            \State $d_i.\text{train}()$
        \EndFor
    \EndIf
   \State Following Eq. 4.0,
    \State Clusters  $\mathcal{C} \gets \text{cluster}(\{d_1, d_2, \ldots, d_n\}, k, method$)
    \For{each cluster $C_i \in \mathcal{C}$}
        \If{$C_i \neq \emptyset$}
        \State \textbf{Device Selection:}
            % \State Select a device $d_i^*$ from $C_i$ using
           \If{Use method from eqn 5.0}
            \State  $d^* = \arg\min_{d_i \in C} f(d_i)$
        \ElsIf{Use method from eqn 5.1}
         \State $P(d^*) = \frac{1}{|C|}$
        \EndIf
            \State \textbf{Train Personalization:} 
\If{Use model from eqn 1.0}
    \State $\theta_i \gets \theta_g$
\ElsIf{Use model from eqn 1.1}
 \State $\theta_i \gets \mu(\theta_g, \theta_s^{i-1})$
\EndIf
\State $\text{Train}(d_s)$
\State $list_c \gets list_c \cup \{\theta_s\}$
\State \textbf{Update Personalization:}
\For{each device in cluster}
    \If{Update using Eq. 2.0}
        \State $\theta_i \gets \theta_s$
    \ElsIf{Update using Eq. 2.1}
        \State $\theta_i \gets \mu(\theta_s, \theta_{i-1})$
    \ElsIf{Update using Eq. 2.2}
        \State $\theta_i \gets \mu(\theta_s, \theta_d, \theta_g)$
    \EndIf
\EndFor
        \EndIf
    \EndFor
 \State $\theta_c \gets \frac{1}{|list_c|} \sum_{\theta \in list_c} \theta$
\EndFor
\State $\theta_g \gets \frac{1}{n} \sum_{i=1}^{n} \theta_i$
\State \textbf{Test Generalization:}
\If{Test model from Eq. 3.0}
    \State $\theta_S \gets \theta_g$
\ElsIf{Test model from Eq. 3.1}
    \State $\theta_S \gets \mu(\theta_g, \theta_c)$
\Else
    \State $\theta_S \gets \theta_c$
\EndIf

\State Perform $S.\text{validate}()$
\State Perform $S.\text{test}()$
\end{algorithmic}
\end{algorithm}

The proposed mdQFL framework
introduces a unique approach that differs from conventional FL methodologies. 
Within each communication round, 
three main steps are involved. 
Initially, individual devices undergo local pre-training
to generate local models. 
Subsequently, 
these devices are grouped on the basis of the distance between their model parameters. 
After this, a global aggregation process is initiated. 
Also, a group of cluster models is derived which is followed by the devices updating their models with variations of options
corresponding to degree of personalization.

The Algorithm \ref{alg:mdQFL-algorithm} shows the steps involved in the algorithm.
The algorithm begins with initializing the weights and training the models on each device. The devices are then clustered according to a specified clustering method (such as k-means, agglomerative clustering, etc.) to group the devices into $k$ clusters. From each group, a representative device is selected to form the subset of devices ($D_s$) that will participate in the training process for that round.

In the initial communication round, each device's model is initialized and trained independently. 
Subsequently, the algorithm clusters the devices and selects representative devices from each cluster. 
For each communication round beyond the first, the selected devices use chosen model weights and undergo local training. 
The server then updates its global model with choice of average weights
and performs testing to evaluate the model's performance. 
This cyclical procedure is repeated for a predetermined number of communication rounds, enhancing model efficiency and performance robustness via deliberate device selection and grouping.
During each phase of clustering, device selection, training, update, and testing, different choices can be made based on the required level of personalization or generalization. 
This is further elaborated in Section \ref{sec:system_model}.

\section{Underlying System Model by Design} \label{sec:system_model}
\begin{figure}[!h]
    \centering
    \includegraphics[width=0.9\linewidth]{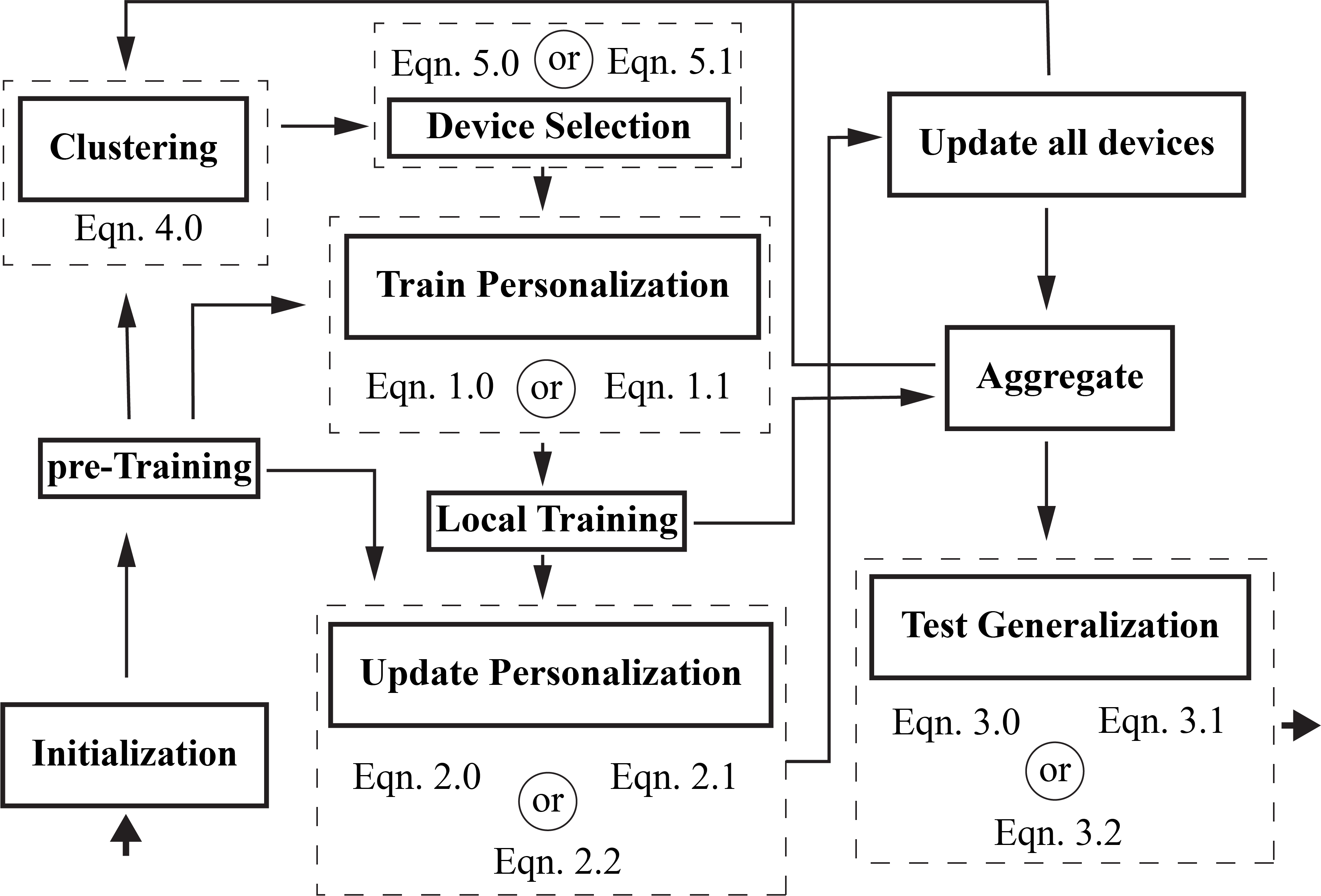}
    \caption{System Model and Design: Schematic view of system model of proposed mdQFL framework.}
    \label{fig:mdQFL_flow_chart}
\end{figure}

In this section, we elaborate on the detailed design of the proposed mdQFL framework, an innovative adaptive QFL framework that improves communication efficiency while improving local training and maintaining stable server performance. The interaction between different components is shown in Fig.~\ref{fig:mdQFL_flow_chart} and their design behind the scene is discussed below.

\subsection{Adaptive Model}\label{sec:adaptive}
Following adaptive design considerations drive our proposed framework for utmost adaptability.

\begin{enumerate}[a)]
    \item \textbf{Adaptive degree of personalization:} The adaptive degree of personalization is used to maintain optimum performance of local devices. 
  Therefore, our proposed method allows for flexible adaptation in updating a device model (\textbf{update personalization}). This can be done using a variety of available models, such as its previous local model, the new cluster model, the global model that covers all devices, or a combination of these models.
    Along with it, we have personalization to how a device model is updated for training (\textbf{training personalization}) i.e. consideration of what model to be used by selected device to perform training.
     Consider $n$ devices, for device $i$, its local model $\theta_i$.
     Based on $\theta_i$, the device is clustered into a group. If it is randomly selected to perform the training, then we have two options to use the model.
     If $\theta_{train}$ is the model to be used then we update the device as, 
    \begin{align}
    \theta_{train} &= \theta_g = \frac{1}{n_d} \sum_{j=1}^{n_d} \theta_j \tag{1.0} \\
    \theta_{train} &= \Phi(\theta_c + \theta_g) = \Phi(\frac{1}{n_c} \sum_{i=1}^{n_c} \theta_i + \theta_g) \tag{1.1}
    \end{align}

    where, $\theta_c$ is the average model of all clusters, $\theta_g$ is the average model of the devices.

    For update personalization we have, 
    \begin{align}
    \theta_{update} &= \theta_c = \frac{1}{n_c} \sum_{i=1}^{n_c} \theta_i \tag{2.0} \\
    \theta_{update} &= \Phi(\theta_c + \theta_d) = \Phi(\frac{1}{n_c} \sum_{i=1}^{n_c} \theta_i + \theta_d) \tag{2.1} \\
    \theta_{update} &= \Phi(\theta_c + \theta_d + \theta_g) = \Phi(\frac{1}{n_c} \sum_{i=1}^{n_c} \theta_i \notag\\
    & +   \theta_d + \frac{1}{n_d} \sum_{j=1}^{n_d} \theta_j) \tag{2.2}
    \end{align}

    where, $\theta_d$ is its old model, $n_c$ number of clusters, and $n_d$ number of devices.
    Based on the need for update personalization or training personalization, we update local models or use training models accordingly. This is represented in Figure \ref{fig:adaptive_device_model}.
    
    \begin{figure}[t]
    \centering
    \includegraphics[width=0.7\columnwidth]{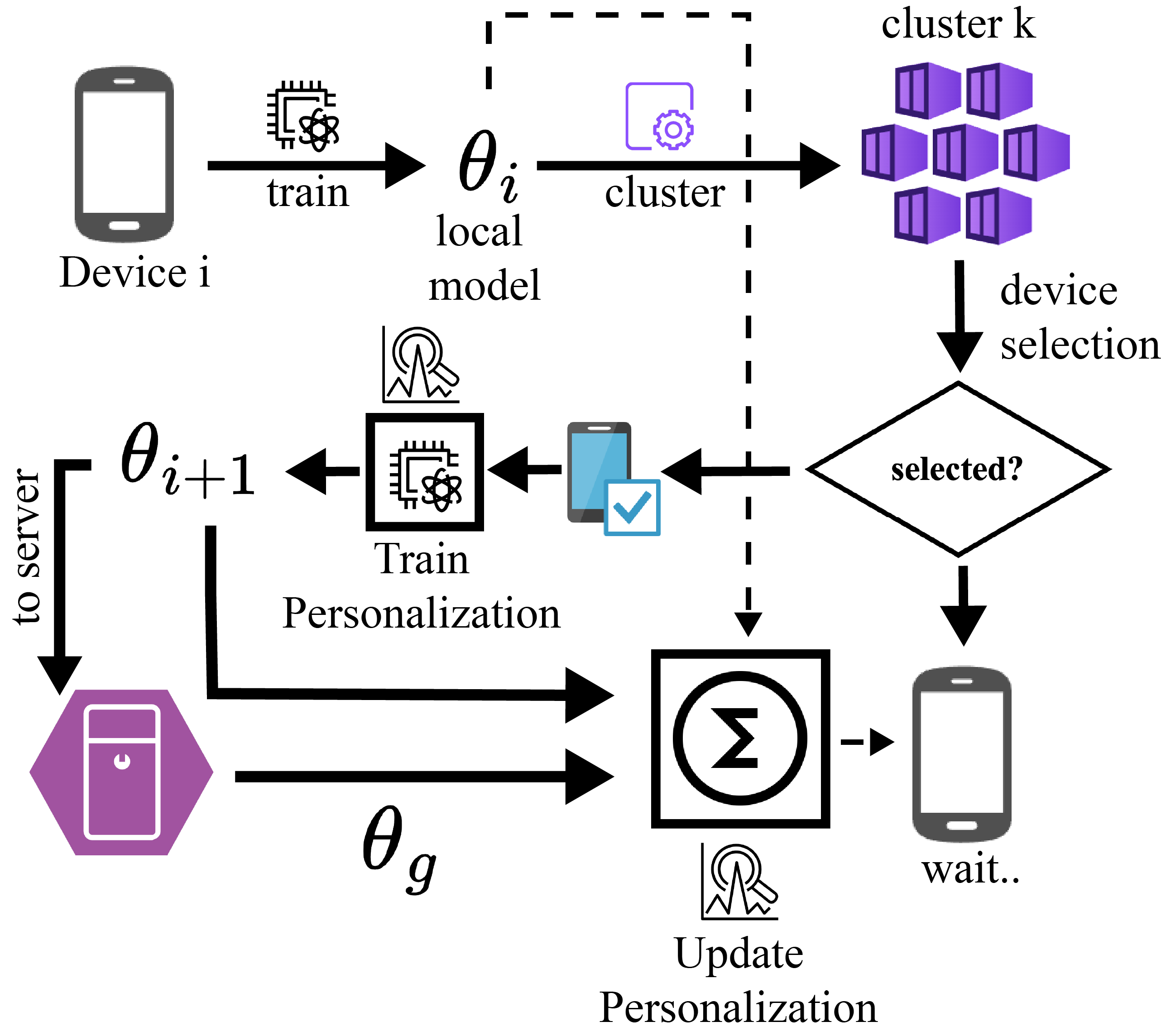}
    \caption{Device Adaptive Model Update: Device $i$ first trains with local dataset. Based on its trained model it will be clustered to its belonging cluster $k$; From cluster $k$ a single device gets selected; The selected device will further train to generate $\theta_{i+1}$ and send it to the server; After global aggregation, the local device can be updated with either cluster model, all device average model or combination or average of those models along with consideration of $\theta_i$ based on need for \textbf{degree of personalization}.}
    \label{fig:adaptive_device_model}
    \end{figure}
    
    \item \textbf{Adaptive degree of generalization:}
    Similarly, for generalization, various options can be used according to the degree of generalization needed for the test model.
    This is depicted in Figure \ref{fig:adaptive_server_model}.

    \begin{align}
    \theta_s &= \theta_g = \frac{1}{n_d} \sum_{j=1}^{n_d} \theta_j \tag{3.0} \\
    \theta_s &= \Phi(\theta_g + \theta_c) = \Phi(\frac{1}{n_d} \sum_{j=1}^{n_d} \theta_j + \frac{1}{n_c} \sum_{i=1}^{n_c} \theta_i) \tag{3.1} \\
    \theta_s &= \theta_c = \frac{1}{n_c} \sum_{i=1}^{n_c} \theta_i \tag{3.2}
    \end{align}

    where, $\theta_s$ is the server model, $\theta_g$ represents the average of all the device models, and $\theta_c$ represents the average of all cluster models.

    \begin{figure}[t]
    \centering
    \includegraphics[width=0.7\columnwidth]{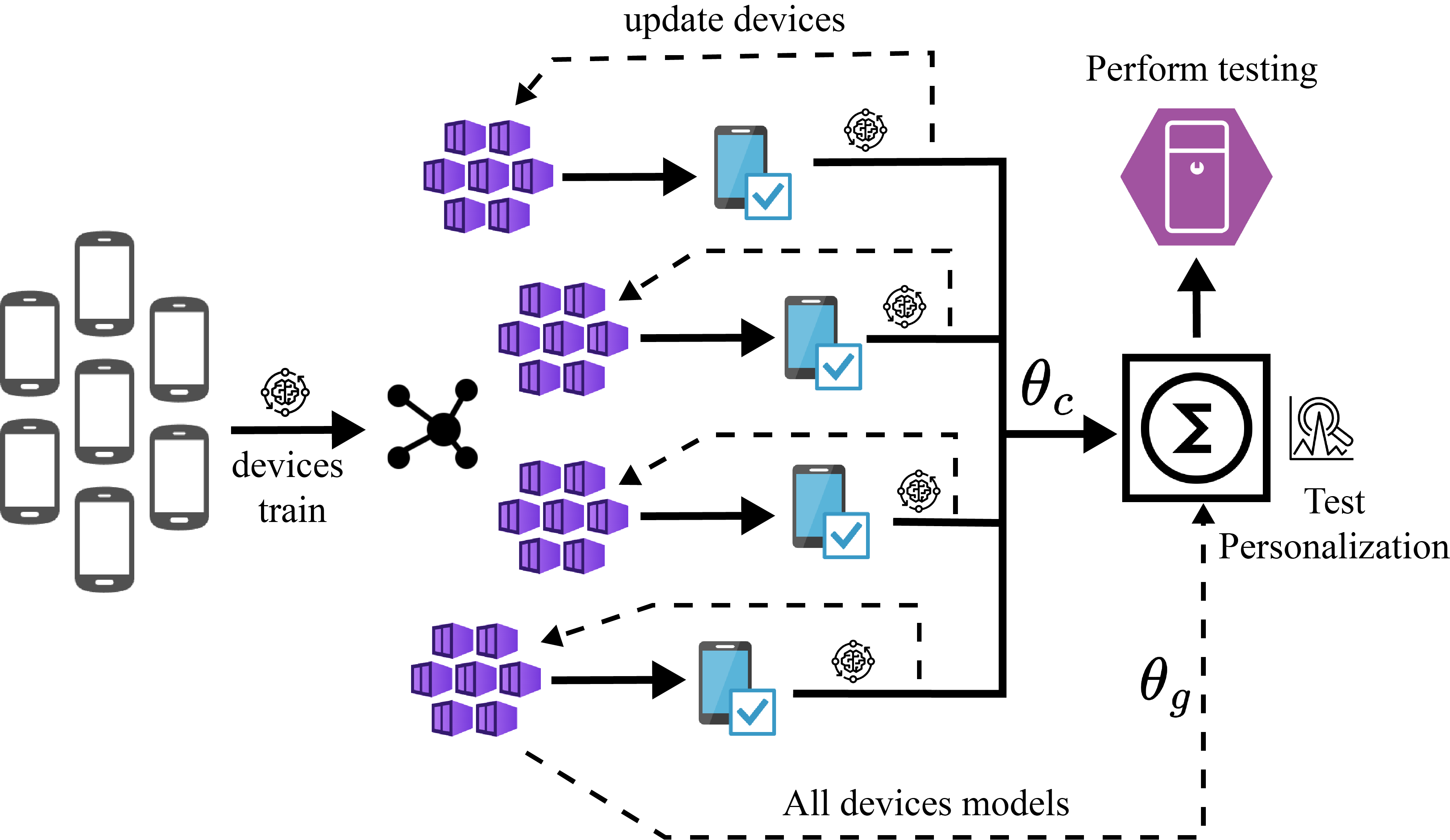}
    \caption{Server Adaptive Model Update: After all devices train, they get grouped based into $k$ clusters; From each cluster one device is selected which trains for the cluster;  All devices in that cluster is updated by the model of selected device; After update, average of all devices as well as all selected devices is performed; Based on the need for \textbf{degree of generalization} either cluster model or all device model or some weighted average of both those models can be used in adaptive fashion depending upon how the performance is achieved.}
    \label{fig:adaptive_server_model}
    \end{figure}
    \item \textbf{Adaptive number of clients:}
    Number of devices selected to participate in training or number of clusters can be either fixed or based on the number of devices as depicted in Figure \ref{fig:adaptive_number_of_device_model}. 
    Mathematically, we have, 
    \begin{align}
         k = \max\left(1, \left\lceil \sqrt{\frac{n}{2}} \right\rceil \right)
         \tag{4.0}
    \end{align}
   
    Here, $k$ is the number of clusters and $n$ is the total number of devices. The function ensures that there is at least one cluster and calculates the number of clusters by rounding up the square root of half the number of devices.
    This is an effective approach; however, the number of clusters can be fixed from the beginning as required as well.
    
     \begin{figure}[!htbh]
    \centering
    \includegraphics[width=0.7\columnwidth]{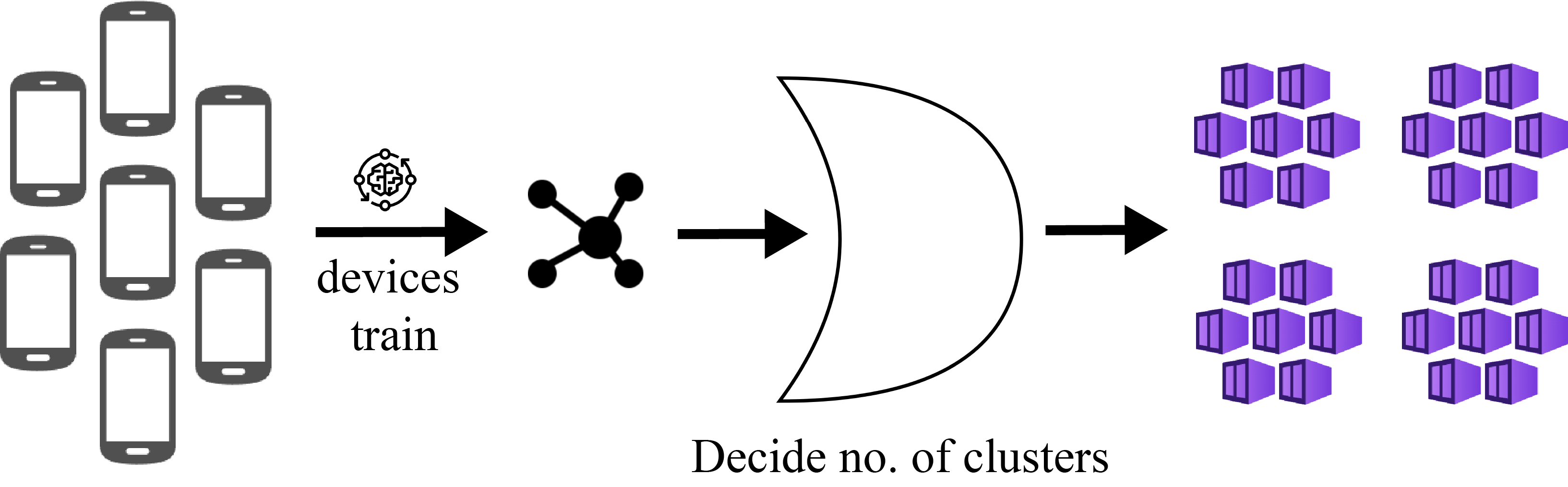}
    \caption{Number of clusters created following grouping mechanism.}
    \label{fig:adaptive_number_of_device_model}
    \end{figure}
    
    \item \textbf{Device Selection:}
    For device selection, we employ two methods, based on optimizer loss or random selection. 
    With objective-based selection, for a given group \( C \) containing devices \( d_i \) where \( i \in \{1, 2, \dots, n\} \), the selected device \( d^* \) is chosen so that it minimizes the objective function value
    \[
    d^* = \arg\min_{d_i \in C} f(d_i), \tag{5.0}
    \]
    where \( f(d_i) \) denotes the most recent objective function value of device \( d_i \).
    However, with the random selection method, the selection of a device \( d^* \) from a group \( C \) is probabilistic. Each device \( d_i \in C \) has an equal likelihood of being selected, mathematically expressed as
    \[
    P(d^*) = \frac{1}{|C|}, \tag{5.1}
    \]
    where, \( |C| \) denotes the number of devices in the cluster.
     \begin{figure}[!htbh]
    \centering
    \includegraphics[width=0.7\columnwidth]{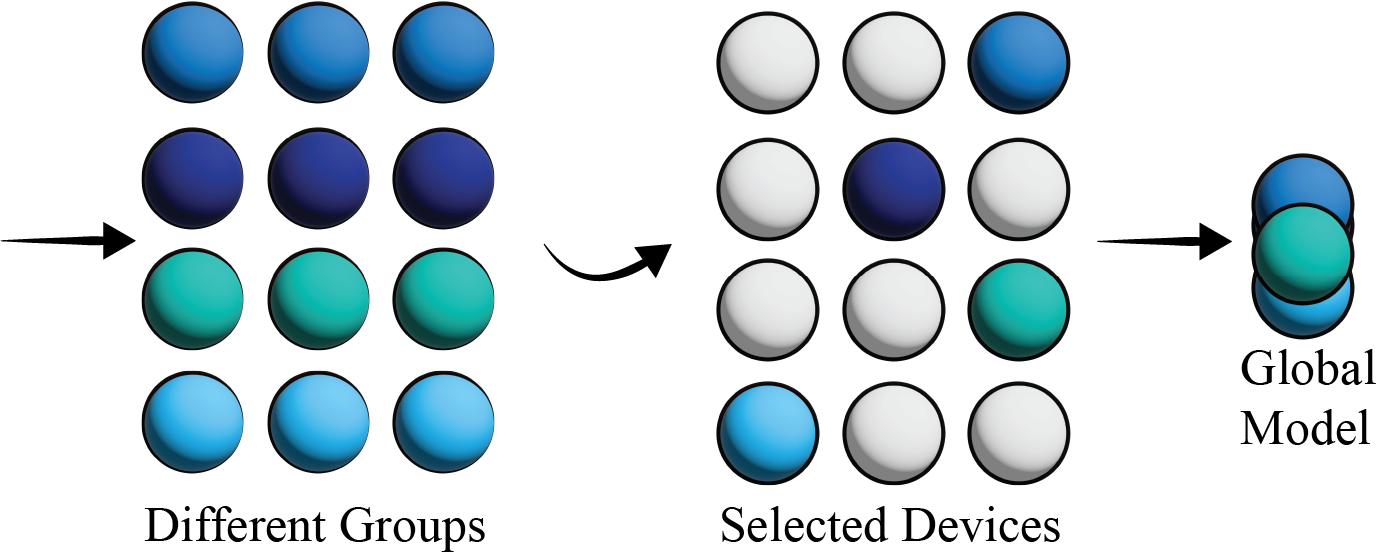}
    \caption{Device selection mechanism: Either probabilistic or objective-based selection; Selected devices contribute to device, cluster and global model.}
    \label{fig:device_selection}
    \end{figure}

\end{enumerate}

\subsection{Communication Model}
Let $T_{comm}$ represent the total communication time, $n_c$ represent the number of clusters, $n_d$ represent the total number of devices, $C_d$ represent the communication cost for a single device, $C_{agg}$ represent the communication cost for aggregating the model on the server, and $T_{train}$ represent the total training time.
The total communication time can be approximated as
\[
T_{comm} = n_c \cdot C_d + C_{agg}
\]
where, $n_c$ is determined by
\[
n_c = \max\left(1, \left\lceil \sqrt{\frac{n_d}{2}} \right\rceil \right)
\]
The total training time $T_{train}$ is proportional to the number of clusters
\[
T_{train} \propto n_c
\]
The complete communication model, which combines both training and communication time, can be written as
\[
T_{total} = T_{train} + T_{comm} = \alpha n_c + \left(n_c \cdot C_d + C_{agg}\right)
\]
where $\alpha$ is a proportional constant representing the per-cluster training time.
To assess the impact of selecting only one device from each cluster, we analyze both communication and training time in the QFL and mdQFL models. By selecting a single device from each cluster for training, we reduce both communication overhead and total training time.
In default QFL algorithm, all devices participate in communication with the server, leading to a communication time proportional to the total number of devices, $n_d$. The communication time can be expressed as
\[
T_{comm}^{QFL} = n_d \cdot C_d
\]
where, $C_d$ represents the communication cost for a single device. However, in the proposed model, only one device per cluster communicates with the server, reducing the communication time to
\[
T_{comm}^{mdQFL} = n_c \cdot C_d + C_{agg}
\]
where $n_c$ is the number of clusters (with $n_c < n_d$) and $C_{agg}$ is the cost of aggregating the model at the server. By selecting fewer devices for communication, we achieve a significant reduction in communication overhead.
Again in default QFL algorithm, all devices participate in training, leading to a training time proportional to the number of devices, $n_d$
\[
T_{train}^{QFL} \propto n_d
\]
In contrast, in proposed approach, only one device from each cluster is selected for training, and the training time is proportional to the number of clusters, $n_c$
\[
T_{train}^{mdQFL} \propto n_c
\]
This leads to a substantial reduction in the total training time, as $n_c$ is significantly smaller than $n_d$.
The total time for communication and training in the QFL can be expressed as:
\[
T_{total}^{QFL} = T_{train}^{QFL} + T_{comm}^{QFL} = \alpha n_d + n_d \cdot C_d
\]
where, $\alpha$ is a proportional constant representing the training time per device.
In the mdQFL model, the total time for communication and training is given by
\[
T_{total}^{mdQFL} = T_{train}^{mdQFL} + T_{comm}^{mdQFL} = \alpha n_c + \left(n_c \cdot C_d + C_{agg}\right)
\]
The overall improvement in performance can be quantified as the ratio of the total time in the QFL model to the total time in the mdQFL model
\[
\text{Performance\_improvement} = \frac{T_{total}^{QFL}}{T_{total}^{mdQFL}}
\]
Since $n_c < n_d$, the mdQFL model demonstrates a significant reduction in both communication and training times, thus improving the overall efficiency of the system.

The proposed approach reduces both the communication and training overhead by selecting only one device per cluster for training and communication with the server. Communication time decreases from $n_d \cdot C_d$ to $n_c \cdot C_d + C_{agg}$, while training time decreases from proportional to $n_d$ to proportional to $n_c$. This reduction results in an overall improvement in performance, as shown by the decrease in $T_{total}$.

\section{Stability and Convergence Analysis} \label{sec:theoretical_analysis}
In this subsection, we provide a range of theoretical analyses, explore multiple aspects, and offer the convergence analysis of mdQFL, while comparing it with the standard QFL.

\subsection{Assumptions}
Initially, we present the assumptions necessary for the theoretical framework.
The first three are common assumptions found in most literature, and the last two can be easily relaxed with slight adjustments to the algorithm.

\begin{assumption}
For COBYLA, in terms of objective functions, we can assume that the same Lipschitz continuity holds, which helps to ensure that the linear approximations used by COBYLA are accurate over the trust region, leading to more stable and faster convergence.

\end{assumption}
This assumption is important to ensure stable and efficient convergence.

\begin{assumption}
    The functions \(\{ F_1, \dots, F_N \}\) are all \( L \)-smooth, where \( L \geq 0 \). For any vectors \( x \) and \( y \), it holds that
    \begin{equation} \label{eqn:lsmooth}
        F_k(y) \leq F_k(x) + \langle y - x, \nabla F_k(x) \rangle + \frac{L}{2} \|y - x\|^2
    \end{equation}
    for each \( k = \{1, 2, \dots, N \} \), which is also a quadratic upper bound for an \( L \)-smooth function.
\end{assumption}

This ensures that the functions can be approximated by a quadratic function, providing a predictable and controlled change in function value for optimization.

\begin{assumption} 
Functions \(\{ F_1, \cdots, F_N \}\) are all \(\mu\)-strongly convex: for all \( x \) and \( y \), 
\begin{equation}\label{eqn:convex}
    F_k(x) \geq F_k(y) + (x - y)^T \nabla F_k(y) + \frac{\mu}{2} \|x - y\|_2^2.
\end{equation}
\end{assumption}

This ensures that the optimization process converges to the global minimum.

\begin{assumption}
Consider a set of \(n\) clients participating, denoted as \(\{C_1, C_2, \dots, C_K\}\). 
We assume that the data distributions of the data sets of these clients, represented by probability distributions \(\{P_1, P_2, \dots, P_K\}\), are heterogeneous and non-identically distributed, meaning that:
\begin{equation} \label{eqn:distribution}
    \exists i, j \in \{1, 2, \dots, K\}, \quad \delta(P_i, P_j) > \epsilon
\end{equation}

where \(\delta(\cdot, \cdot)\) denotes a divergence measure (e.g. total variation distance, Kullback-Leibler divergence) between the distributions and \(\epsilon\) is a positive constant representing a threshold beyond which the divergence between any pair of distributions is considered significant.
\end{assumption}

This assumption in an experimental context can be interpreted as an MNIST dataset being allocated across devices, with each device holding only certain labels.
This guarantees that optimal conditions are taken into account in the analysis.

\begin{assumption}

Along with the previous assumption, we also need to consider that at least a group of devices should be closely related in such a way that there is some similarity in character between the devices.
\end{assumption}
This assumption is necessary because completely unique devices with different datasets cannot be trained for the same model. Thus, we consider that there is some degree of homogeneity between some devices.
However, we do not introduce extra parameters to address this in our experimental analysis. In the case of the MNIST dataset with 10 labels 0 to 9, if we distributed in a l-cycle structure, we distributed at least 2 labels to each device. Distributing only one label to each device would be extremely non-IID. However, there will be some devices that will overlap just one label with the previous and next devices.

\subsection{Convergence Conditions}
A distributed optimization model can be formulated as \cite{liConvergenceFedAvgNonIID2019}:
\[
\min_\theta \left\{ F(\theta) \triangleq \sum_{i=1}^{n} p_i F_i(\theta) \right\}
\]

where, $n$ represents the total number of devices and $p_i$ denotes the weight factor of the device $i^{th}$, ensuring $p_i \geq 0$ and \(\sum_{i=1}^{N} p_i = 1\). 
Assume that each $i^{th}$ device has $m_i$ samples of training data \{$|\psi(x)\rangle_{i,1}, |\psi(x)\rangle_{i,2}, \dots, |\psi(x)\rangle_{i,m_i}\}$. 
The local objective function $F_i(\cdot)$ is defined as
% \cite{liConvergenceqFedAvgNonIID2019}
\[
F_i(\theta) = \frac{1}{m_i} \sum_{j=1}^{m_i} \ell(\theta; |\psi(x)\rangle_{i,j}), \tag{2}
\]
where $\ell(\cdot; \cdot)$ is a loss function defined by the device. 
In the standard FedAvg algorithm, the central server distributes the most recent global model, 
$\theta_g$, across all devices. 
Each device, particularly $i^{th}$, sets $\theta_{i}^{t} = \theta_g$ and then executes $E$ local updates:

\begin{equation} \label{eqn:sgd_update}
    \theta_{t+1} = \theta_{t} - \eta \nabla F_i(\theta_{t}, \xi_{t}), \quad t =\{0, 1, \dots, E - 1\}
\end{equation}
where, $\eta$ is the learning rate and $\xi_{t}$ is a randomly selected sample from the local data set. The server then compiles the local models, $\{\theta_{0}, \dots, \theta_{n}\}$, to form the updated global model, $\theta_{g}$.

\begin{lemma}
Let \( F \) be a continuous Lipschitz function with Lipschitz constant \( L \). The regret \( R(T) \) of the COBYLA optimizer 
 \cite{Powell1994} 
after \( T \) iterations is bounded by:
\begin{equation}
   R_F(T) = \sum_{k=1}^{T} [ F(\theta_t) - F(\theta^*) ] \leq L \sum_{t=1}^{T} \Delta_t
\end{equation}

where, \( \theta_t \) is the point in iteration \( t \), \( \theta^* \) is the optimal solution, and \( \Delta_t \) is the radius of the trust region at iteration \( t \).
\end{lemma}

The error bound at iteration \( t \) can be described in terms of the radius of the trust region \( \Delta_t \) and the accuracy of the linear approximations. 
With trust region radius bound,
\[
\| \theta_{t+1} - \theta^* \| \leq \Delta_t
\]
where \( \theta^* \) is the optimal solution and \( \Delta_t \) is the radius of the trust region at iteration \( t \). 
With an error bound of the objective function,
\[
| f(\theta_{t+1}) - f(\theta^*) | \leq C \Delta_t
\]
where, \( C \) is a constant dependent on the Lipschitz continuity of the objective function \( F \) and the accuracy.

\begin{theorem}[mdQFL]
    Let Assumptions 1 to 4 hold, and let \(L, \mu, \sigma_k, G\) be defined. Choose \(\kappa = \frac{L}{\mu}\), \(\gamma = \max\{8\kappa, N\}\), and the learning rate \(\eta_t = \frac{2}{\mu(\gamma + t)}\). Then, with partial participation of the device \cite{liConvergenceFedAvgNonIID2019}:

\begin{align}
    \mathbb{E}[F(\theta_T)] - F^* \leq
    &\frac{2\kappa}{\gamma + T} \left( \frac{B^{\ket{\psi}} + C}{\mu} + 2L \|\theta_0 - \theta^*\|^2 \right) \notag\\
    &+ L \sum_{t=1}^{T} \Delta_t
\end{align}
where
$B = \sum_{k=1}^N \sigma_k^2 + 6L \Gamma + 8(N - 1)^2 G^2 + \frac{1}{iter} + \frac{1}{\Delta_t},
$
and \(\Gamma = \sum_{k=1}^N (F^* - F_k^*)\).
\end{theorem}

The proofs of the lemmas and theorems are provided in the appendix \ref{sec:appendix}.

\subsection{Segregating Redundant Models}

Let $\theta_i$ and $\theta_j$ be the local model parameters of devices $i$ and $j$, respectively. If these devices have similar data distributions, their local model vectors will tend to be similar to some extent such that
\[
\|\theta_i - \theta_j\| \leq \epsilon,
\]
where, $\epsilon > 0$ is a small constant indicating the degree of similarity. Including both $\theta_i$ and $\theta_j$ in the aggregation can introduce redundancy without adding significant new information.

\section{Performance Evaluation}\label{sec:experimental_results}
In this section, we present an analysis of the experimental findings, underscoring the significance and validation of the proposed theoretical model.

\subsection{Experiment 1: Against baseline QFL}
\subsubsection{Setup}
In these experiments, the data set is divided into a training device set and a server device set. The data of the training set are subsequently distributed among all devices, and the test set is used for validation and testing on the server device. Each device also further splits its local data into training and testing subsets.
We assign data to devices using the l-cycle method to create a non-IID scenario with varying degrees. Initially, we allocate only $2$, $5$, or $8$ class labels (nClass = $2$, $5$, or $8$) in each l-cycle. In addition, we performed experiments with different numbers of devices. Specifically, for $50$ devices, the COBYLA optimizer's local iteration is set to $5$, while with $10$ devices it is set to $50$. We used the MNIST dataset and experimented with different sample sizes in some cases to speed up the experimental process.
The experiments were carried out in Google Colab using Qiskit library.

\begin{figure}[!htb]
    \centering
    \begin{subfigure}[]{0.4\columnwidth}
    \centering
    \includegraphics[width=\columnwidth]{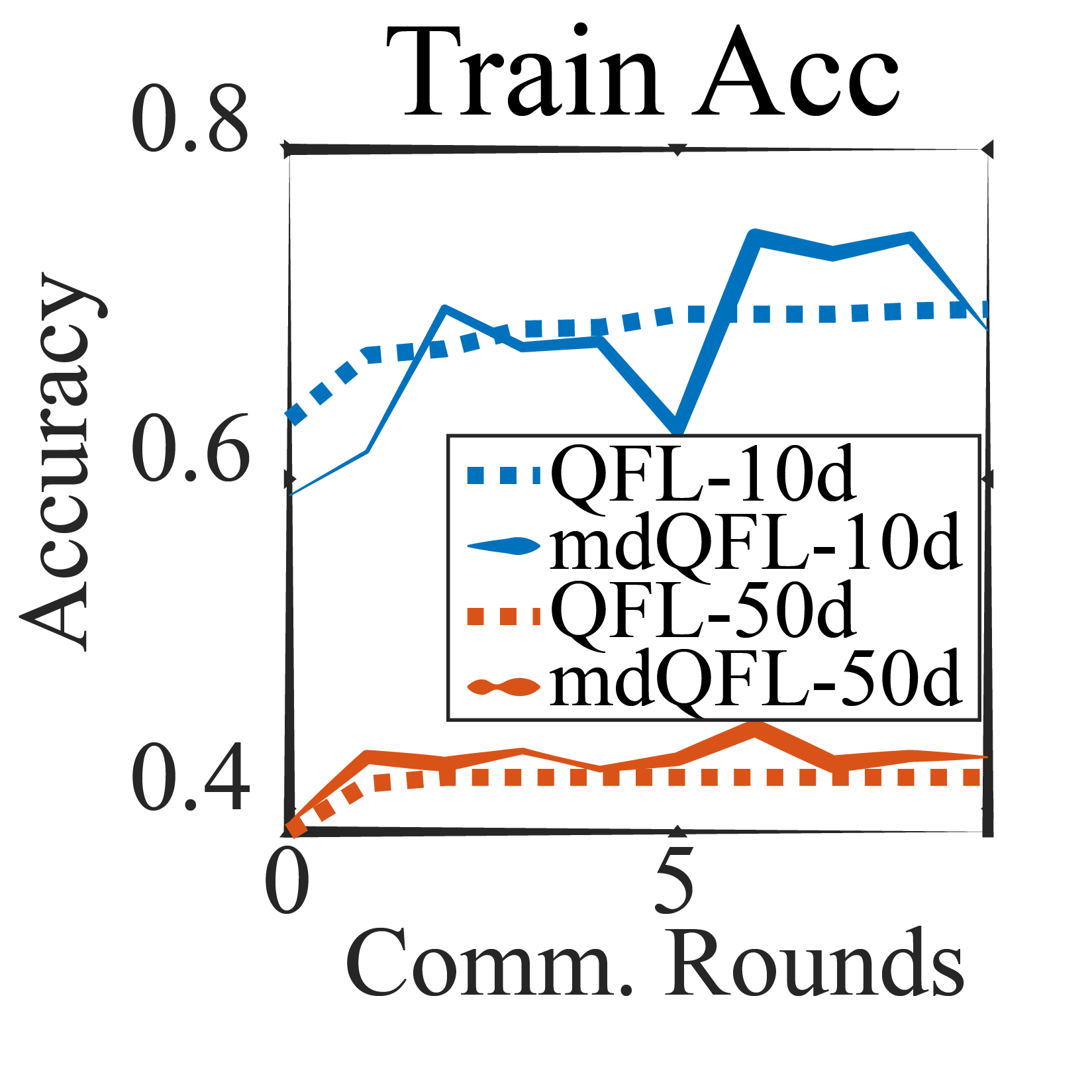}
    \caption{Train Accuracy}
    \label{fig:devices_train_acc}
    \end{subfigure}
    % \hfill
    \begin{subfigure}[]{0.4\columnwidth}
    \centering
    \includegraphics[width=\columnwidth]{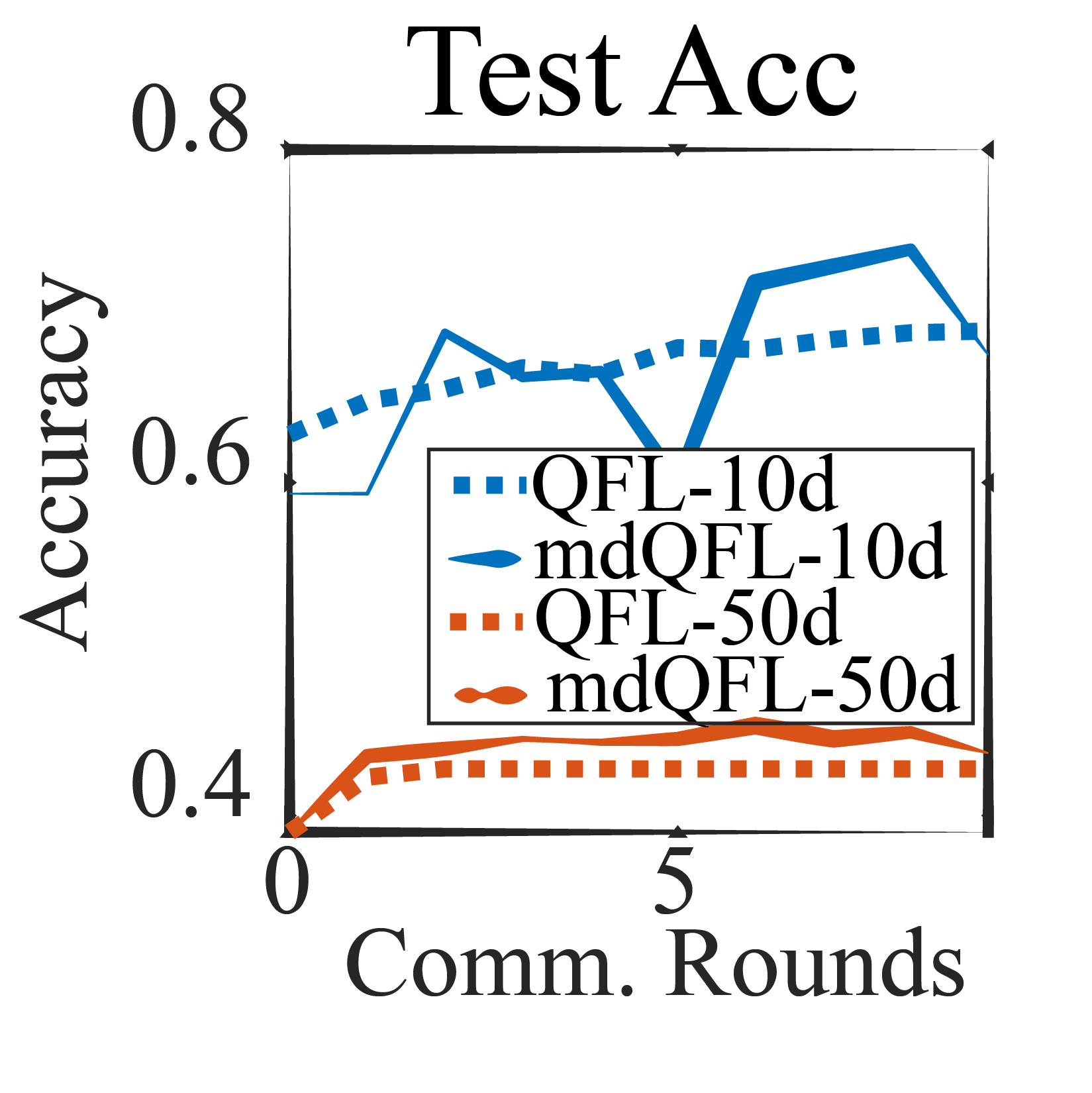}
    \caption{Test Accuracy}
    \label{fig:devices_test_acc}
    \end{subfigure}
     \centering
    \caption{Average Devices Performance: Average train and test accuracy of all devices in each communication round. The participating number of devices is $10$ ($50$ iter) and $50$ ($10$ iter).}
    \label{fig:devices_performance_accuracy}
\end{figure}

\subsubsection{Local Performance}
Figure \ref{fig:devices_performance_accuracy} shows a comparison between mdQFL and QFL with respect to the average performance of all devices in each round of communication. The proposed mdQFL appears to outperform QFL in terms of training and testing accuracy for varying numbers of devices.
This is because mdQFL is more personalized towards the local device.
The improvement in the local device performance is really interesting as it helps in the overall performance of the system.
It should be noted that, for mdQFL, the number of devices participating in training is equal to the number of clusters, which is always less than in QFL where all devices participate in training in each communication round. 
Thus, the average accuracy is obtained by varying the number of devices for both QFL and mdQFL.
In terms of performance variation, which we can see between $10$ and $50$ devices, it is due to the local iteration of the COBYLA optimizer. 
For $50$ devices, each device only runs $5$ iterations of the optimizer, which can be reflected in the performance degradation of the devices.

\subsubsection{Global Performance}
\begin{figure}[!htb]
    \centering
    \begin{subfigure}[]{0.45\columnwidth}
    \centering
    \includegraphics[width=\columnwidth]{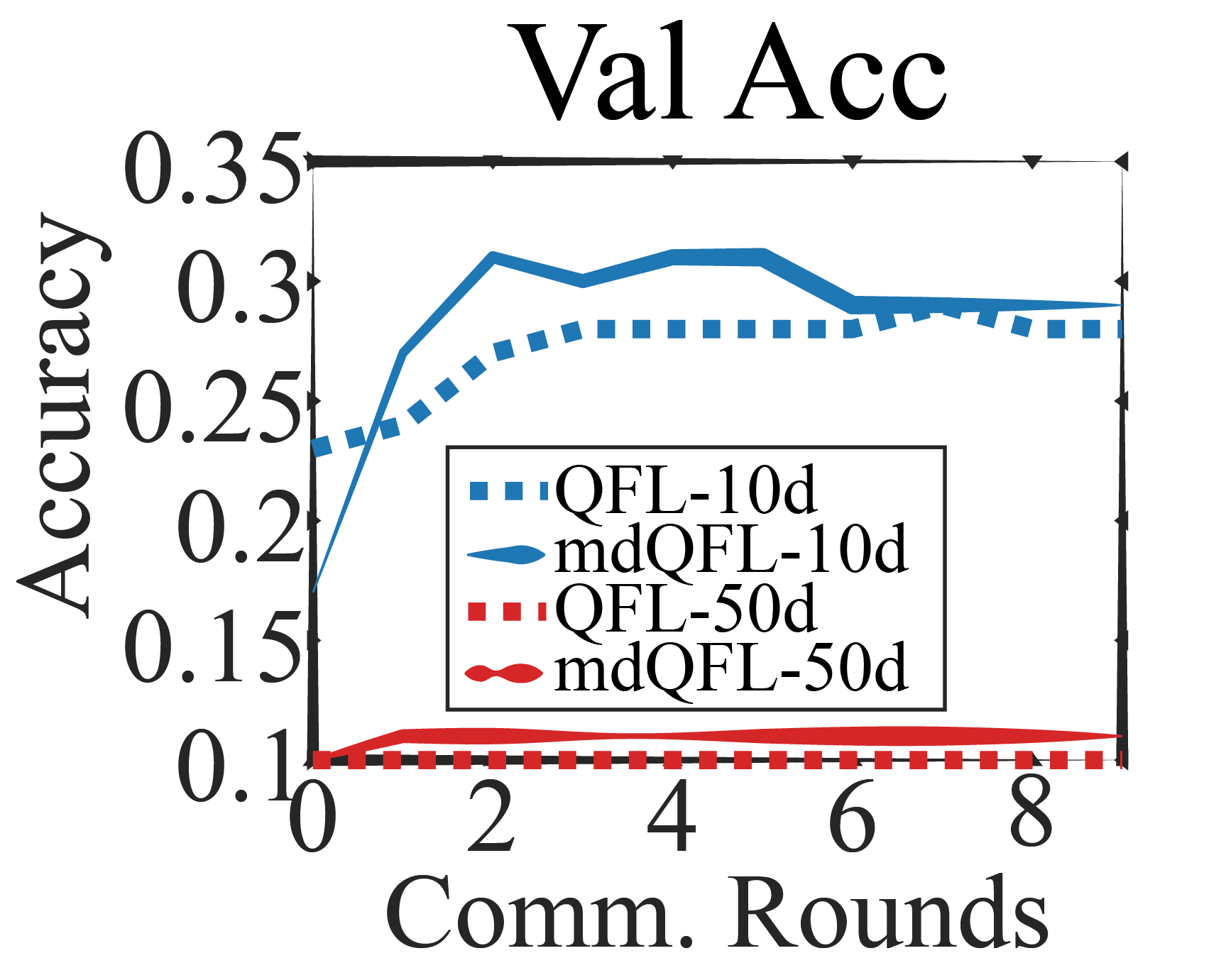}
    \caption{Val Accuracy}
    \label{fig:server_train_acc}
    \end{subfigure}
    % \hfill
    \begin{subfigure}[]{0.45\columnwidth}
    \centering
    \includegraphics[width=\columnwidth]{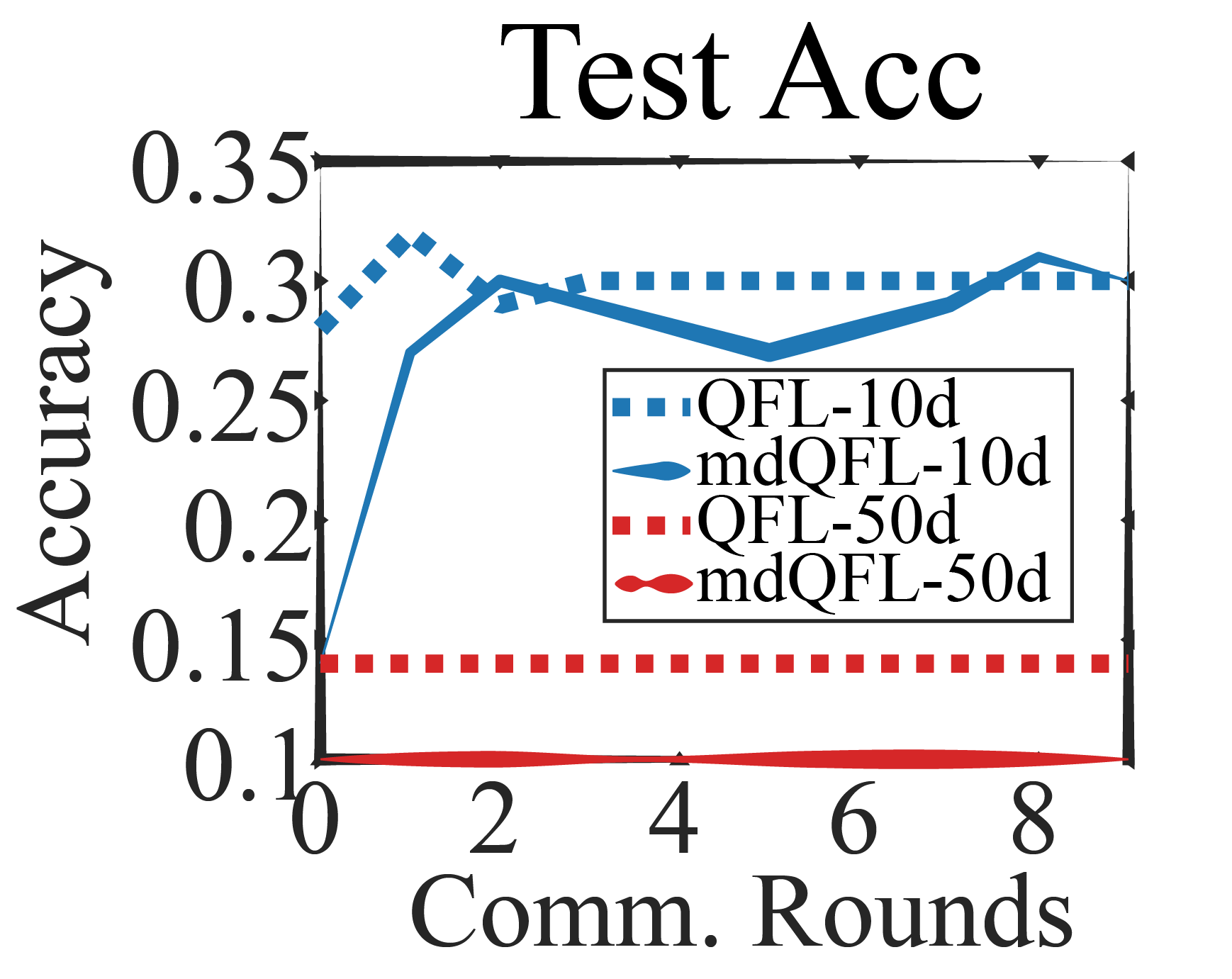}
    \caption{Test Accuracy}
    \label{fig:server_test_acc}
    \end{subfigure}
     \centering
    \caption{Server Performance: Validation and Test Loss on server validation and test datasets with varying number of devices and local iterations.}
    \label{fig:server_performance_accuracy}
\end{figure}

In Figure \ref{fig:server_train_acc} we can see that mdQFL demonstrates a sharp
increase in accuracy during initial communication rounds before stabilizing in terms of validation accuracy. 
This indicates that mdQFL is highly efficient in improving accuracy quickly. 
Similarly, QFL shows a steady increase in accuracy, though it remains
slightly below mdQFL after the initial rounds, indicating robust but
slightly less efficient performance. On the other hand, with 50 devices,  
mdQFL performs better than QFL. 
In Figure \ref{fig:server_test_acc}, in terms of test accuracy, the result is comparable between mdQFL and QFL.
 In contrast to validation accuracy, especially with 50 devices, QFL is slightly better than mdQFL.
The inefficiency observed in 50-device configurations of mdQFL could be indication of the need to create more clusters and thus more devices for training for better performance.

\begin{figure}[!htb]
    \centering
    \begin{subfigure}[]{0.45\columnwidth}
    \centering
    \includegraphics[width=\columnwidth]{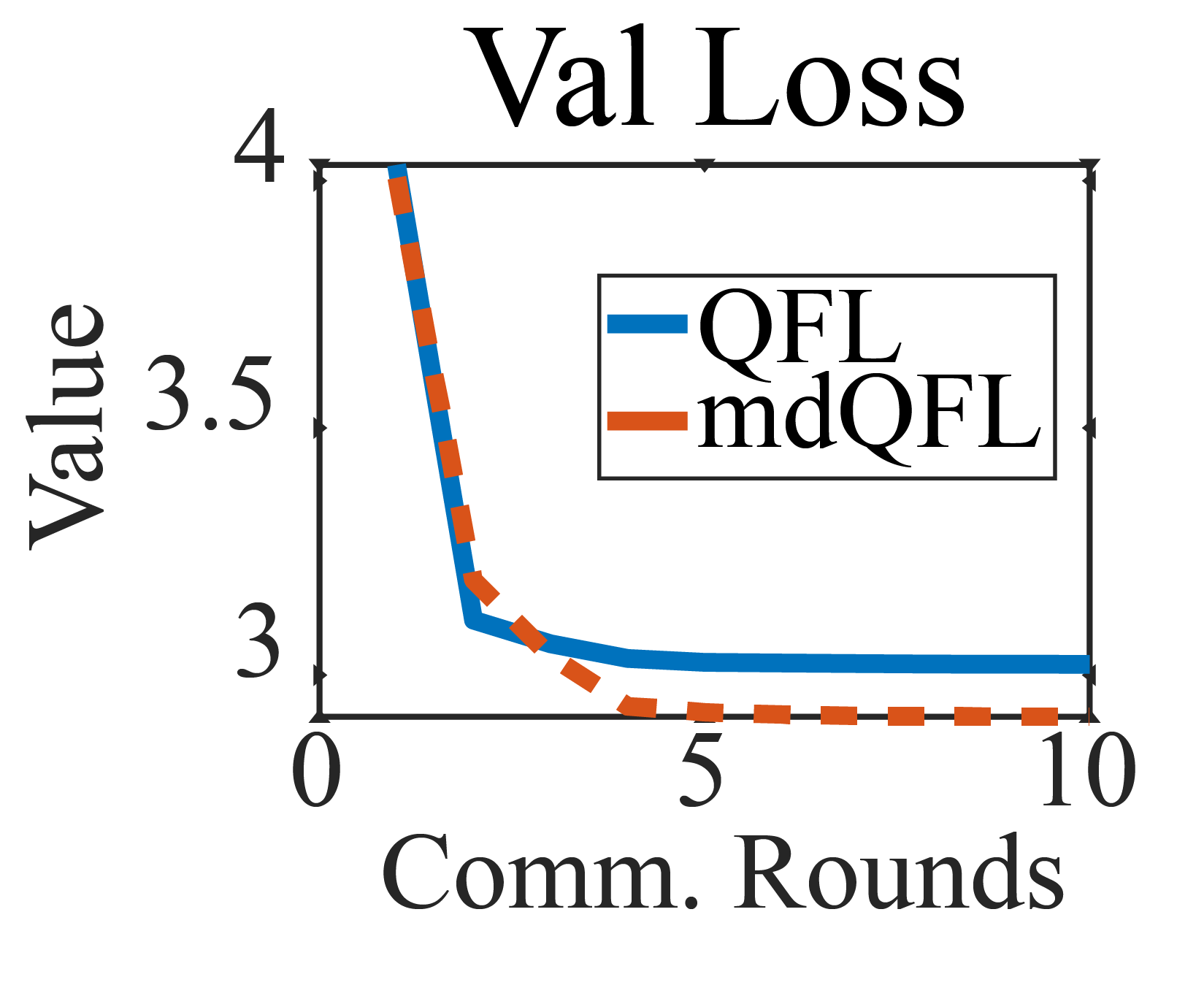}
    \caption{10 Devices; 50 iter.}
    \label{fig:validation_loss_10}
    \end{subfigure}
    \begin{subfigure}[]{0.45\columnwidth}
    \centering
    \includegraphics[width=\columnwidth]{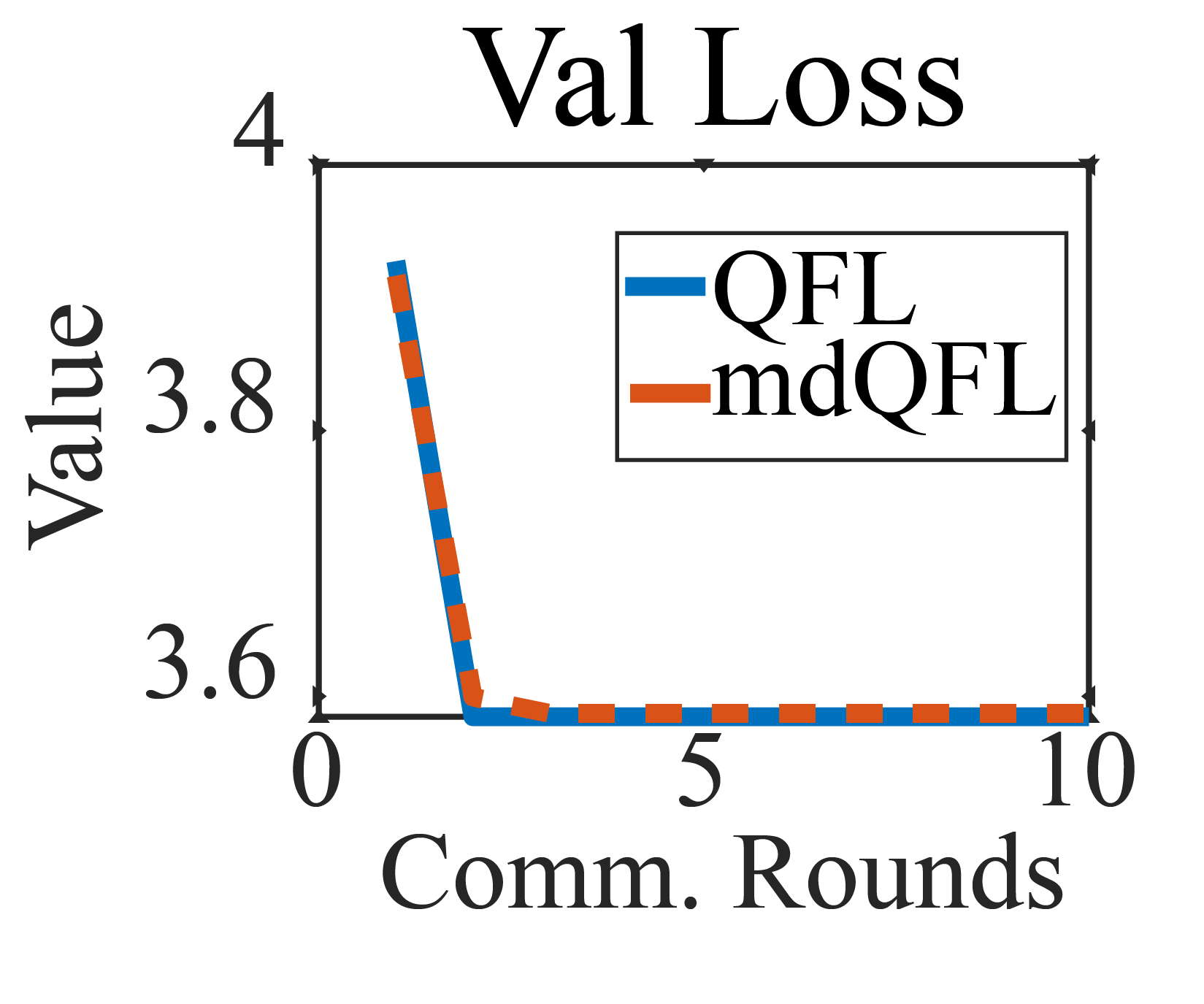}
    \caption{50 Devices; 5 iter.}
    \label{fig:validation_loss_50}
    \end{subfigure}
     \centering
    \caption{Server Performance: Validation loss of global model on validation dataset.}
    \label{fig:server_validation_loss}
\end{figure}

In Figure \ref{fig:validation_loss_10}, mdQFL performs consistently better than QFL, as evidenced by its
lower objective values throughout iterations and its continued decline, while QFL stabilizes around a higher objective value. 
This shows that the global model is capable of generalizing the unseen data pretty well. 
This suggests that mdQFL is more effective in reducing the objective value and
potentially offering better optimized performance.
However, as seen in Figure \ref{fig:validation_loss_50}, with a higher number of devices, the performance is almost equivalent, indicating that more customization is required in terms of the number of clusters or devices participating either to increase or decrease.

\subsubsection{Communication Time}
\begin{figure}[!htb]
    \centering
    \begin{subfigure}[h]{0.4\columnwidth}
    \centering
    \includegraphics[width=\columnwidth]{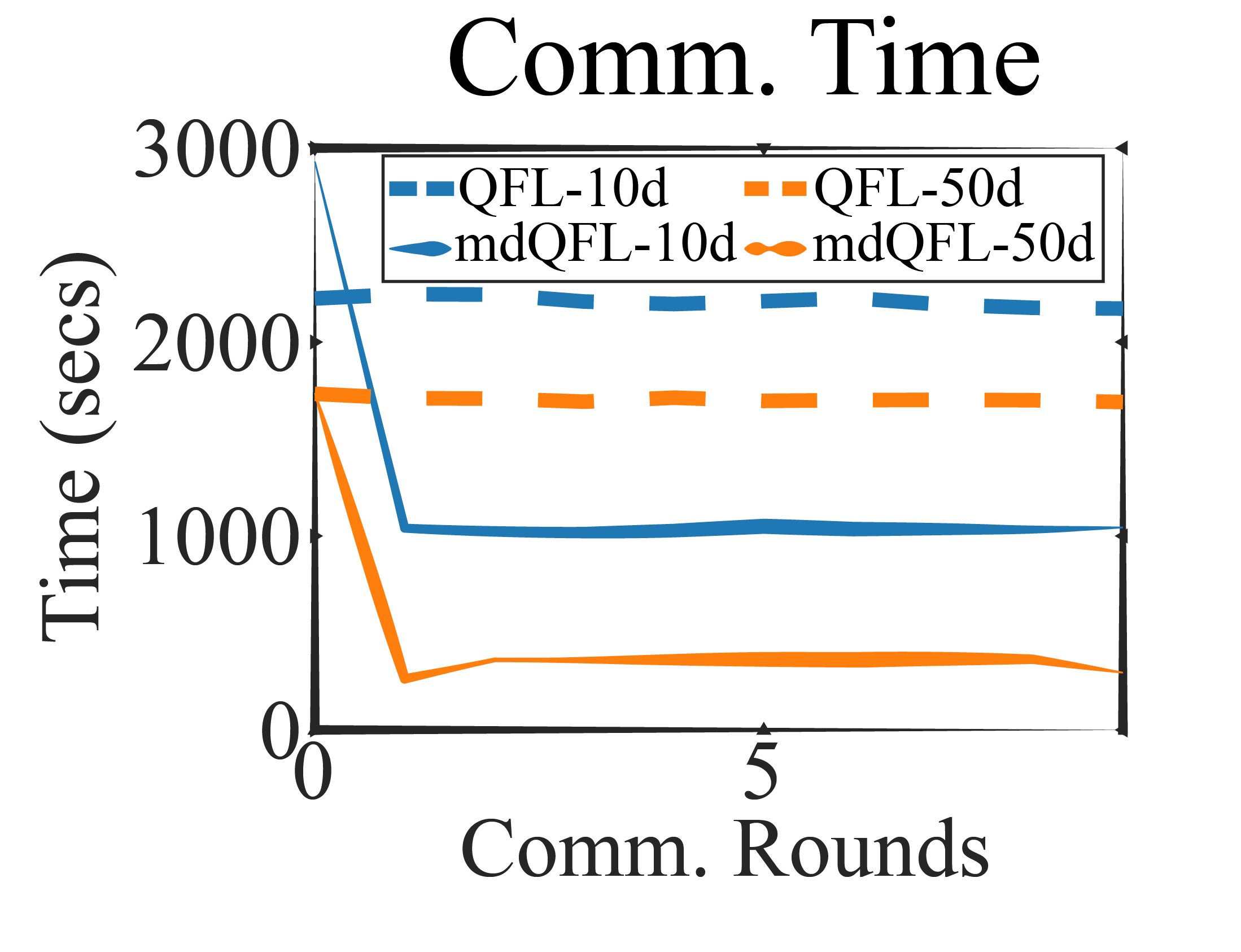}
   \caption{10d, 50d; MNIST}
   \label{fig:comm_time_nclass2)}
    \end{subfigure}
    \begin{subfigure}[h]{0.36\columnwidth}
    \centering
    \includegraphics[width=\columnwidth]{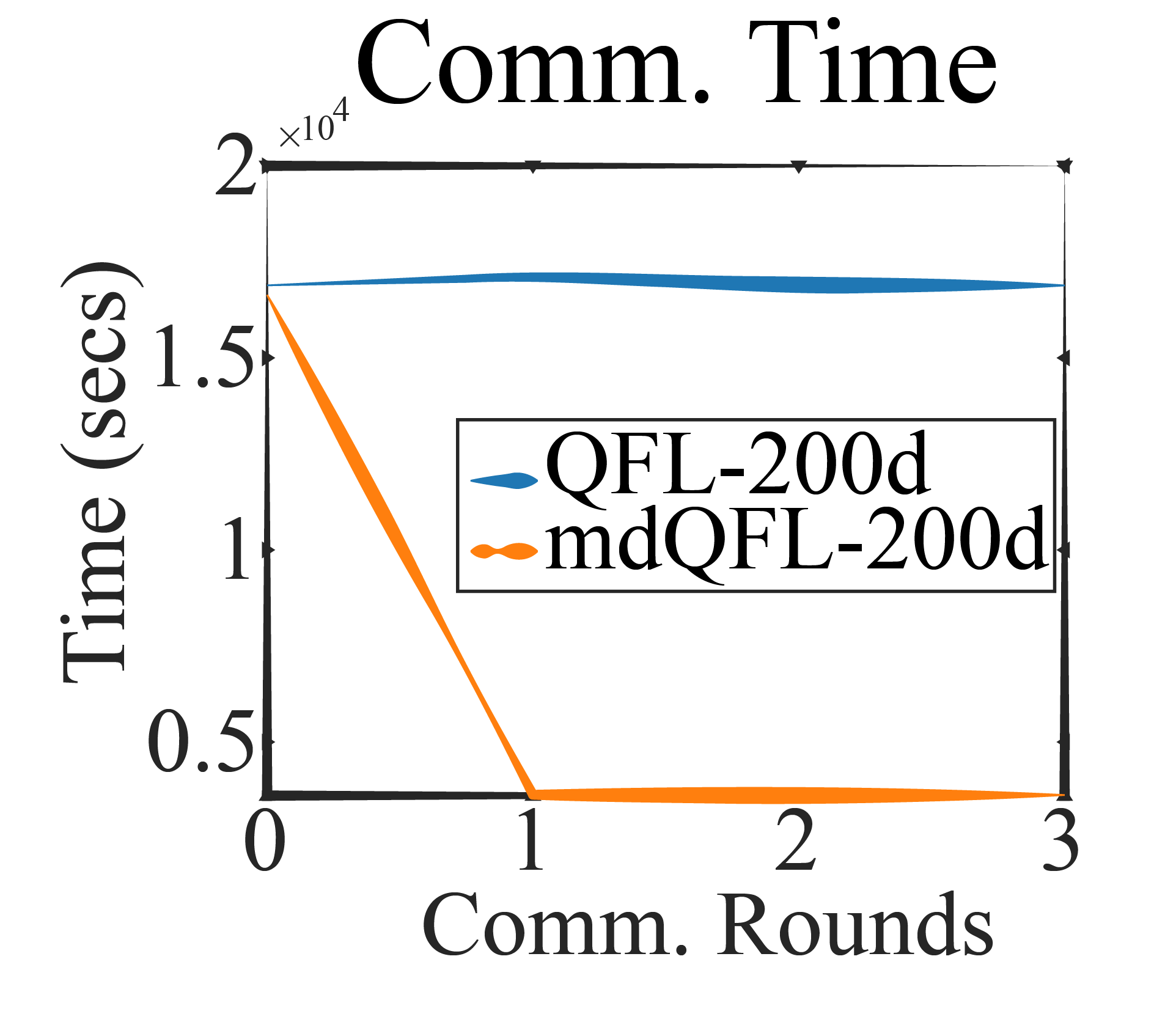}
   \caption{200d; Genomic}
   \label{fig:comm_time_200d)}
    \end{subfigure}
    \caption{Comparison between mdQFL and QFL with 10, 50,  200 devices in terms of communication time with MNIST and Genomic dataset.}
    \label{fig:comm_time}
\end{figure}

Figure \ref{fig:comm_time} demonstrates the communication times for both mdQFL and QFL. The graph reveals that mdQFL configurations are notably more efficient compared to QFL setups over numerous rounds. Although mdQFL has a higher initial communication time due to setup overhead, configurations using 50 and 10 devices show a significant decrease and stabilization in the following rounds, indicating superior efficiency. However, QFL setups maintain relatively consistent communication times, despite the efficiency gains seen in mdQFL. In general, the considerable reduction in communication time for mdQFL after the first round highlights its efficiency and potential for faster convergence. The initial overhead in mdQFL is due to the need to initialize the quantum circuit for all devices, along with conducting a training round for all devices during the first communication round.
Considering the performance metrics of both the server and the devices from prior results, the enhancement in communication time is notable, nearly accomplishing a reduction in communication overhead by a factor of two to four.
With almost similar or sometimes improvement in some cases with so much reduction in communication overhead, developing more efficient QFL frameworks with insights into unique new approaches can be a crucial step forward.
It is also interesting that the efficiency is directly proportional to the number of devices or the number of local iterations.
With 200 devices, QFL takes $1.8$ x ${10^4}$ seconds, while mdQFL takes $0.4$ x $10^4$ seconds, which is almost 4 times, whereas with 10 and 50 devices, the improvements are 2 to 3 times better. 

\subsection{Experiment 2: Variations of mdQFL}
In this set of experiments, we perform various experimental analyzes based on degree of personalization and generalization to see how they compare with each other in terms of communication time, local and server performance along with the standard QFL approach mentioned in Section \ref{sec:adaptive}.
We perform an experiment with a combination of various customizations to study how they impact the overall performance of the mdQFl framework
as in Table \ref{table:experiments_adaptive}.
The combination [0,2,0] means that we use eqn. 1.0 for train personalization, 2.2 for update personalization, and 3.0 for test generalization.
There can be many variations and combinations, but we choose these combinations from [0,2,0] to [1,1,2] ranging from more global model influence to local model influence. 

\begin{table}[h!]
\centering
\caption{Varying experimental combinations for mdQFL}
\begin{tabularx}{\columnwidth}{|c|X|}
\hline
\raisebox{1.5ex}{\textbf{Values}} & \raisebox{1.5ex}{\textbf{Customization to personalization and generalization.}} \\ \hline
{[0, 2, 0]} & High generalization, all global models involved. Uses from eqn. 1.0, 2.2, 3.0. \\ \hline
{[1, 0, 2]} & All cluster models involved. Uses from eqn. 1.1, 2.0, 3.2. \\ \hline
{[1, 1, 1]} & Device and cluster models involvement. Uses from eqn. 1.1, 2.1, 3.1. \\ \hline
{[1, 2, 1]} & Hybrid model with involvement of global and clusters. Uses from eqn. 1.0, 2.2, 3.1. \\ \hline
{[1, 1, 2]} & Cluster, global, and more device models involvement. Uses from eqn. 1.0, 2.1, 3.2. \\ \hline
\end{tabularx}
\label{table:experiments_adaptive}
\end{table}

We perform various sets of experiments with MNIST dataset with sample sizes from 1000 (small1000), 10000 (small10000) and normal to achieve the experimental analysis. The local COBYLA optimizer has local iterations of 10 and 100 with a varying number of devices between 20d and 50d. 
For the class label distribution, we experiment with the l-cycle structure with n2, n3, and n5.

\subsubsection{Training Time}
\begin{figure}[!htb]
\centering
\hspace{0.5em}
    \begin{subfigure}[]{0.45\columnwidth}
        \centering
        \includegraphics[width=0.9\columnwidth]{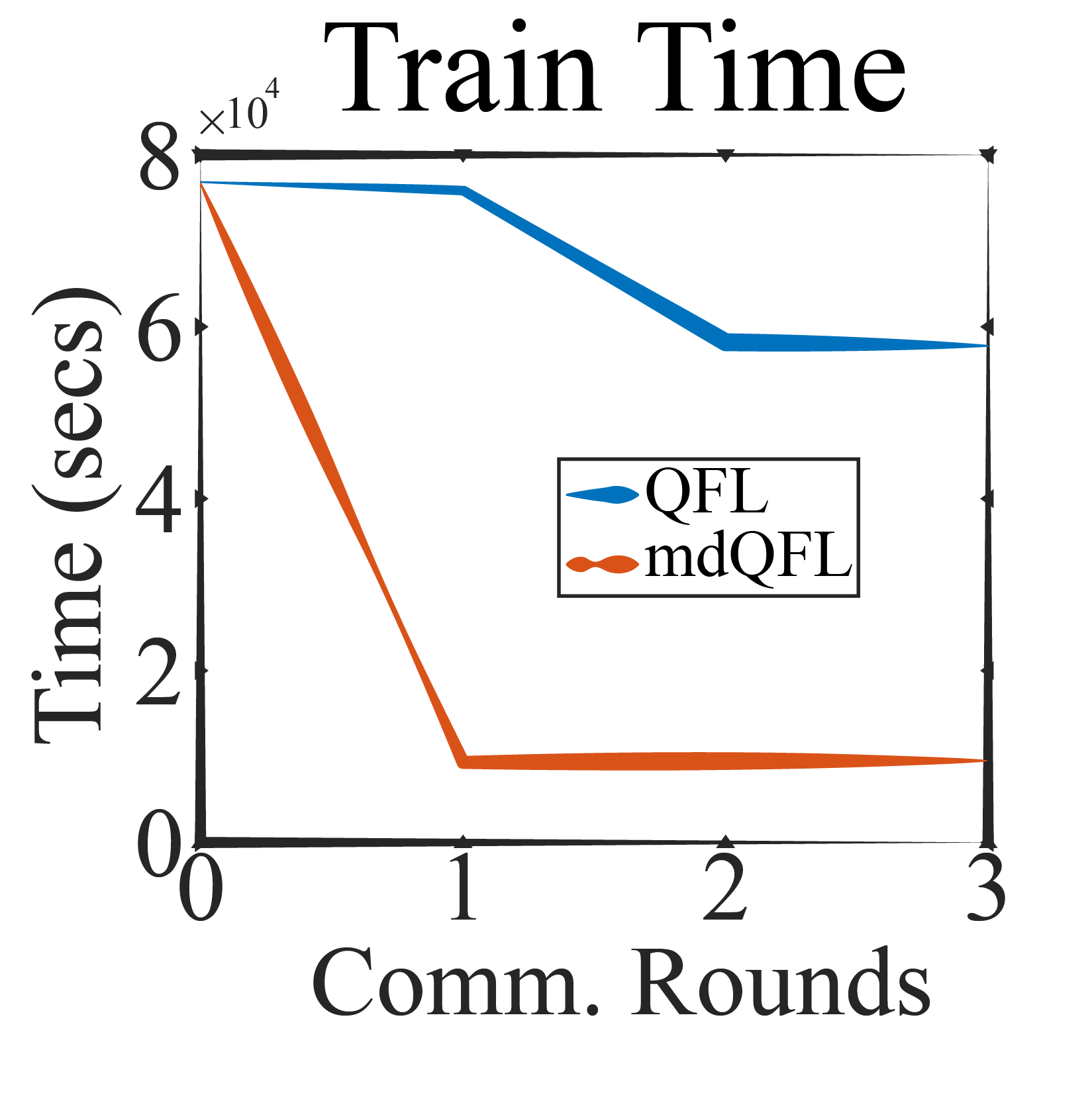}
        \caption{Normal, n2, 50d}
        \label{fig:normal_100iter_50d_nclass2}
    \end{subfigure}
    \begin{subfigure}[]{0.45\columnwidth}
        \centering
        \includegraphics[width=0.9\columnwidth]{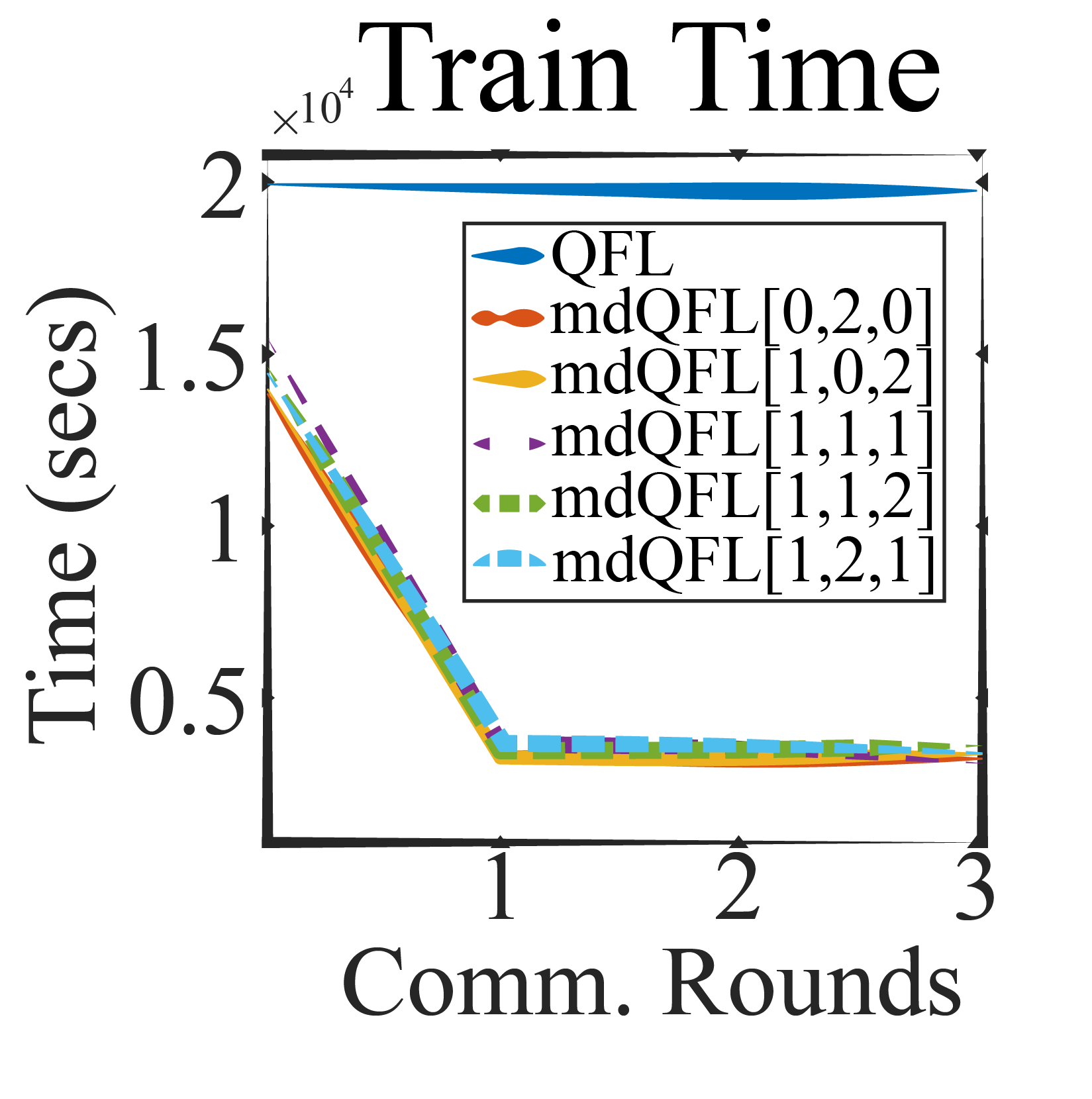}
        \caption{Small10000, n3, 20d}
        \label{fig:mnist_small10000_100iter_20d_nclass3}
    \end{subfigure}
    \begin{subfigure}[]{0.45\columnwidth}
        \centering
        \includegraphics[width=\columnwidth]{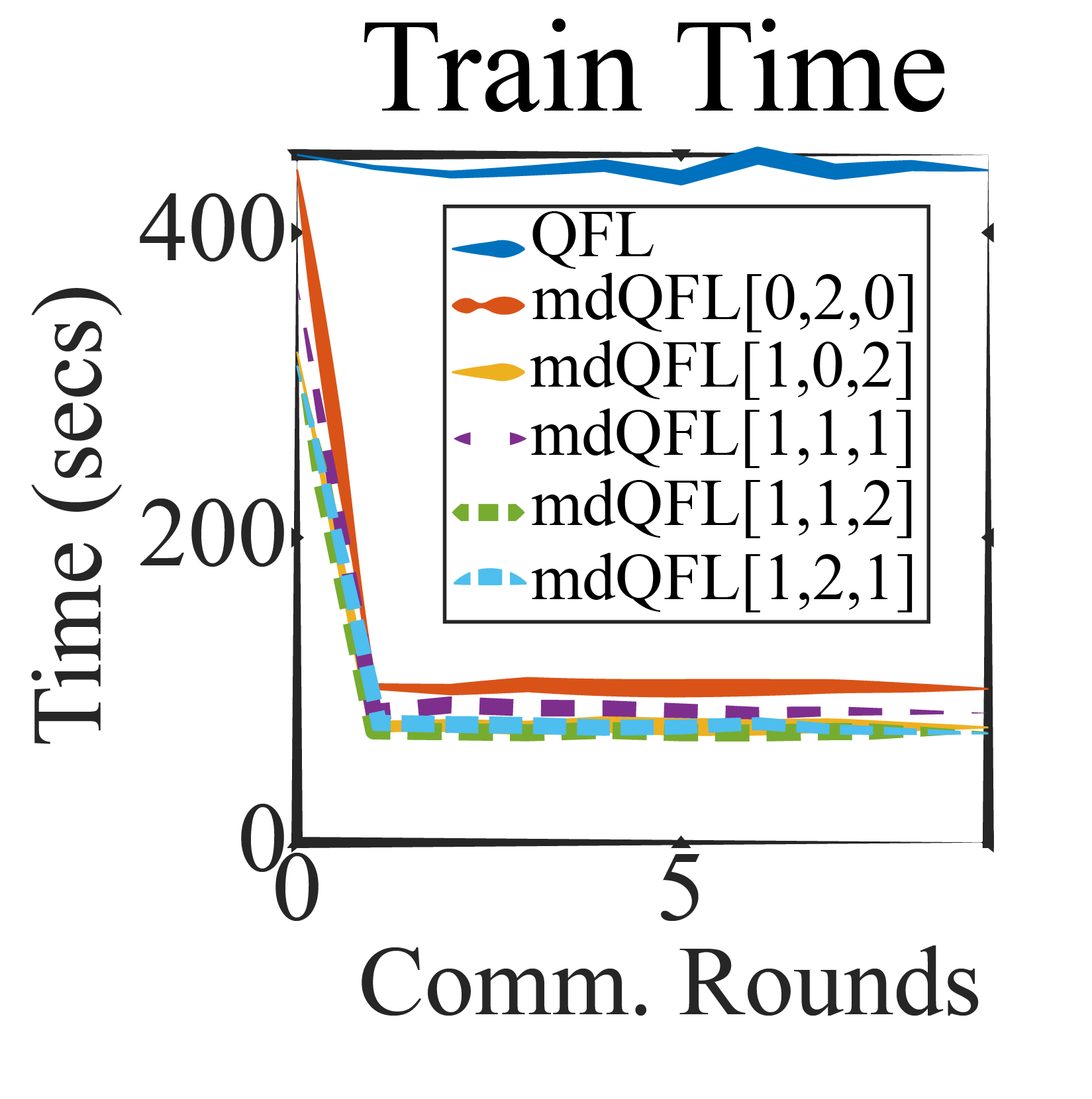}
        \caption{Small2000, n3, 20d}
        \label{fig:mnist_small2000_10iter_20d_nclass3}
    \end{subfigure}
    \begin{subfigure}[]{0.45\columnwidth}
        \centering
        \includegraphics[width=\columnwidth]{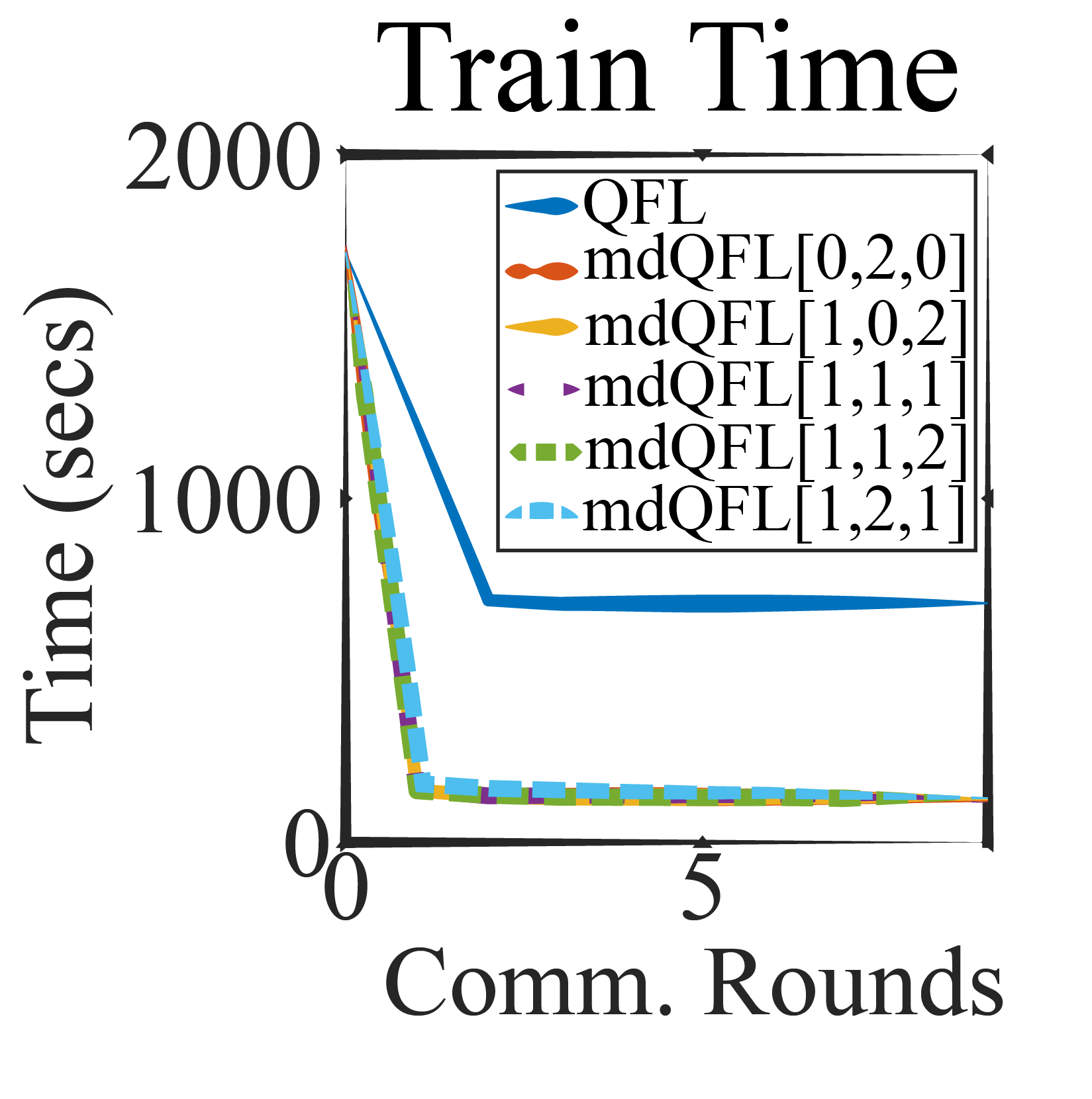}
        \caption{Small1000, n5, 50d}
        \label{fig:mnist_small1000_10iter_50d_nclass5}
    \end{subfigure}
    \caption{Training time comparison between variations of mdQFL vs QFL with varying data size, number of devices, local iterations, and class distribution.}
    \label{fig:training_time}
\end{figure}

As illustrated in Figure \ref{fig:training_time}, the proposed mdQFL framework demonstrates a significant benefit by consistently improving communication efficiency, thereby substantially lowering communication costs, an essential factor in distributed machine learning.
The outcomes consistently depict a significant decrease in communication overhead across different device counts, local iterations, class distributions, and data sizes.
This is essential because as we continue to see a significant increase in data volume, the demand for greater computational power and optimizations intensifies, necessitating quicker training and more efficient machine learning frameworks.

\subsubsection{Global Performance}
\begin{figure}[!htb]
    \centering
    \begin{subfigure}[]{0.4\columnwidth}
    \centering
    \includegraphics[width=\columnwidth]{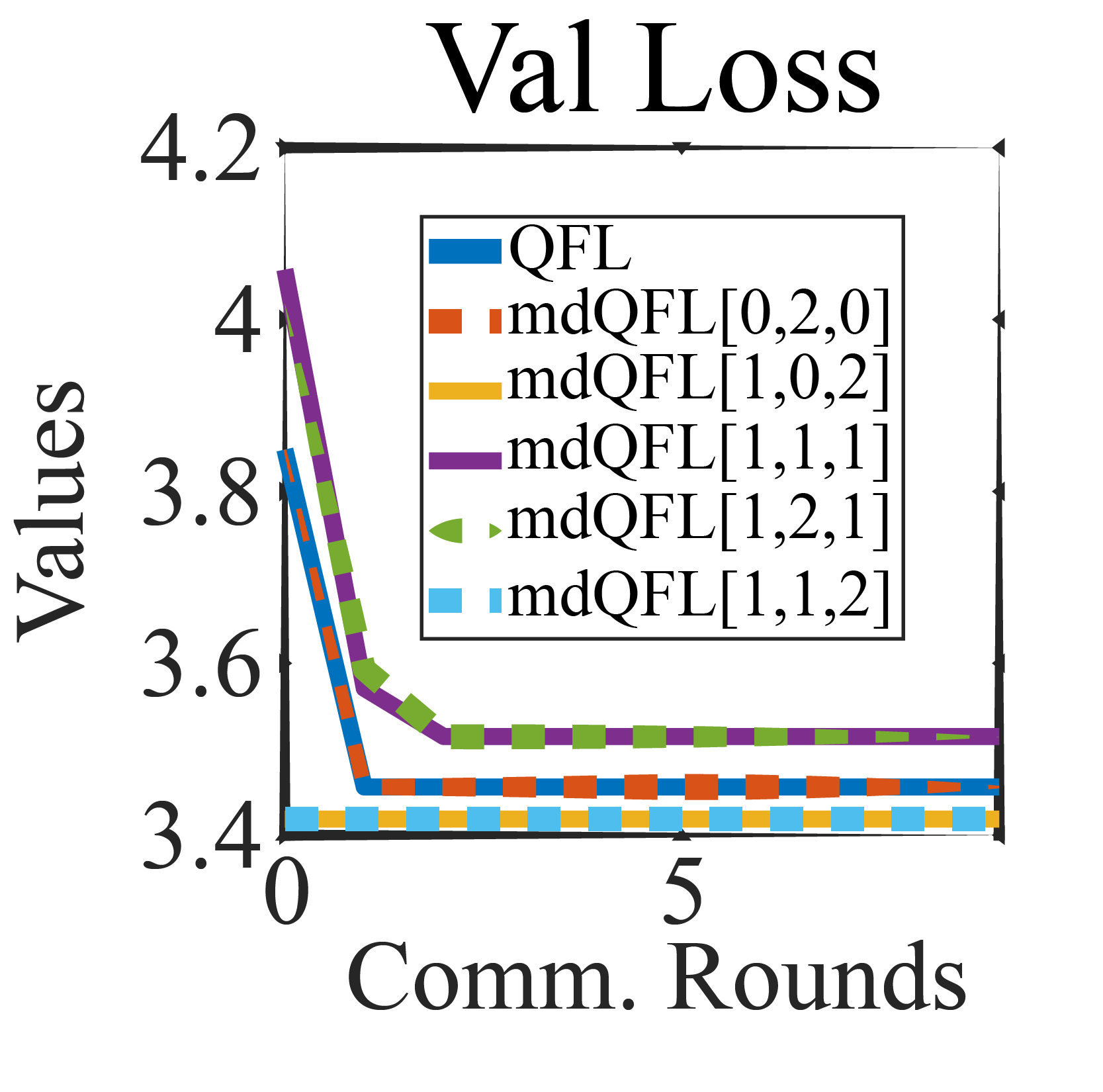}
   \caption{Normal, n3}
   \label{fig:val_loss_small2000_10iter_20d_nclass3}
    \end{subfigure}
    \begin{subfigure}[]{0.4\columnwidth}
    \centering
    \includegraphics[width=\columnwidth]{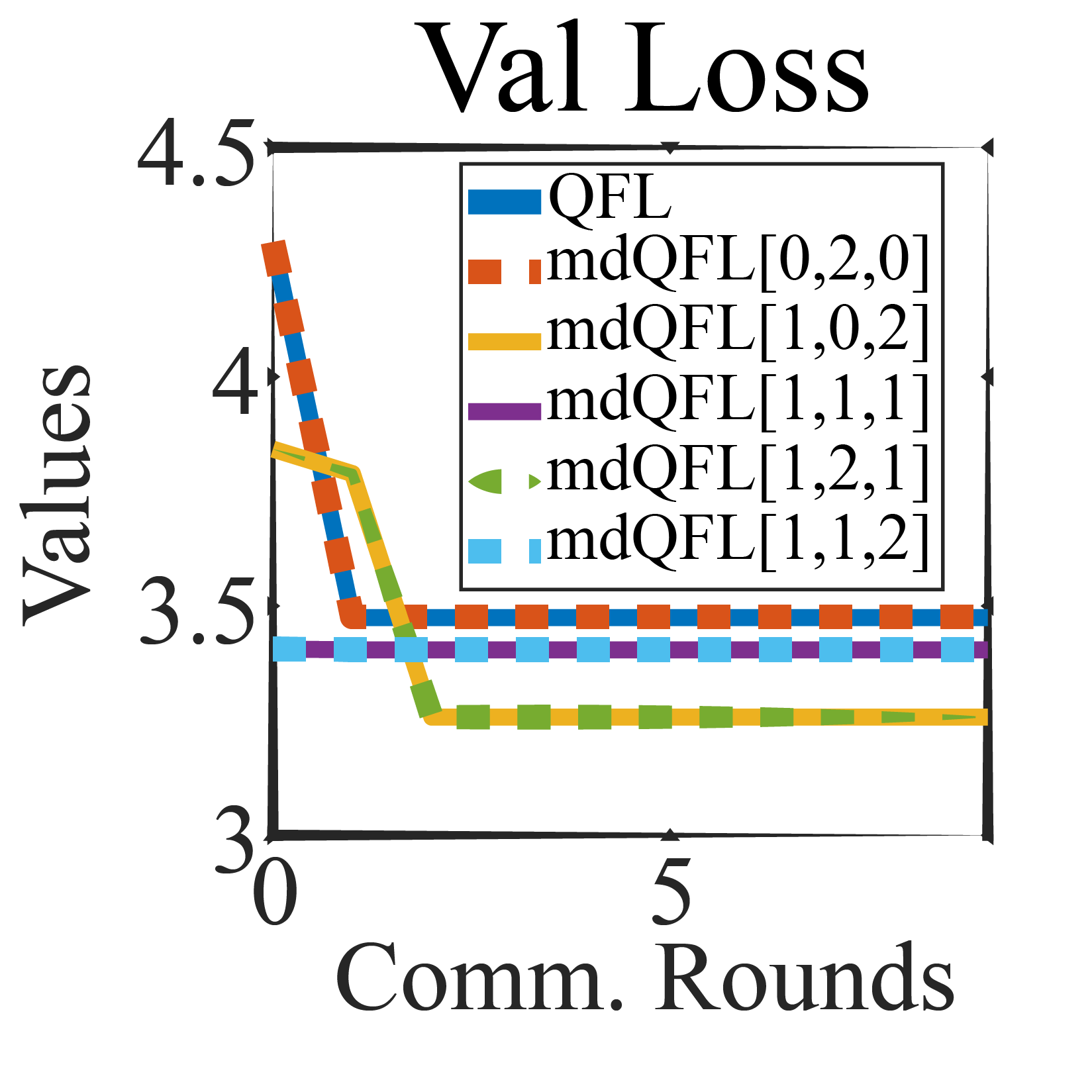}
   \caption{Small1000, n5}
   \label{fig:val_loss_small1000_10iter_50d_nclass5}
    \end{subfigure}
     \begin{subfigure}[]{0.4\columnwidth}
    \centering
    \includegraphics[width=\columnwidth]{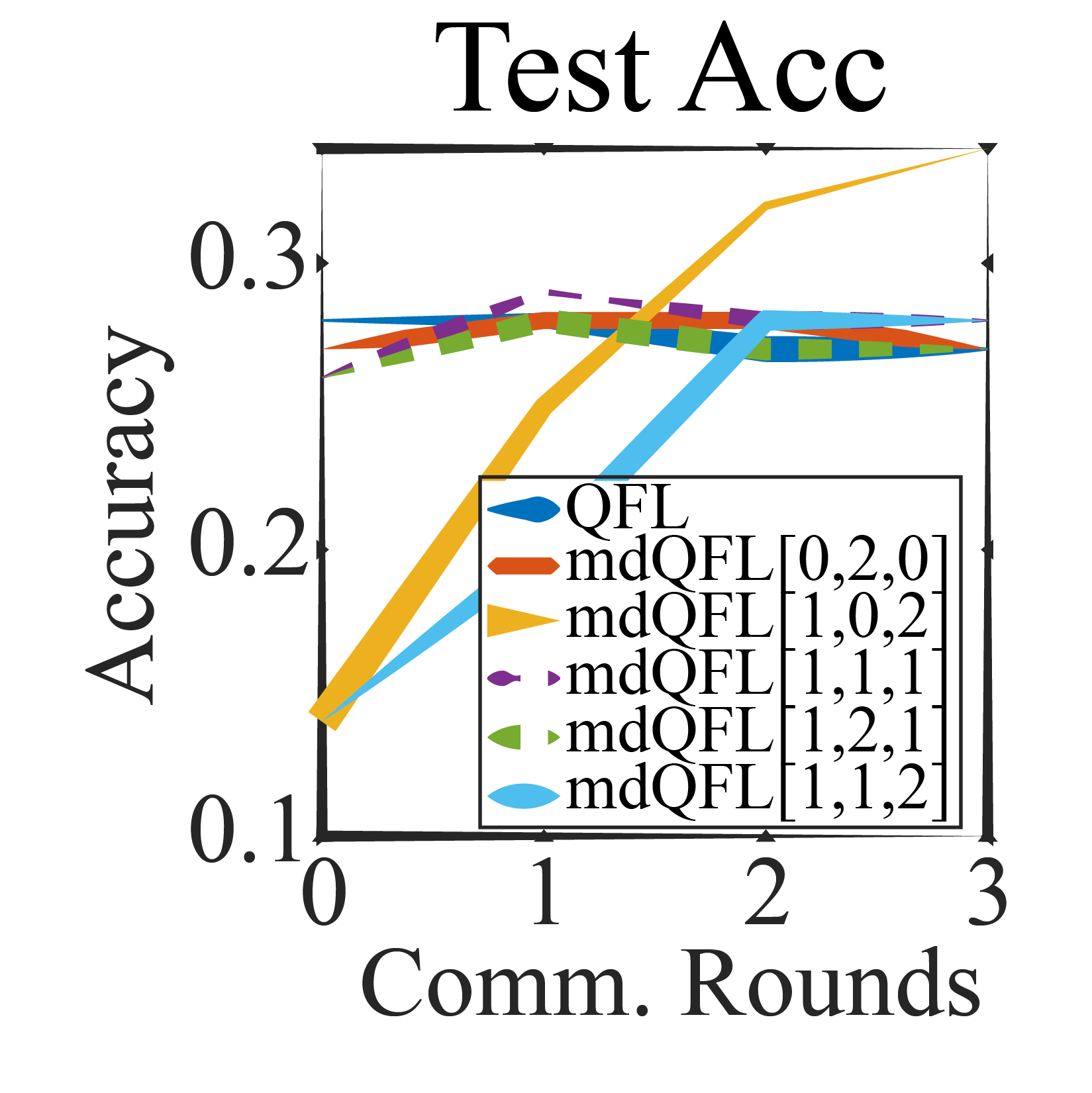}
   \caption{Small10000, n3}
   \label{fig:test_acc_small10000_20D}
    \end{subfigure}
    \begin{subfigure}[]{0.4\columnwidth}
    \centering
    \includegraphics[width=\columnwidth]{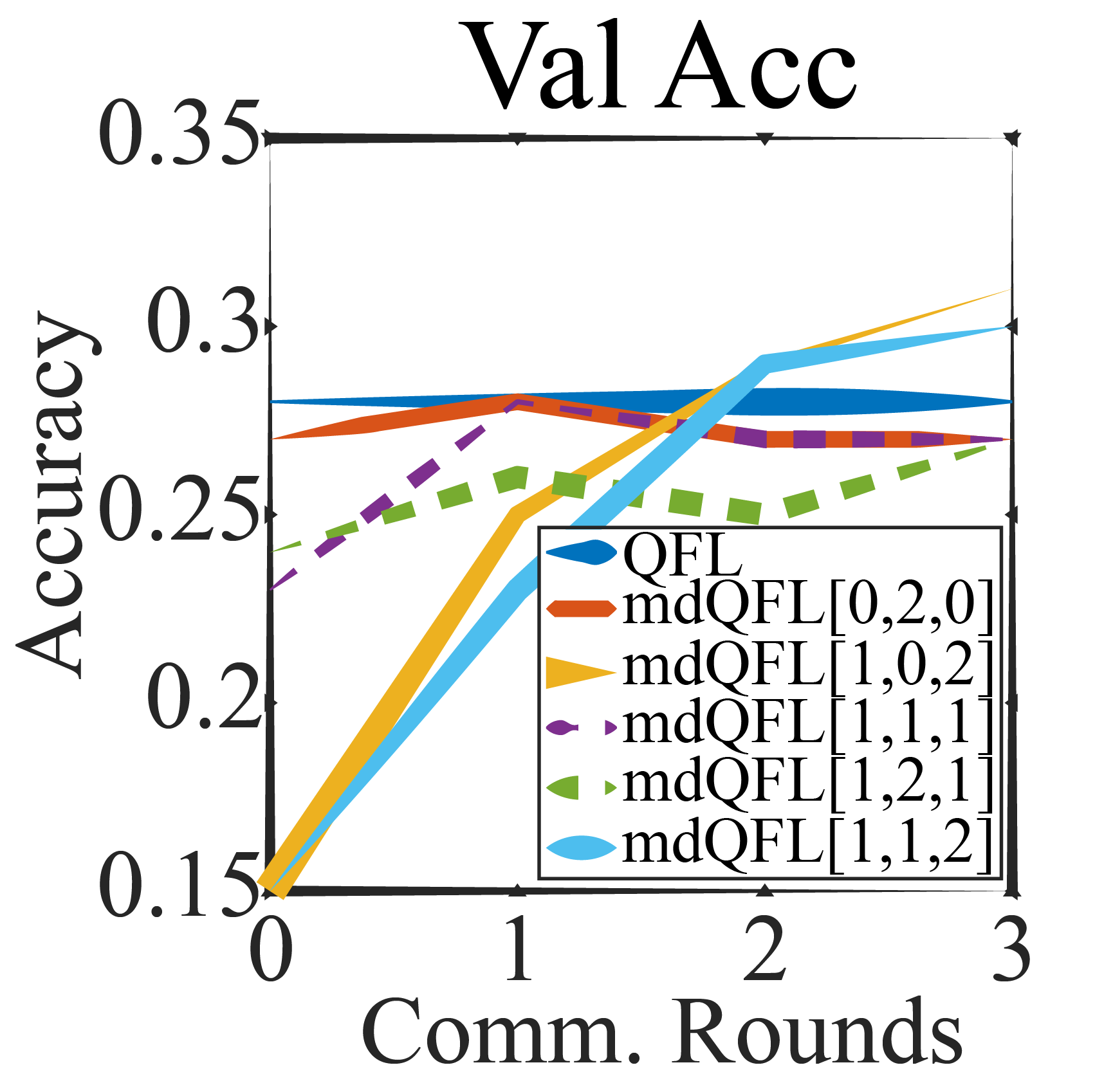}
   \caption{Small10000, n3}
   \label{fig:val_acc_small10000_20D}
    \end{subfigure}
    \caption{Global model performance: Validation and Test loss and accuracies.}
    \label{fig:server_performance_adaptive}
\end{figure}

With improvement in communication overhead, we now compare the performance comparison between varying mdQFL and QFL.
From Figure \ref{fig:server_performance_adaptive}, we can see how the global model is performing.
In Figure \ref{fig:val_loss_small2000_10iter_20d_nclass3}, mdQFL[1,1,2] and mdQFL[1,0,2] achieve the best performance. 
The mdQFL[1,1,2] configuration mainly employs cluster and local models, while mdQFL [1,0,2] focuses mainly on cluster models. 
In contrast, mdQFL[0,2,0] emphasizes the use of global models, which enhances generalization. Similarly, mdQFL[1,1,1] and mdQFL[1,2,1] adopt hybrid strategies, balancing personalization and generalization.
In Figure \ref{fig:val_loss_small2000_10iter_20d_nclass3}, using the small2000 dataset, 20 devices each performing 10 local iterations, and nclass set to 3, the higher levels of personalization outperform both QFL and greater generalization. However, mid-range levels of both personalization and generalization appear to experience a decline in performance. 

Whereas in Figure \ref{fig:val_loss_small1000_10iter_50d_nclass5}, with 50 devices, nclass=5, 10 local iteration and small1000 dataset, QFL and mdQFL[0,2,0] shows worst validation loss whereas rest of the approaches perform better than both. 
This results shows that adaptation to the degree of personalization and generation in fact impacts the system and in better ways in this experimental set, as shown in the Figures \ref{fig:val_loss_small2000_10iter_20d_nclass3} and \ref{fig:val_loss_small1000_10iter_50d_nclass5}.

Similarly, in terms of validation and test accuracy, both mdQFL[1,0,2] and mdQFl[1,1,2] perform better than QFL in both scenarios, as seen in Figures \ref{fig:test_acc_small10000_20D} and \ref{fig:val_acc_small10000_20D}. Also, from the results, we can conclude that adaptations are required based on degree of non-IID scenario, as we can see varying results with different number of nclass.

\subsubsection{Impact of personalization}
While in terms of device performance, we can see that the higher the degree of personalization, the better the consistent performance, as seen with mdQFL[1,1,2], mdQFL[1,0,2] and mdQFL[1,1,1].
The higher degree of generalization represented by mdQFL[0,2,0] shows a degradation of the performance in both the average train and the test accuracy of all devices in each communication round. 
This result confirms the need and importance of an adaptive model dependent on the varying scenario.
\begin{figure}[!htb]
    \centering
    \begin{subfigure}[]{0.4\columnwidth}
    \centering
    \includegraphics[width=\columnwidth]{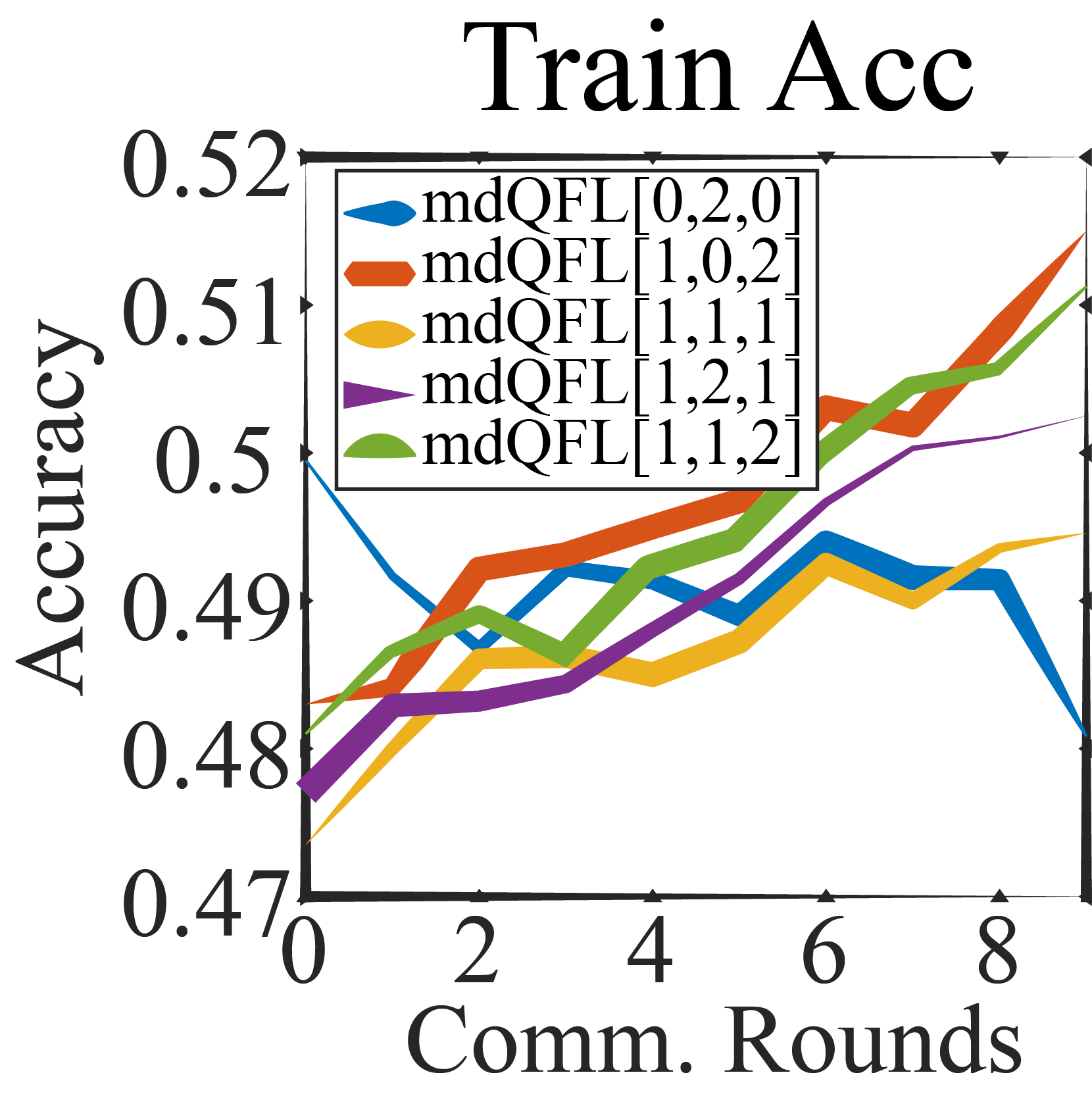}
   \caption{Normal, n2}
   \label{fig:v}
    \end{subfigure}
    \begin{subfigure}[]{0.4\columnwidth}
    \centering
    \includegraphics[width=\columnwidth]{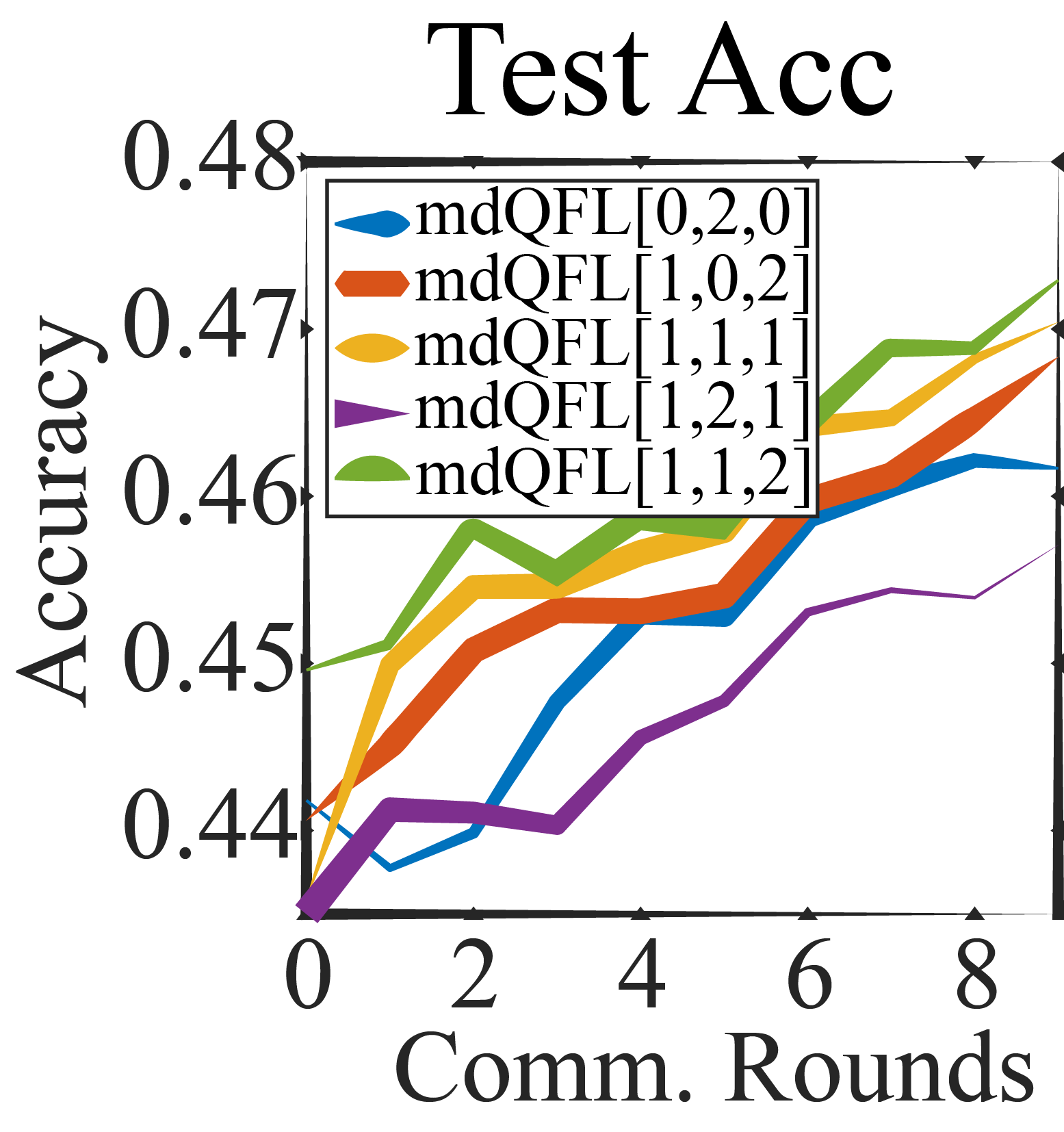}
   \caption{Small 10000, n3}
   \label{fig:adaptive_comparison_test_acc}
    \end{subfigure}
    \caption{Average train and test accuracy of all devices with variations of mdQFL.}
    \label{fig:impact_of_personalization}
\end{figure}

\subsection{Experiment 3: Impact of Various Factors}
\subsubsection{Impact of number of devices, cluster method, local iteration}
\begin{figure}[!htb]
    \centering
    \begin{subfigure}[]{0.38\columnwidth}
    \centering
    \includegraphics[width=\columnwidth]{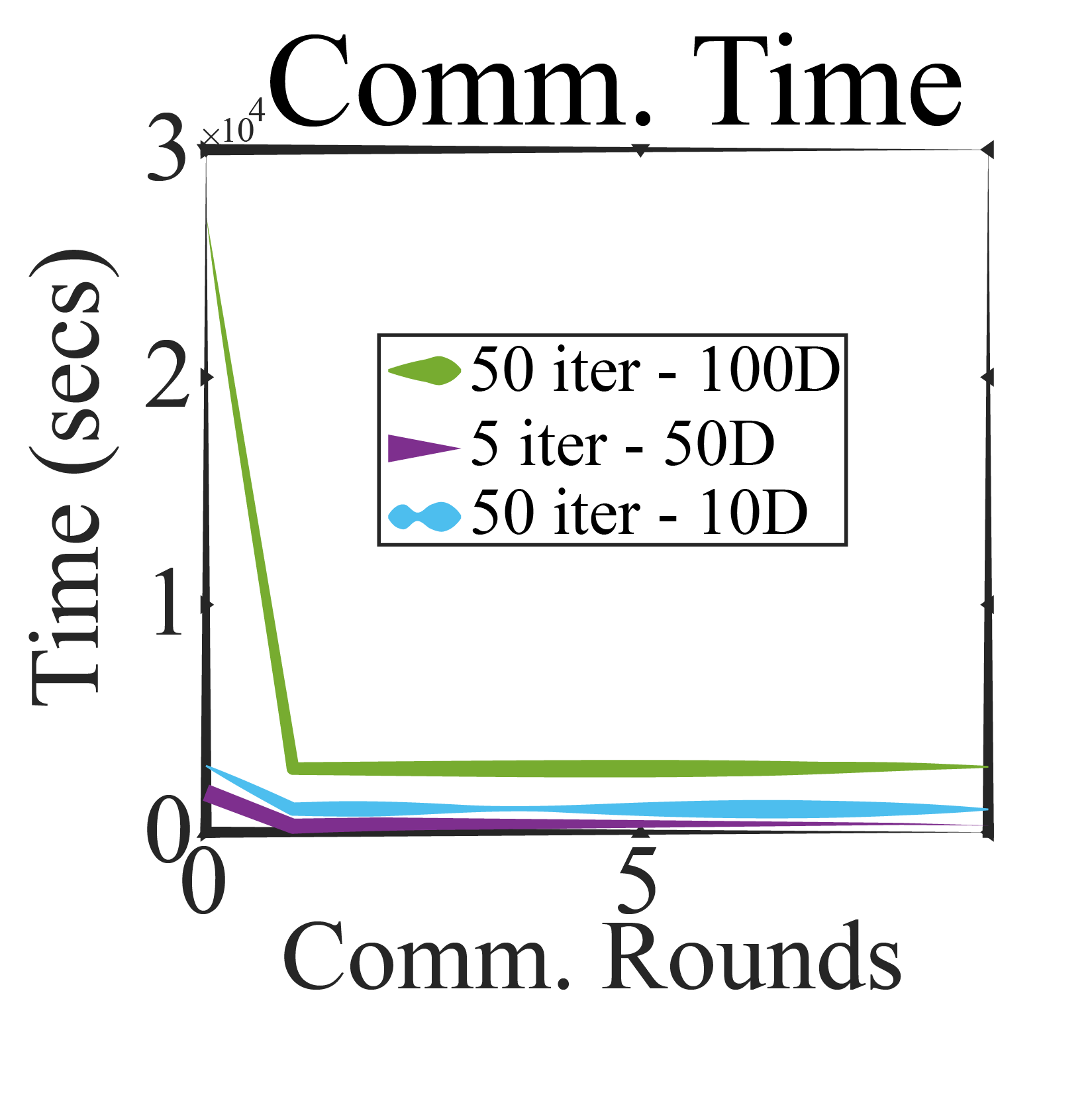}
   \caption{Number of Devices}
   \label{fig:comm_time_ndevices)}
    \end{subfigure}
    \begin{subfigure}[]{0.38\columnwidth}
    \centering
    \includegraphics[width=\columnwidth]{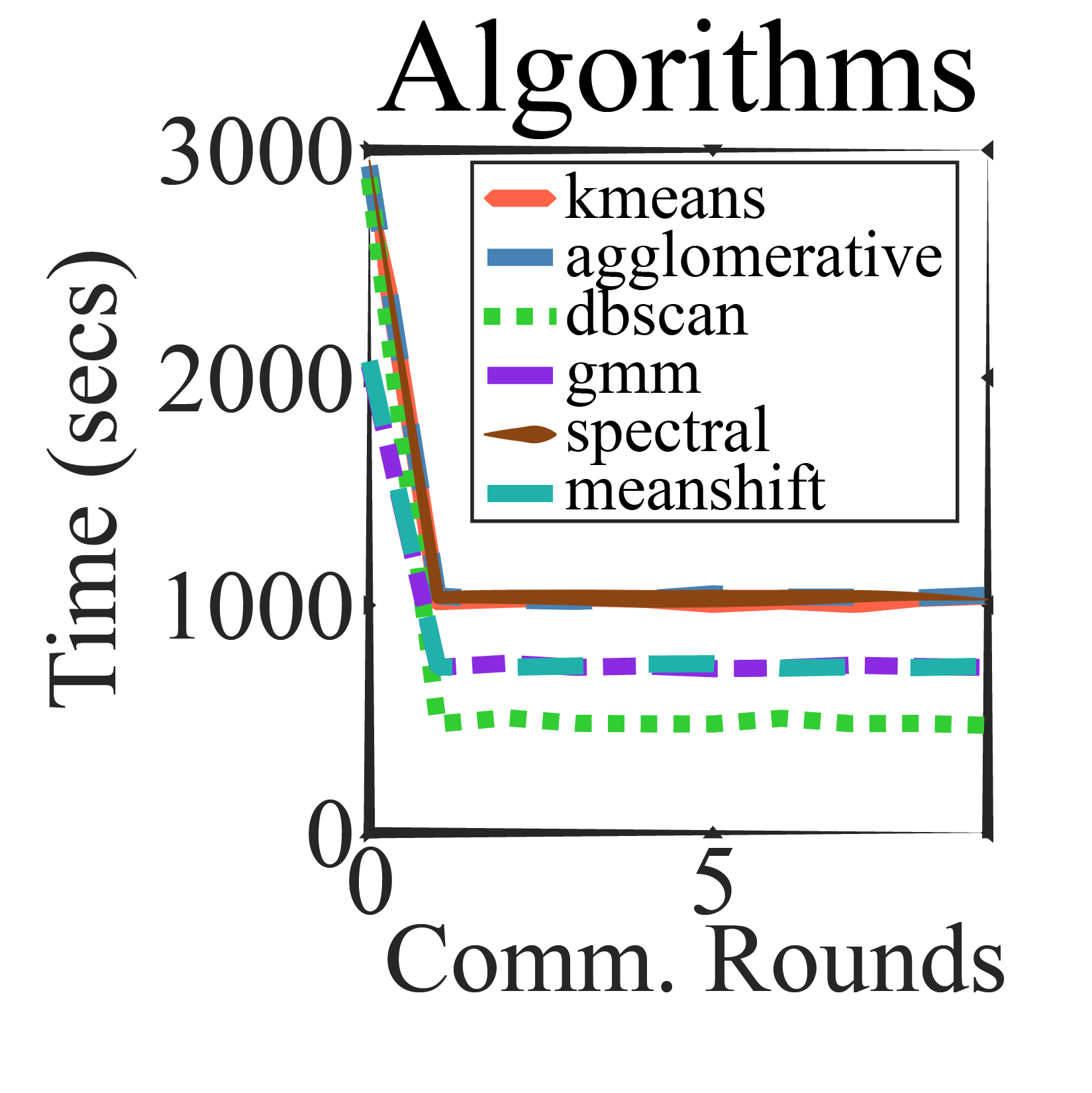}
   \caption{Clustering}
   \label{fig:comm_time_cluster)}
    \end{subfigure}
    \caption{Impact of local iteration and number of devices.}
    \label{fig:impact_of_number_devices}
\end{figure}

From Figure \ref{fig:impact_of_number_devices}, we see how different factors such as the number of devices, the cluster method and iteration within each device affect the overall performance of the system.
In Figure \ref{fig:comm_time_ndevices)}, we see how the overall communication overhead is impacted by the number of devices and the number of local iterations performed by each device.
With a total number of $50$ devices, each performing only $5$ iterations, the overall process finished faster compared to the
others.
Although keeping the local iterations at $50$ local iterations, with $100$ devices, the system performs slower than with $10$ devices.

Similarly, the method chosen for clustering clearly affects overall performance as shown in Figure \ref{fig:comm_time_cluster)}.
We conclude that DBSCAN is the top-performing clustering algorithm, followed closely by Mean-Shift and Agglomerative methods in terms of communication delay. 
This underscores the necessity of choosing the most suitable algorithm, as not all clustering techniques produce the same results.

\subsubsection{Impact of optimizer}
In terms of optimizers, we experimented with four optimizers, namely the Constrained Optimization by Linear Approximation Optimizer (COBYLA), GradientDescent (GD), Adaptive Moment Estimation (ADAM) and Analytic Quantum Gradient Descent (AQGD).
With same maxiter=100 for all optimizers with default values for their respective optimizers, our preliminary results show that in terms of training time COBYLA is the best, whereas as GD suffers the most.
Similarly, in terms of training and test accuracy, COBYLA and AQGD optimizers seem to perform similarly which is better than ADAM whereas SG is the worst optimizer. This shows the importance of choosing the best optimizer. 

\begin{figure}[!htb]
    \centering
    \begin{subfigure}[]{0.33\columnwidth}
    \centering
    \includegraphics[width=\columnwidth]{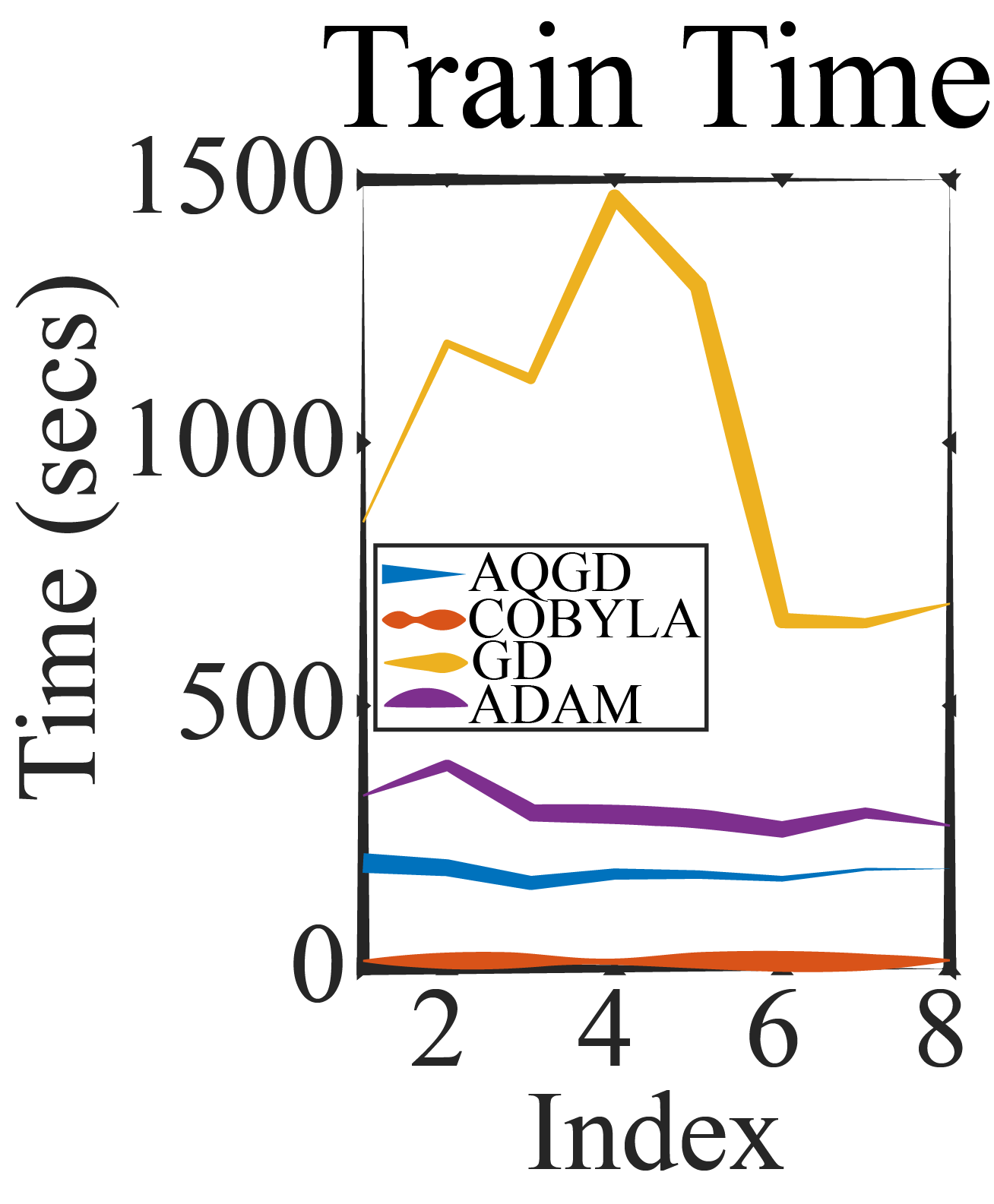}
   \caption{Training Time}
   \label{fig:training_time_optimizer)}
    \end{subfigure}
    \begin{subfigure}[]{0.3\columnwidth}
    \centering
    \includegraphics[width=\columnwidth]{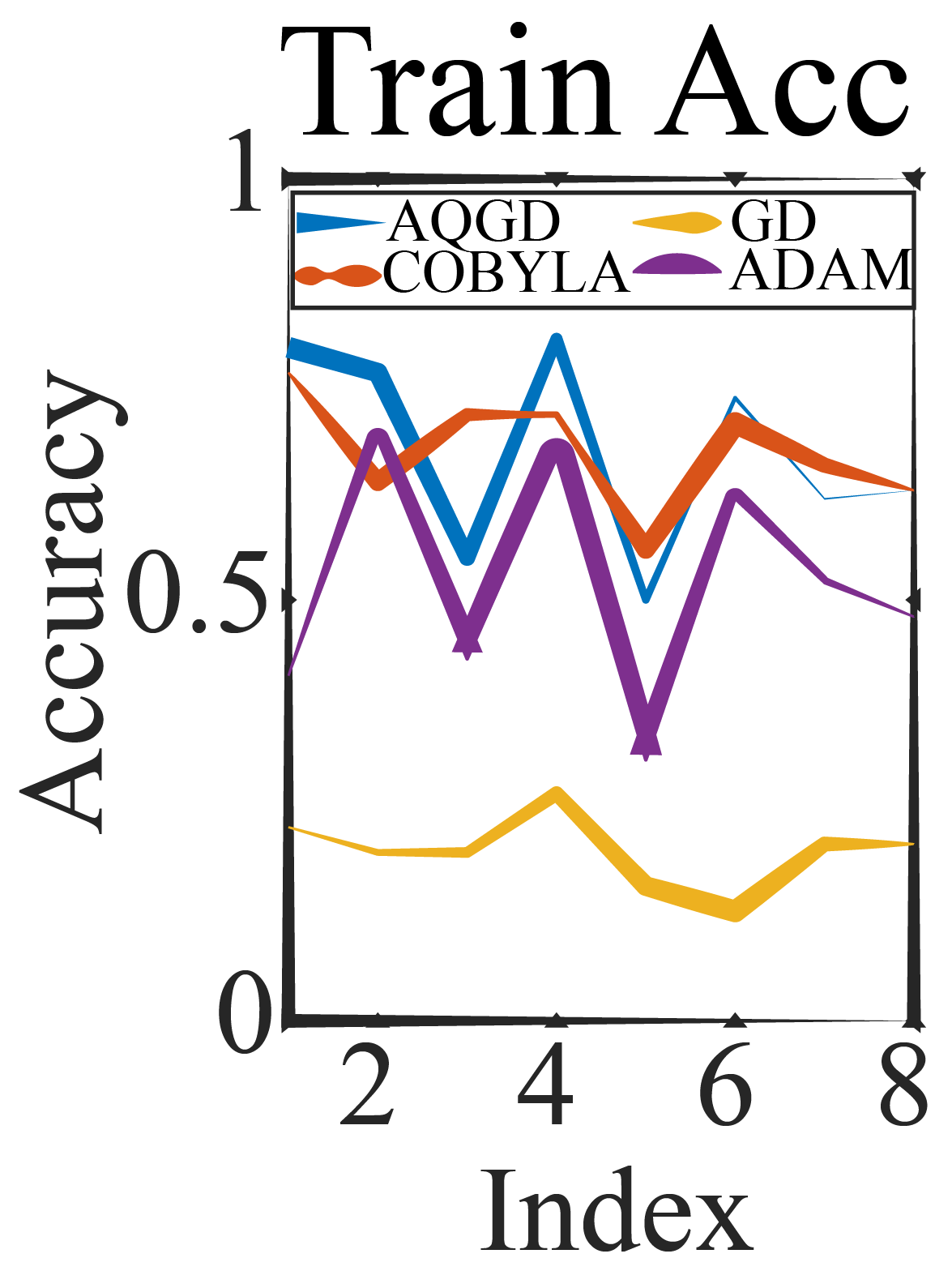}
   \caption{Train Accuracy}
   \label{fig:train_acc_optimizer)}
    \end{subfigure}
    \begin{subfigure}[]{0.3\columnwidth}
    \centering
    \includegraphics[width=\columnwidth]{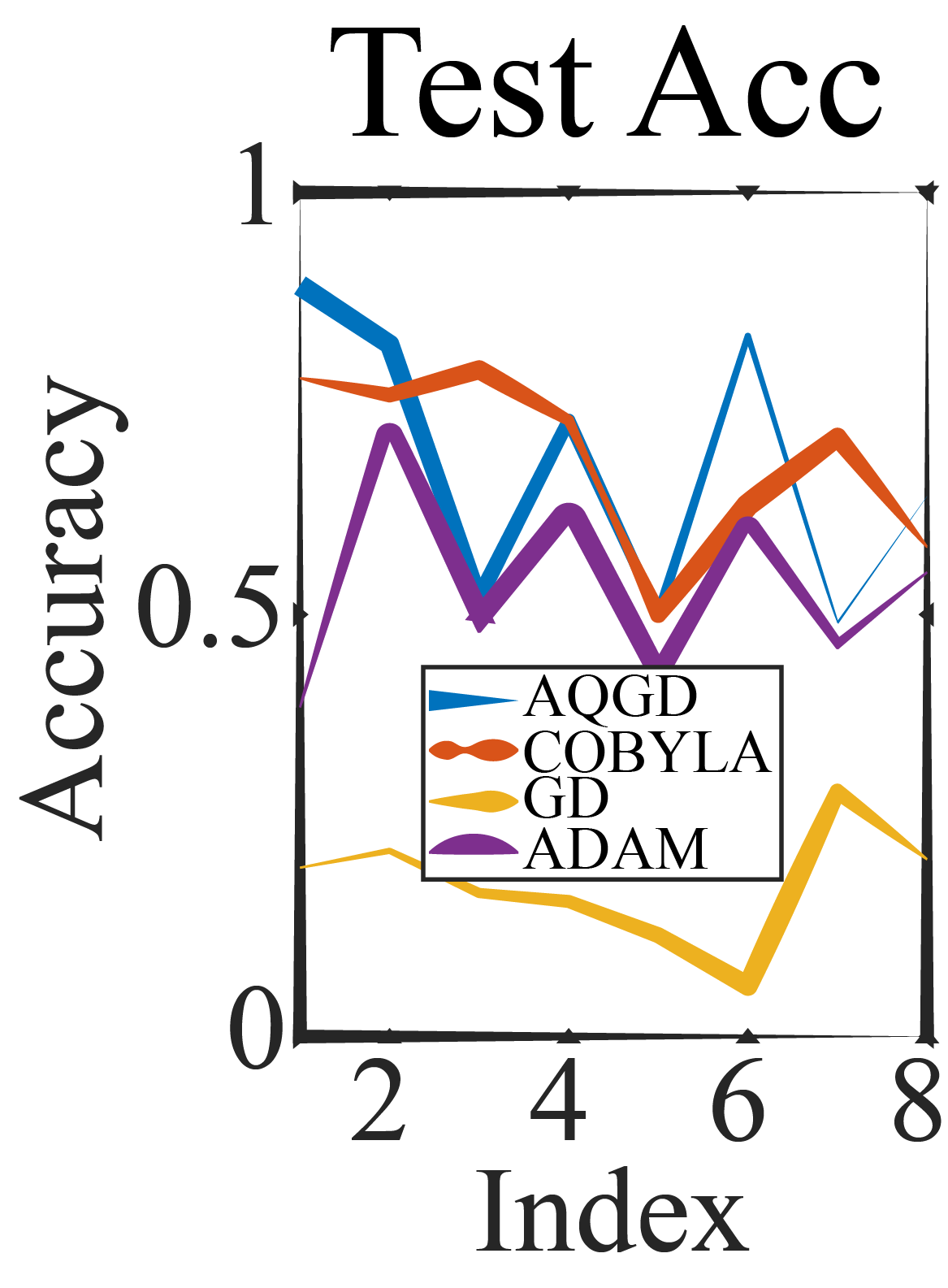}
   \caption{Test Accuracy}
   \label{fig:test_acc_optimizer)}
    \end{subfigure}
    \caption{Impact of optimizer used on local device training time and average accuracies.}
    \label{fig:impact_of_optimizers}
\end{figure}

\subsubsection{Impact of Device Selection Approach}
\begin{figure}[!htb]
    \centering
    \begin{subfigure}[]{0.4\columnwidth}
    \centering
    \includegraphics[width=\columnwidth]{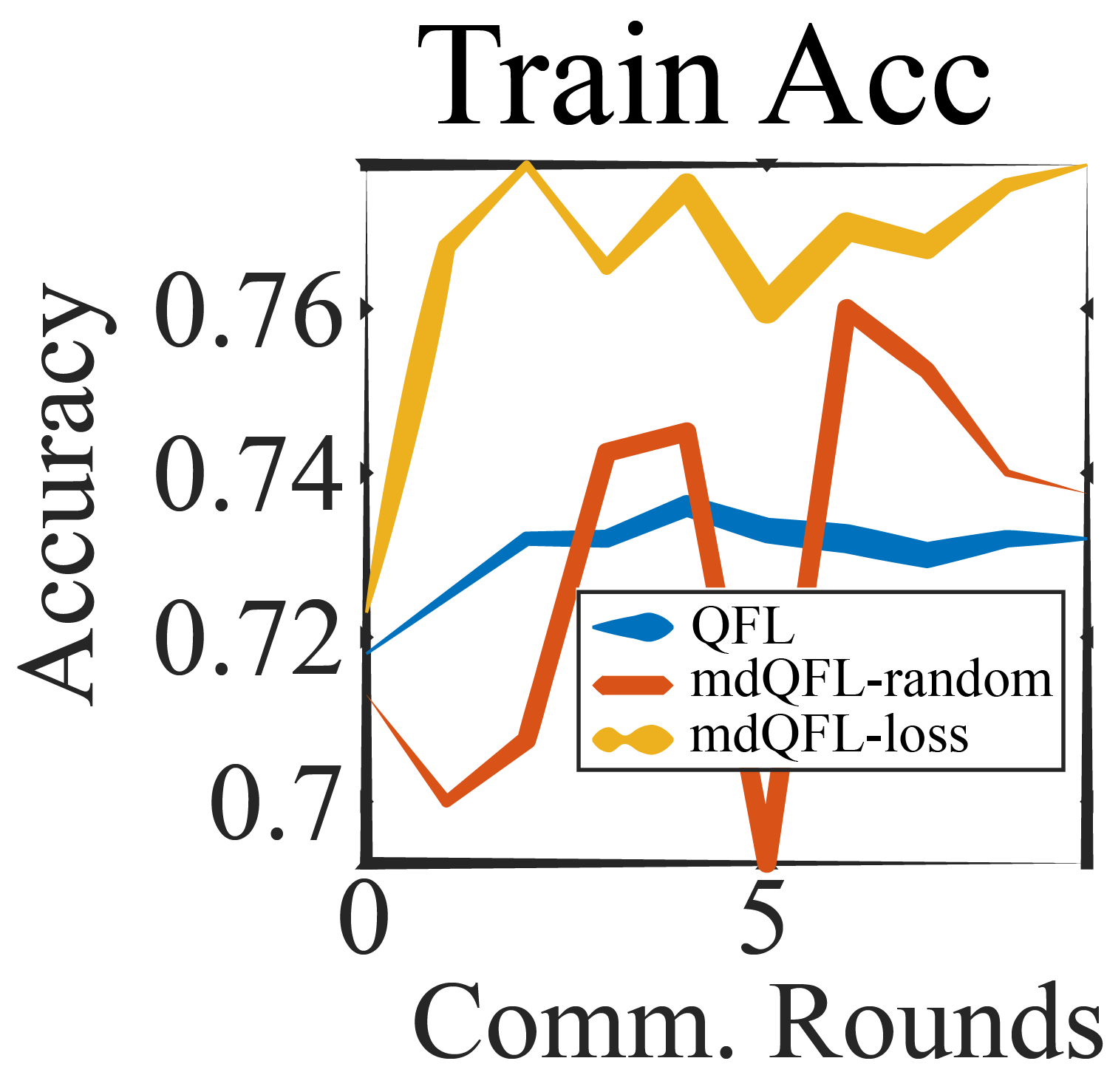}
   \caption{Train Accuracy}
   \label{fig:average_train_acc_selection)}
    \end{subfigure}
    \begin{subfigure}[]{0.4\columnwidth}
    \centering
    \includegraphics[width=\columnwidth]{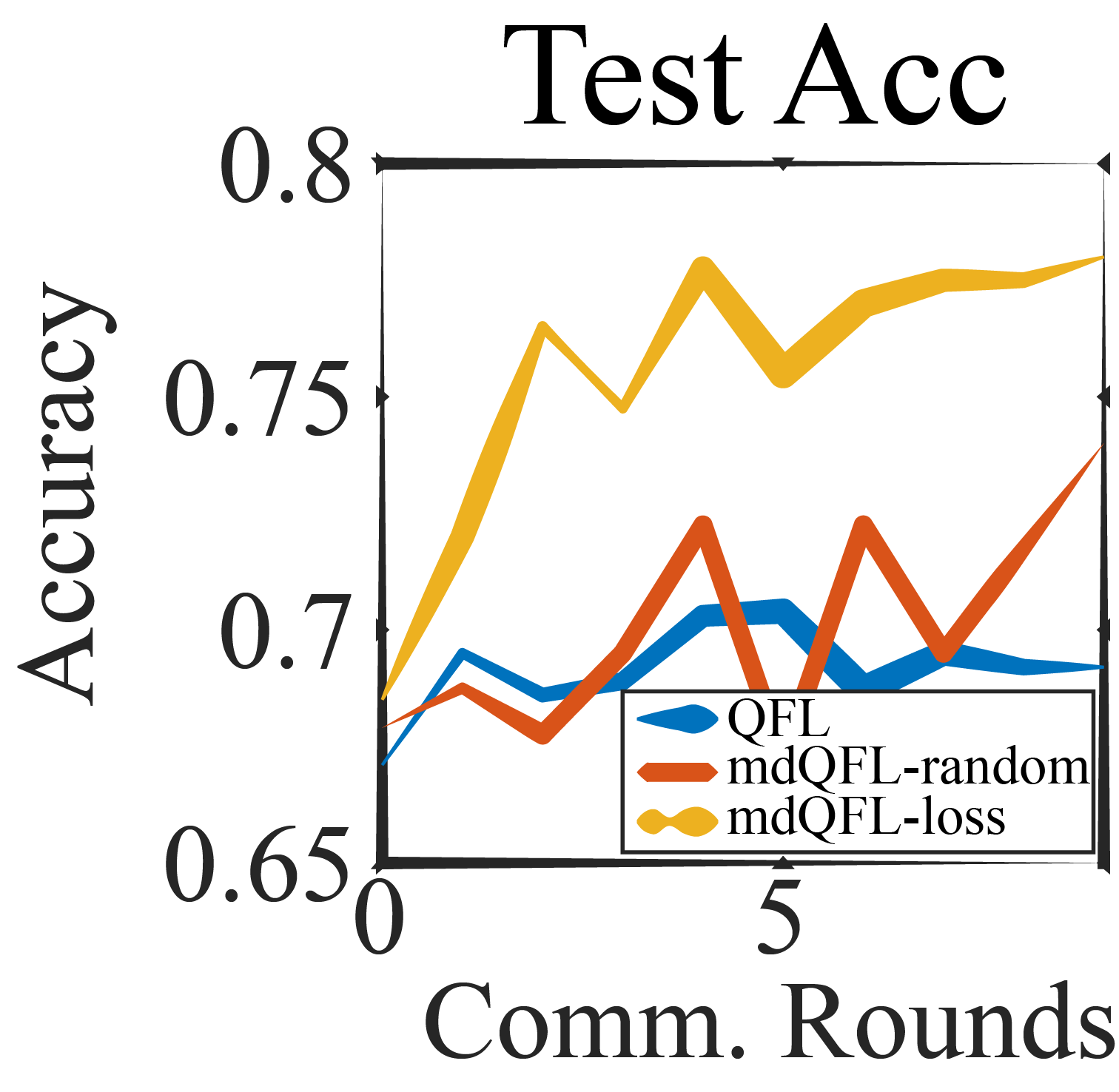}
   \caption{Test Accuracy}
   \label{fig:average_test_acc_selection)}
    \end{subfigure}
    \begin{subfigure}[]{0.4\columnwidth}
    \centering
    \includegraphics[width=\columnwidth]{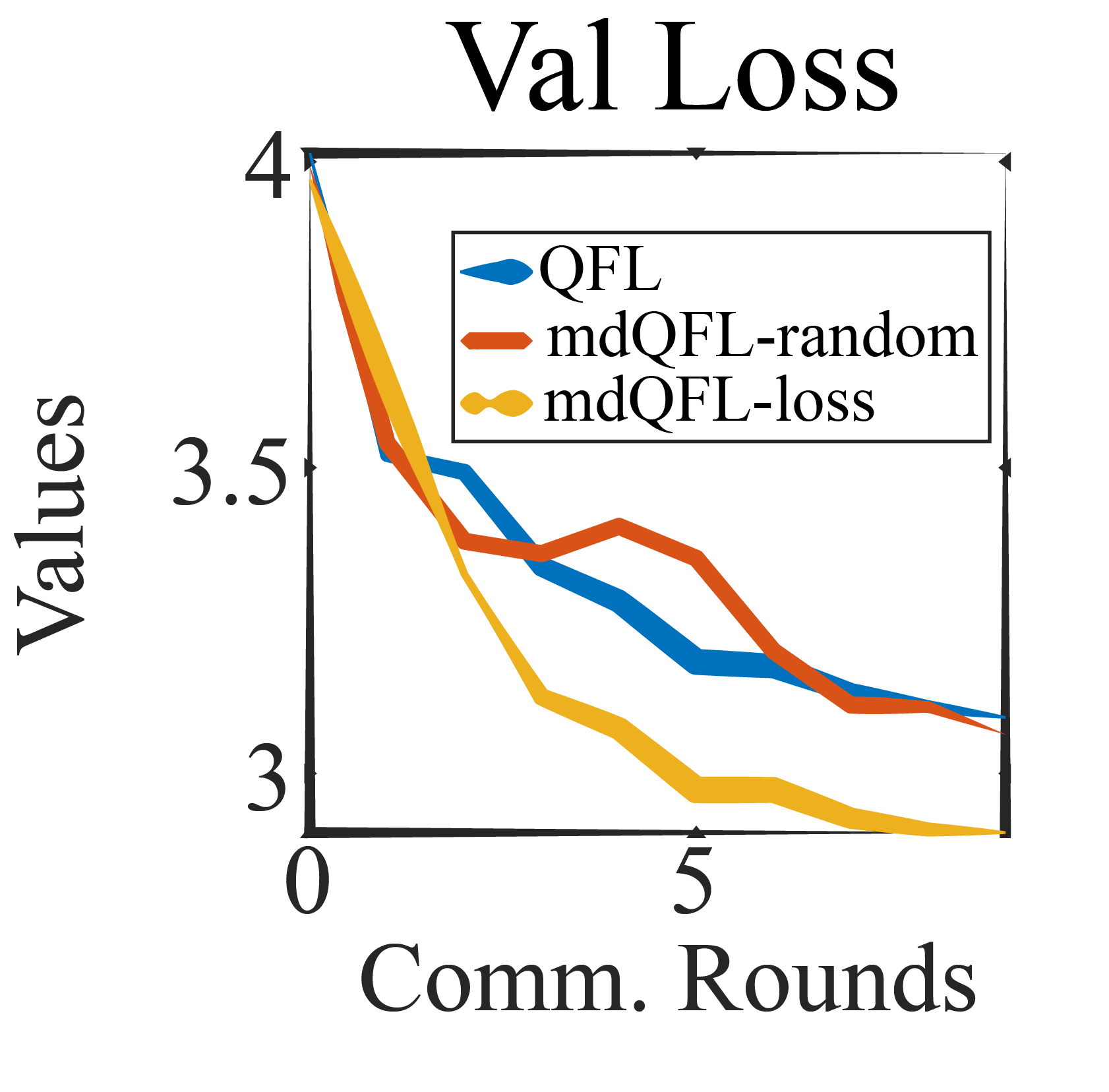}
   \caption{Validation Loss}
   \label{fig:val_loss_selection)}
    \end{subfigure}
     \begin{subfigure}[]{0.4\columnwidth}
    \centering
    \includegraphics[width=\columnwidth]{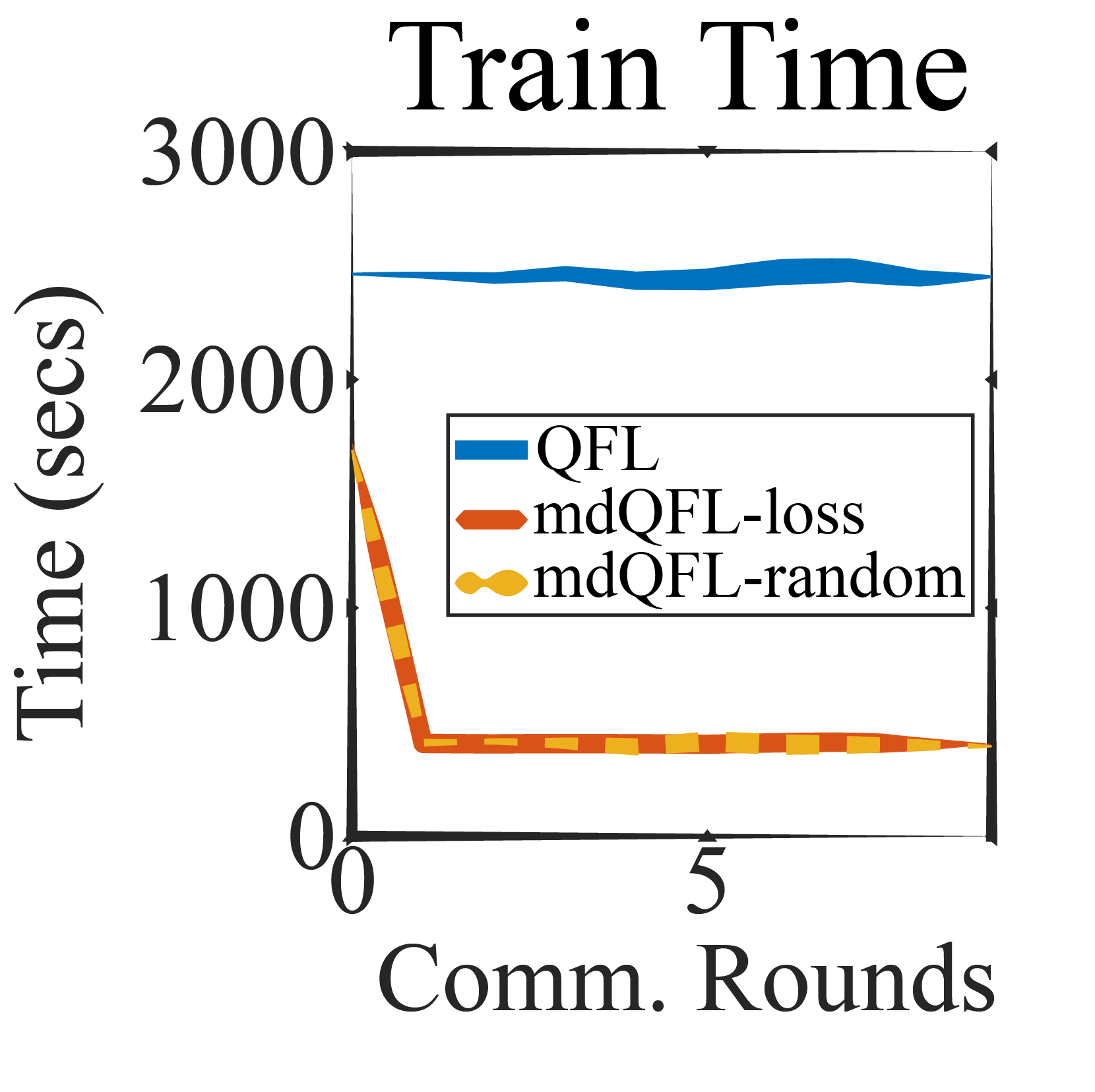}
   \caption{Training Time}
   \label{fig:training_time_selection)}
    \end{subfigure}
    \caption{Impact of device selection method (probabilistic: mdQFL-random or  gradient based: mdQFL-loss) used on training time, server validation loss and local average device accuracies.}
    \label{fig:impact_of_selection_method}
\end{figure}

The method chosen to select a device in the cluster significantly affects the overall performance of the framework, as illustrated by the average training and testing accuracy of all devices and the validation loss of the global model on the unseen data set of a server in Figure \ref{fig:impact_of_selection_method}.
We can clearly see a huge improvement in all cases with device selection based on optimizer loss values (mdQFL-loss).
In mdQFL-loss, the devices are selected from clusters with best performing objective function values, whereas mdQFL-random follows probabilistic approach to select a device from cluster.
This indicates the importance of the device selection approach in improving the overall performance of the system.
The communication overhead is similar to both approaches of mdQFL.
However, both mdQFL approaches work better than the standard QFL algorithm, highlighting the robustness of our proposed algorithm, both with a probabilistic or heuristic approach to the selection of devices.
These sets of comparative experiments were carried out with a small set of MNIST (1000 samples) for training among 20 devices and 100 samples for the test set for the server device.
The maximum iteration for the COBYLA optimizer was set to 100 with nclass = 3 labels distributed between devices, and the device grouping mechanism used DBSCAN.

\section{Conclusion}
In this work, we introduced a communication-efficient adaptive algorithm for QFL (mdQFL) designed to tackle communication overhead challenges when training with numerous devices, large datasets, and non-IID scenarios.
Our experimental evaluation indicates an almost 50\% improvement in communication efficiency, along with consistent improvements in local performance while maintaining stable global convergence.
Our framework integrates adaptive training and update personalization, alongside test generalization, effectively functioning with varying numbers of devices and accommodating different levels of personalization and generalization.
Both experimental and theoretical analyses validate that our framework is robust and efficient in managing large datasets, non-IID settings, and a significant number of devices. As a result, by improving both communication overhead and performance, we have demonstrated the feasibility and practicality of the framework.
% \bibliography{yourbib}

\begin{thebibliography}{10}
\providecommand{\url}[1]{#1}
\csname url@samestyle\endcsname
\providecommand{\newblock}{\relax}
\providecommand{\bibinfo}[2]{#2}
\providecommand{\BIBentrySTDinterwordspacing}{\spaceskip=0pt\relax}
\providecommand{\BIBentryALTinterwordstretchfactor}{4}
\providecommand{\BIBentryALTinterwordspacing}{\spaceskip=\fontdimen2\font plus
\BIBentryALTinterwordstretchfactor\fontdimen3\font minus \fontdimen4\font\relax}
\providecommand{\BIBforeignlanguage}[2]{{%
\expandafter\ifx\csname l@#1\endcsname\relax
\typeout{** WARNING: IEEEtran.bst: No hyphenation pattern has been}%
\typeout{** loaded for the language `#1'. Using the pattern for}%
\typeout{** the default language instead.}%
\else
\language=\csname l@#1\endcsname
\fi
#2}}
\providecommand{\BIBdecl}{\relax}
\BIBdecl

\bibitem{mcmahanCommunicationEfficientLearningDeep2023}
H.~B. McMahan, E.~Moore, D.~Ramage, S.~Hampson, and B.~A. y~Arcas, ``Communication-{{Efficient Learning}} of {{Deep Networks}} from {{Decentralized Data}},'' Jan. 2023.

\bibitem{dinhFederatedLearningWireless2021}
\BIBentryALTinterwordspacing
C.~T. Dinh, N.~H. Tran, M.~N.~H. Nguyen, C.~S. Hong, W.~Bao, A.~Y. Zomaya, and V.~Gramoli, ``Federated {Learning} {Over} {Wireless} {Networks}: {Convergence} {Analysis} and {Resource} {Allocation},'' \emph{IEEE/ACM Transactions on Networking}, vol.~29, no.~1, pp. 398--409, Feb. 2021, conference Name: IEEE/ACM Transactions on Networking. [Online]. Available: \url{https://ieeexplore.ieee.org/document/9261995}
\BIBentrySTDinterwordspacing

\bibitem{liFederatedLearningChallenges2020}
T.~Li, A.~K. Sahu, A.~Talwalkar, and V.~Smith, ``Federated learning: {{Challenges}}, methods, and future directions,'' \emph{IEEE signal processing magazine}, vol.~37, no.~3, pp. 50--60, 2020.

\bibitem{knillQuantumComputing2010}
E.~Knill, ``Quantum computing,'' \emph{Nature}, vol. 463, no. 7280, pp. 441--443, Jan. 2010.

\bibitem{fedProx_2020}
T.~Li, A.~K. Sahu, M.~Zaheer, M.~Sanjabi, A.~Talwalkar, and V.~Smith, ``Federated {{Optimization}} in {{Heterogeneous Networks}},'' Apr. 2020.

\bibitem{fedBN_ICLR2021}
X.~Li, M.~Jiang, X.~Zhang, M.~Kamp, and Q.~Dou, ``{{FEDBN}}: {{FEDERATED LEARNING ON NON-IID FEATURES VIA LOCAL BATCH NORMALIZATION}},'' \emph{ICLR 2021}, 2021.

\bibitem{FedDCFederatedLearning2022}
L.~Gao, H.~Fu, L.~Li, Y.~Chen, M.~Xu, and C.-Z. Xu, ``{{FedDC}}: {{Federated Learning}} with {{Non-IID Data}} via {{Local Drift Decoupling}} and {{Correction}},'' in \emph{2022 {{IEEE}}/{{CVF Conference}} on {{Computer Vision}} and {{Pattern Recognition}} ({{CVPR}})}.\hskip 1em plus 0.5em minus 0.4em\relax {New Orleans, LA, USA}: {IEEE}, Jun. 2022, pp. 10\,102--10\,111.

\bibitem{daiDisPFLCommunicationEfficientPersonalized2022}
R.~Dai, L.~Shen, F.~He, X.~Tian, and D.~Tao, ``{{DisPFL}}: {{Towards Communication-Efficient Personalized Federated Learning}} via {{Decentralized Sparse Training}},'' in \emph{Proceedings of the 39th {{International Conference}} on {{Machine Learning}}}.\hskip 1em plus 0.5em minus 0.4em\relax {PMLR}, Jun. 2022, pp. 4587--4604.

\bibitem{zhaoExactDecompositionQuantum2022a}
H.~Zhao, ``Exact {{Decomposition}} of {{Quantum Channels}} for {{Non-IID Quantum Federated Learning}},'' Sep. 2022.

\bibitem{xiaQuantumFedFederatedLearning2021a}
Q.~Xia and Q.~Li, ``{{QuantumFed}}: {{A Federated Learning Framework}} for {{Collaborative Quantum Training}},'' in \emph{2021 {{IEEE Global Communications Conference}} ({{GLOBECOM}})}, Dec. 2021, pp. 1--6.

\bibitem{chehimiQuantumFederatedLearning2022}
M.~Chehimi and W.~Saad, ``Quantum {{Federated Learning}} with {{Quantum Data}},'' in \emph{{{ICASSP}} 2022 - 2022 {{IEEE International Conference}} on {{Acoustics}}, {{Speech}} and {{Signal Processing}} ({{ICASSP}})}, May 2022, pp. 8617--8621.

\bibitem{huangQuantumFederatedLearning2022}
R.~Huang, X.~Tan, and Q.~Xu, ``Quantum {{Federated Learning With Decentralized Data}},'' \emph{IEEE Journal of Selected Topics in Quantum Electronics}, vol.~28, no. 4: Mach. Learn. in Photon. Commun. and Meas. Syst., pp. 1--10, Jul. 2022.

\bibitem{chenFederatedQuantumMachine2021}
S.~Y.-C. Chen and S.~Yoo, ``Federated {{Quantum Machine Learning}},'' \emph{Entropy}, vol.~23, no.~4, p. 460, Apr. 2021.

\bibitem{zhangFederatedLearningQuantum2022}
Y.~Zhang, C.~Zhang, C.~Zhang, L.~Fan, B.~Zeng, and Q.~Yang, ``Federated {{Learning}} with {{Quantum Secure Aggregation}},'' Jul. 2022.

\bibitem{liDataHeterogeneityRobustFederated2022}
Z.~Li, Y.~He, H.~Yu, J.~Kang, X.~Li, Z.~Xu, and D.~Niyato, ``Data {{Heterogeneity-Robust Federated Learning}} via {{Group Client Selection}} in {{Industrial IoT}},'' Feb. 2022.

\bibitem{ghoshEfficientFrameworkClustered2020a}
A.~Ghosh, J.~Chung, D.~Yin, and K.~Ramchandran, ``An {{Efficient Framework}} for {{Clustered Federated Learning}},'' in \emph{Advances in {{Neural Information Processing Systems}}}, vol.~33.\hskip 1em plus 0.5em minus 0.4em\relax Curran Associates, Inc., 2020, pp. 19\,586--19\,597.

\bibitem{liConvergenceFedAvgNonIID2019}
X.~Li, K.~Huang, W.~Yang, S.~Wang, and Z.~Zhang, ``On the {{Convergence}} of {{FedAvg}} on {{Non-IID Data}},'' https://arxiv.org/abs/1907.02189v4, Jul. 2019.

\bibitem{Powell1994}
\BIBentryALTinterwordspacing
M.~J.~D. Powell, ``A direct search optimization method that models the objective and constraint functions by linear interpolation,'' \emph{Springer Netherlands,}, pp. 51--67, 1994. [Online]. Available: \url{https://doi.org/10.1007/978-94-015-8330-5_4}
\BIBentrySTDinterwordspacing

\end{thebibliography}
% \bibliographystyle{IEEEtran}

\section{APPENDIX} \label{sec:appendix}
\subsection{MNIST Dataset preparation}
\begin{figure}[!htbh]
    \centering
    \includegraphics[width=0.8\linewidth]{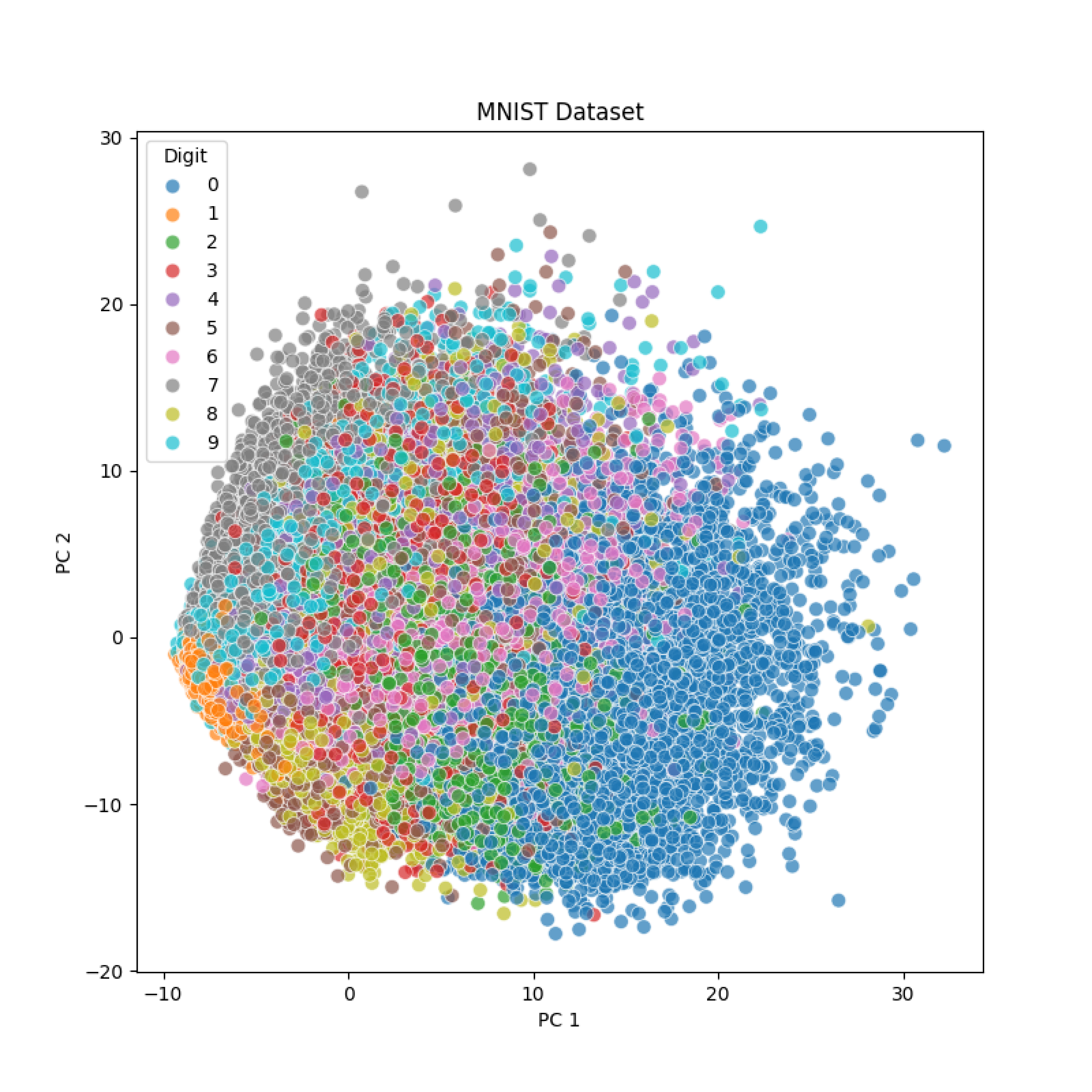}
    \caption{MNIST dataset after PCA}
    \label{fig:mnist_pca}
\end{figure}

Principal Component Analysis (PCA) was used to reduce dimensionality in the MNIST dataset and others in this study. After loading the original data set, the images were flattened from 28x28 to 784 after which normalization was done on it using Standard Scaler.
After that, PCA was used with n\_components=4. This was done to use 4 qubits in our quantum circuit.
The plot is shown in Figure \ref{fig:mnist_pca}. 
There is overlap between the classes, which indicates that better approaches might be needed further to obtain a clearer separation.

\subsection{Labels per device}
\begin{figure}[!htbh]
    \centering
    \includegraphics[width=0.6\linewidth]{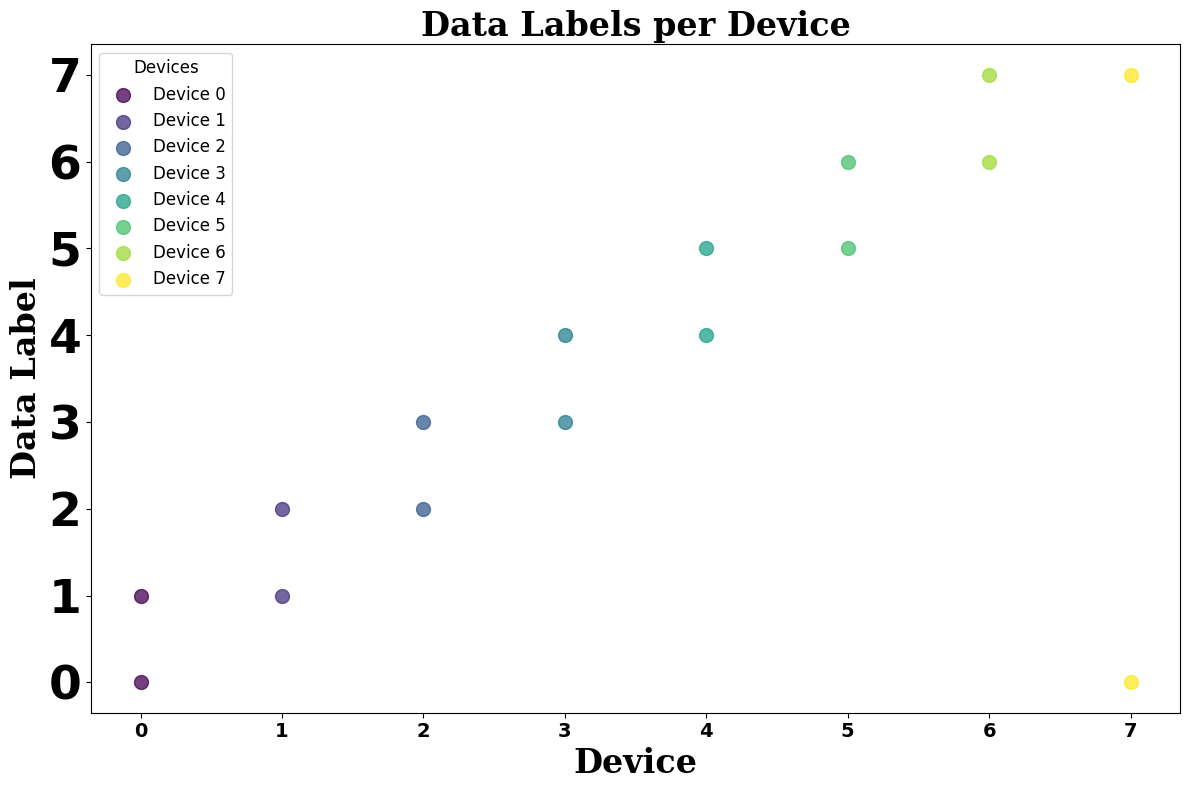}
    \caption{Label per device}
    \label{fig:label_distribution}
\end{figure}

The label distribution between devices is done as shown in Figure \ref{fig:label_distribution}.
Distributing only two unique labels per device or more can be considered to reflect varying degree of non-IID scenario in real world scenario where devices mostly have unique characteristics.

\section{Pre-Algorithm Steps}
\subsection{Data Preparation and Processing} 
\begin{algorithm}[!htbh]
\caption{Data Preparation and Processing}
\begin{algorithmic}[1]
\Require $(X_{\text{train}}, Y_{\text{train}}), (X_{\text{test}}, Y_{\text{test}})$: Raw data samples and labels
\Require $d$: Original feature dimension, $n_{\text{train}}, n_{\text{test}}$: Subset sizes, $\alpha$: Train-validation split ratio, $r$: Random state for reproducibility, $k$: Number of PCA components

\State \textbf{Step 1: Load Dataset}
% \If{$\text{data\_size} = \text{"small"}$}
\State $X_{\text{train}} \gets X_{\text{train}}[:n_{\text{train}}]$,  $Y_{\text{train}} \gets Y_{\text{train}}[:n_{\text{train}}]$
\State $X_{\text{test}} \gets X_{\text{test}}[:n_{\text{test}}]$, $Y_{\text{test}} \gets Y_{\text{test}}[:n_{\text{test}}]$
% \EndIf

\State \textbf{Step 2: Flatten Features}
\State $X_{\text{train\_flat}} \gets \text{Reshape}(X_{\text{train}}, (|X_{\text{train}}|, d))$
\State $X_{\text{test\_flat}} \gets \text{Reshape}(X_{\text{test}}, (|X_{\text{test}}|, d))$

\State \textbf{Step 3: Normalize Features}
\State $\text{Scaler} \gets \text{StandardScaler()}$
\State $X_{\text{train\_norm}} \gets \text{Scaler.fit\_transform}(X_{\text{train\_flat}})$
\State $X_{\text{test\_norm}} \gets \text{Scaler.transform}(X_{\text{test\_flat}})$

\State \textbf{Step 4: Apply PCA}
\State $\text{PCA} \gets \text{PCA}(k)$
\State $X_{\text{train\_pca}} \gets \text{PCA.fit\_transform}(X_{\text{train\_norm}})$
\State $X_{\text{test\_pca}} \gets \text{PCA.transform}(X_{\text{test\_norm}})$

\State \textbf{Step 5: Split Data for Training and Testing}
\State $(X_{\text{train}}, X_{\text{validation}}, Y_{\text{train}}, Y_{\text{validation}}) \gets \text{Split}(X_{\text{train\_pca}}, Y_{\text{train}}, \text{train\_size} = \alpha, \text{random\_state} = r)$

\end{algorithmic}
\label{alg:data_preparation}
\end{algorithm}

The Algorithm \ref{alg:data_preparation} begins by loading the dataset, separating it into training and testing sets. If a smaller subset of the data is specified, the algorithm reduces the dataset size by selecting the first $n_{\text{train}}$ samples for training and $n_{\text{test}}$ samples for testing. This subset selection helps in reducing computational costs for quick experimentation. Next, the algorithm flattens the input data, originally in a higher-dimensional form, into two-dimensional arrays where each sample is converted into a flat vector of dimension $d$. This flattening process is essential for compatibility with certain machine learning models that require a linear input structure.

After flattening, the algorithm normalizes the features of the dataset using a \textit{StandardScaler}. This step standardizes the feature values to have a mean of 0 and a standard deviation of 1, which is important to ensure that each feature contributes equally to the learning process. Following normalization, the algorithm applies Principal Component Analysis (PCA) to the data to reduce its dimensionality. By projecting the data into a $k$-dimensional space, PCA captures the most informative features while mitigating the curse of dimensionality and reducing computational complexity.

Finally, the processed dataset is split into training and validation sets based on a specified train-validation ratio $\alpha$. A random state $r$ is utilized to ensure reproducibility of the split. This final split allows model evaluation on a validation set that was not used during training, providing a robust estimate of model performance.

\subsection{Data Distribution} The Algorithm \ref{alg:data_distribution_initialization} distributes training data to multiple devices in a chain structure. For each device, a specific label range is determined using a cyclic distribution of classes. Data points falling within the label range are selected and assigned to the respective device. The algorithm appends the selected data and labels to lists for each device. Finally, we obtain the data and labels for each device, facilitating distributed training or evaluation.
Finally, devices are initialized with their respective data, id $i$, $maxiter$ for optimizer, $True$ for warm start, and $P_o$ for the initial point to start training which is initially random parameters.

\begin{algorithm}
\caption{Data Distribution using $l$-cycle}
\begin{algorithmic}[1]
\Require $N$: Number of devices, $C$: Number of classes per device, $\mathcal{I}_i$: Device indices, $s_i$, $e_i$: label start and label end
\Require $\mathbf{X}$: Training features, $\mathbf{y}$: Training labels
\State \textbf{Initialize} $\mathcal{D}_{\text{data}} \gets []$, $\mathcal{D}_{\text{labels}} \gets []$

\For{$i = 0$ \textbf{to} $N-1$}
    \State \textbf{Determine label range for device } $i$
    \State $s_i \gets i \bmod 10$
    \State $e_i \gets (i + C) \bmod 10$
    
    \If{$e_i > s_i$}
        \State $\mathcal{I}_i \gets \{ j \mid s_i \leq y_j < e_i \}$
    \Else
        \State $\mathcal{I}_i \gets \{ j \mid y_j \geq s_i \lor y_j < e_i \}$
    \EndIf

    \State \textbf{Assign data and labels:}
    \State $\mathbf{X}_i \gets \mathbf{X}[\mathcal{I}_i]$
    \State $\mathbf{y}_i \gets \mathbf{y}[\mathcal{I}_i]$
    
    \State $\mathcal{D}_{\text{data}} \gets \mathcal{D}_{\text{data}} \cup \{\mathbf{X}_i\}$
    \State $\mathcal{D}_{\text{labels}} \gets \mathcal{D}_{\text{labels}} \cup \{\mathbf{y}_i\}$
\EndFor
\State \textbf{Define Ansatz Circuit:}
\State $\mathcal{A} \gets \text{RealAmplitudes}(Q = 4, R = 3)$

\State \textbf{Set Initial Point:}
\State $\mathbf{p_0} \gets [0.5]^{|\mathcal{A}|}$

\State \textbf{Initialize Devices:}
\For{$i = 0$ \textbf{to} $N-1$}
    \State $\text{device}_i \gets \text{Device}(i, \mathcal{D}_{\text{data}}[i], \mathcal{D}_{\text{labels}}[i], \text{maxiter}, \text{True}, \mathbf{p_0})$
\EndFor
\end{algorithmic}
\label{alg:data_distribution_initialization}
\end{algorithm}

\subsection{Device Entity} The algorithm initializes a device with input data, labels, and training parameters. It normalizes the data, sets up the quantum feature map and ansatz, selects an optimizer, and performs training using a Variational Quantum Classifier (VQC), and eventually computes many metrics such as training time, performance scores etc.
\begin{algorithm}
\caption{Device Entity}
\begin{algorithmic}[1]
\Require $i$: Device Id, $\mathbf{X}$: Input data, $\mathbf{y}$: Labels, $\mathbf{p_0}$: Initial point, $\mathcal{O}$: Optimizer type, $\mathbf{X}$: Input feature matrix, $\mathbf{y}$: Label vector, $iter$: Maximum iterations, $\mathbf{p_0}$: Initial parameter vector

\State \textbf{Device Attributes:}
\State $\mathbf{X}_{\text{norm}} \gets \text{MinMaxScaler}(\mathbf{X})$
\State $\mathbf{w} \gets \emptyset$ \Comment{Cluster ID}
\State $\tau \gets 0$ \Comment{Training time}
\State $q_{\text{train}} \gets 0$, $q_{\text{test}} \gets 0$ \Comment{Training and test scores}
\State $\mathcal{S} \gets \text{Sampler}()$
\State $\mathcal{F} \gets \text{ZZFeatureMap}(n_f, 1)$ \Comment{$n_f$: Number of features}
\State $\mathcal{A} \gets \text{RealAmplitudes}(n_f, 3)$

\If{$\mathcal{O} = \text{sgd}$}
    \State $\mathcal{O} \gets \text{GradientDescent}(iter)$
\ElsIf{$\mathcal{O} = \text{cobyla}$}
    \State $\mathcal{O} \gets \text{COBYLA}(iter)$
\EndIf

\State $\mathbf{X}_{\text{train}}, \mathbf{X}_{\text{test}}, \mathbf{y}_{\text{train}}, \mathbf{y}_{\text{test}} \gets \text{TrainTestSplit}(\mathbf{X}_{\text{norm}}, \mathbf{y}, 0.8)$

\State \textbf{VQC Initialization:}
\State $\text{VQC} \gets \text{VQC}(\mathcal{S}, \mathcal{F}, \mathcal{A}, \mathcal{O}, \mathbf{p_0})$

\State \textbf{Training:}
\State $\text{VQC.fit}(\mathbf{X}_{\text{train}}, \mathbf{y}_{\text{train}})$
\State $q_{\text{train}} \gets \text{VQC.score}(\mathbf{X}_{\text{train}}, \mathbf{y}_{\text{train}})$
\State $q_{\text{test}} \gets \text{VQC.score}(\mathbf{X}_{\text{test}}, \mathbf{y}_{\text{test}})$
\State \textbf{Output:} $q_{\text{train}}$, $q_{\text{test}}$, $\tau$
\end{algorithmic}
\label{alg:device_initialization_training}
\end{algorithm}

\section{Further Results}

\subsection{Clustering Algorithms}
The Algorithm \ref{alg:clustering_algorithms} clusters a list of devices into $k$ clusters using a specified method $\mathcal{M}$. First, it extracts the weights from all devices in $\mathbf{W}$. Based on the chosen clustering method (e.g., k-means, agglomerative, DBSCAN), a clustering model $\mathcal{C}$ is initialized. The model is then fitted to $\mathbf{W}$, and the cluster labels $\mathbf{L}$ are assigned to each device. The devices are grouped into clusters $\mathcal{S}$ according to these labels. Finally, the clustered groups $\mathcal{S}$ are outputted for further processing.
The main clustering algorithms used are KMeans, AgglomerativeClustering, DBSCAN, SpectralClustering, GaussianMixture and MeanShift from Sklearn library.

\begin{algorithm}[!htbh]
\caption{Algorithms used for Clustering Devices}
\label{alg:clustering_algorithms}
\begin{algorithmic}[1]
\Require $\mathcal{D}$: List of devices, $K$: Number of clusters, $\mathcal{M}$: Clustering method, $S$: clusters list containing devices, $\mathcal{U}$: Unique labels, $\mathcal{I}$: Cluster indices, $\mathbf{L}$: result of clustering.
\State $\mathbf{W} \gets \text{extract weights from } \mathcal{D}$
\State \textbf{Initialize clustering model:}
\If{$\mathcal{M} = \text{kmeans}$}
    \State $\mathcal{C} \gets \text{KMeans}(n\_clusters = K, random\_state = 0, n\_init = 10)$
\ElsIf{$\mathcal{M} = \text{agglomerative}$}
    \State $\mathcal{C} \gets \text{AgglomerativeClustering}(n\_clusters = K)$
\ElsIf{$\mathcal{M} = \text{dbscan}$}
    \State $\mathcal{C} \gets \text{DBSCAN}(eps = 0.5, min\_samples = 5)$
\ElsIf{$\mathcal{M} = \text{gmm}$}
    \State $\mathcal{C} \gets \text{GaussianMixture}(n\_components = K, random\_state = 0)$
\ElsIf{$\mathcal{M} = \text{spectral}$}
    \State $\mathcal{C} \gets \text{SpectralClustering} (n\_clusters = K, affinity = \text{nearest\_neighbors}, random\_state = 0)$
\ElsIf{$\mathcal{M} = \text{mean\_shift}$}
    \State $\mathcal{C} \gets \text{MeanShift}()$
\EndIf

\State $\mathbf{L} \gets \mathcal{C}.\text{fit\_predict}(\mathbf{W})$

\State \textbf{Assign Clusters:}
\For{$i = 0$ \textbf{to} $|\mathcal{D}| - 1$}
    \State $\mathcal{D}[i].\text{cluster} \gets \mathbf{L}_i$
\EndFor

\State \textbf{Group Devices by Cluster:}
\State $\mathcal{U} \gets \text{unique labels in } \mathbf{L}$
\State $\mathcal{S} \gets []$ 
\For{$l \in \mathcal{U}$}
    \State $\mathcal{I} \gets \{j \mid \mathbf{L}_j = l\}$
    \State $\mathcal{S} \gets \mathcal{S} \cup \{\mathcal{D}[j] \mid j \in \mathcal{I}\}$
\EndFor

\State \textbf{Output:} $\mathcal{S}$
\end{algorithmic}
\label{alg:cluster_devices}
\end{algorithm}

\subsection{Impact of Clustering Algorithms}
\begin{figure}[!htbh]
    \centering
    \begin{subfigure}[]{0.45\columnwidth}
    \centering
    \includegraphics[width=\columnwidth]{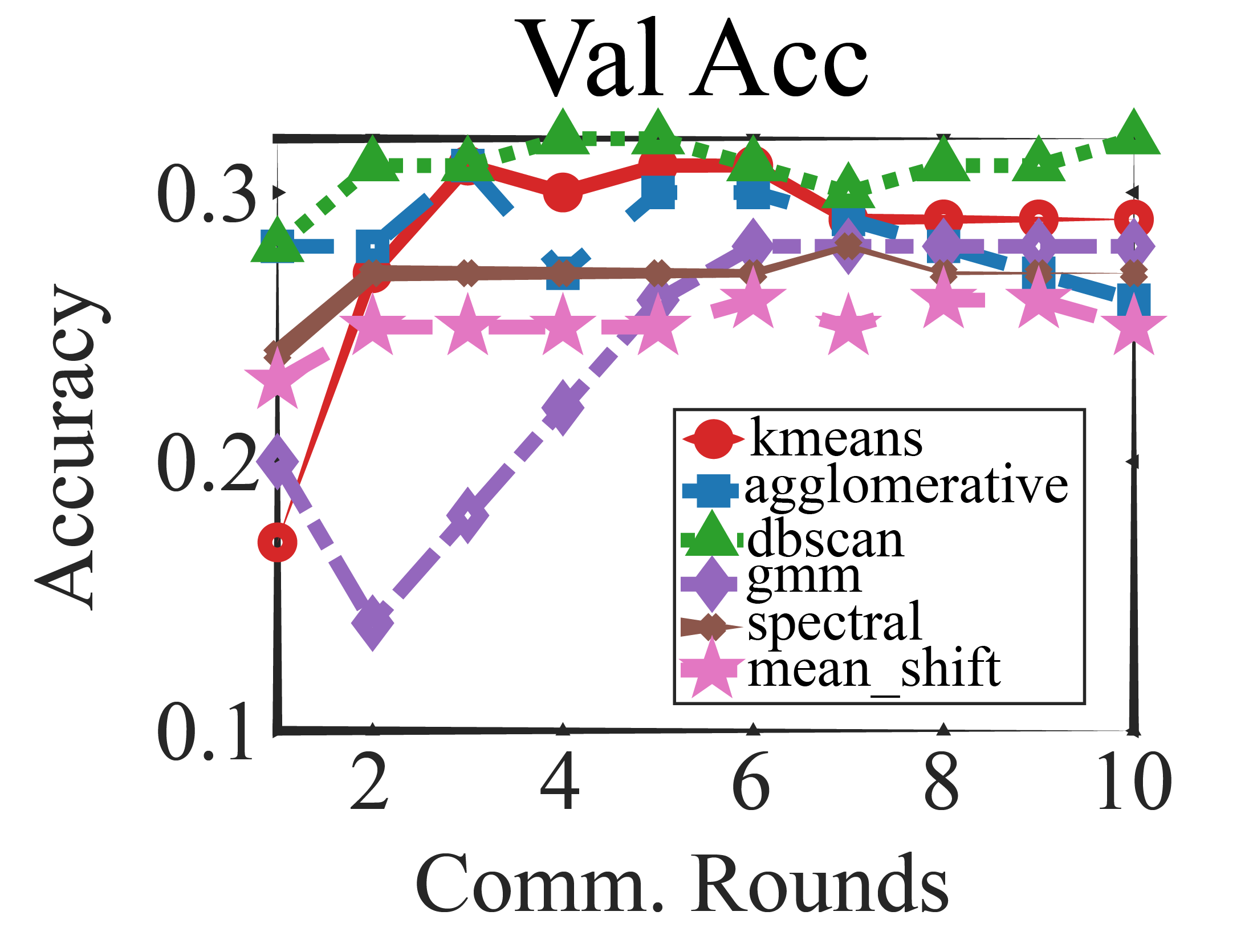}
    \caption{Validation Accuracy}
    \label{fig:server_train_methods}
    \end{subfigure}
    % \hfill
    \begin{subfigure}[]{0.45\columnwidth}
    \centering
    \includegraphics[width=\columnwidth]{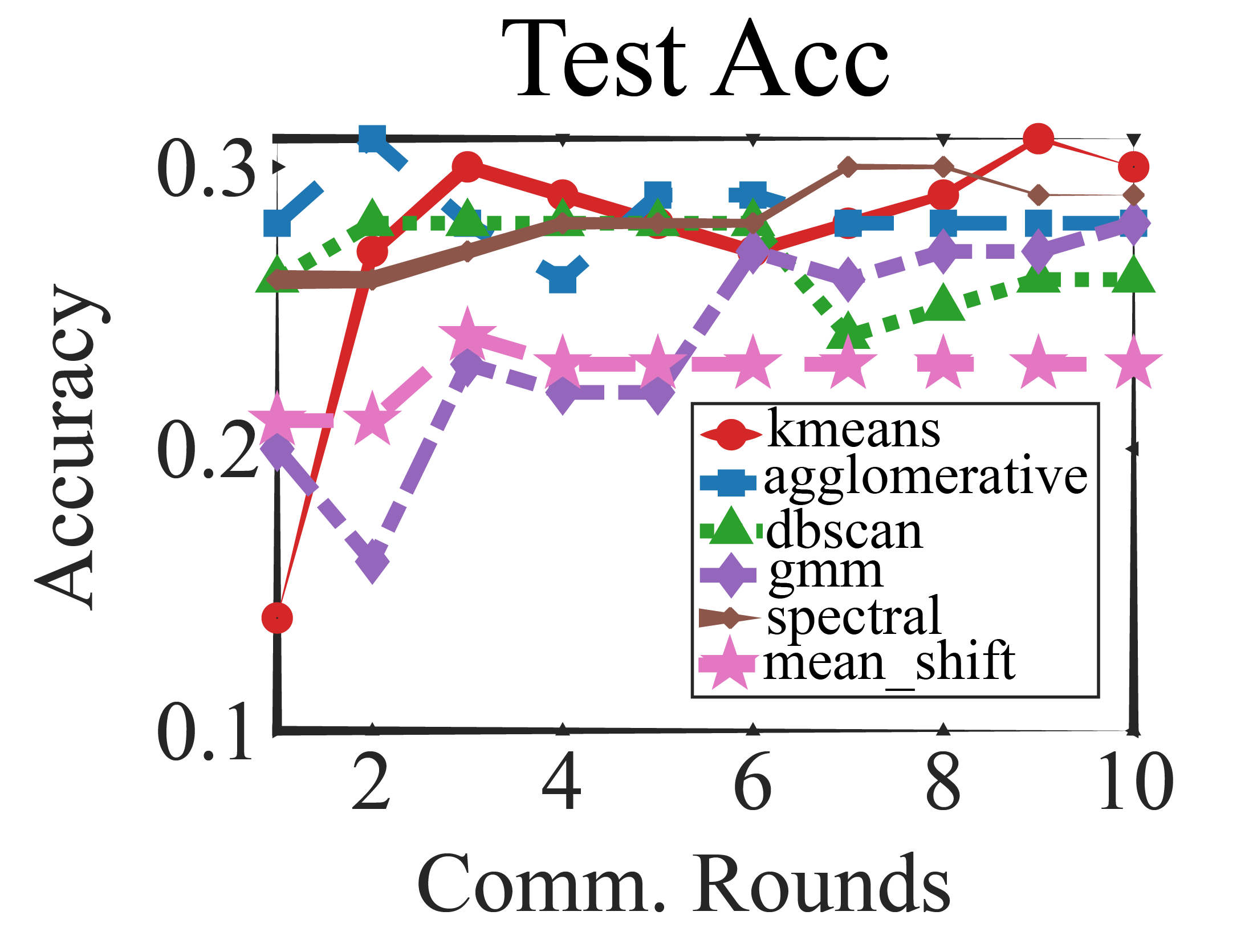}
    \caption{Test Accuracy}
    \label{fig:server_test_methods}
    \end{subfigure}
     \centering
    \caption{Server Performance}
    \label{fig:server_performance_methods}
\end{figure}

\begin{figure}[!htb]
    \centering
    \centering
    \includegraphics[width=0.45\columnwidth]{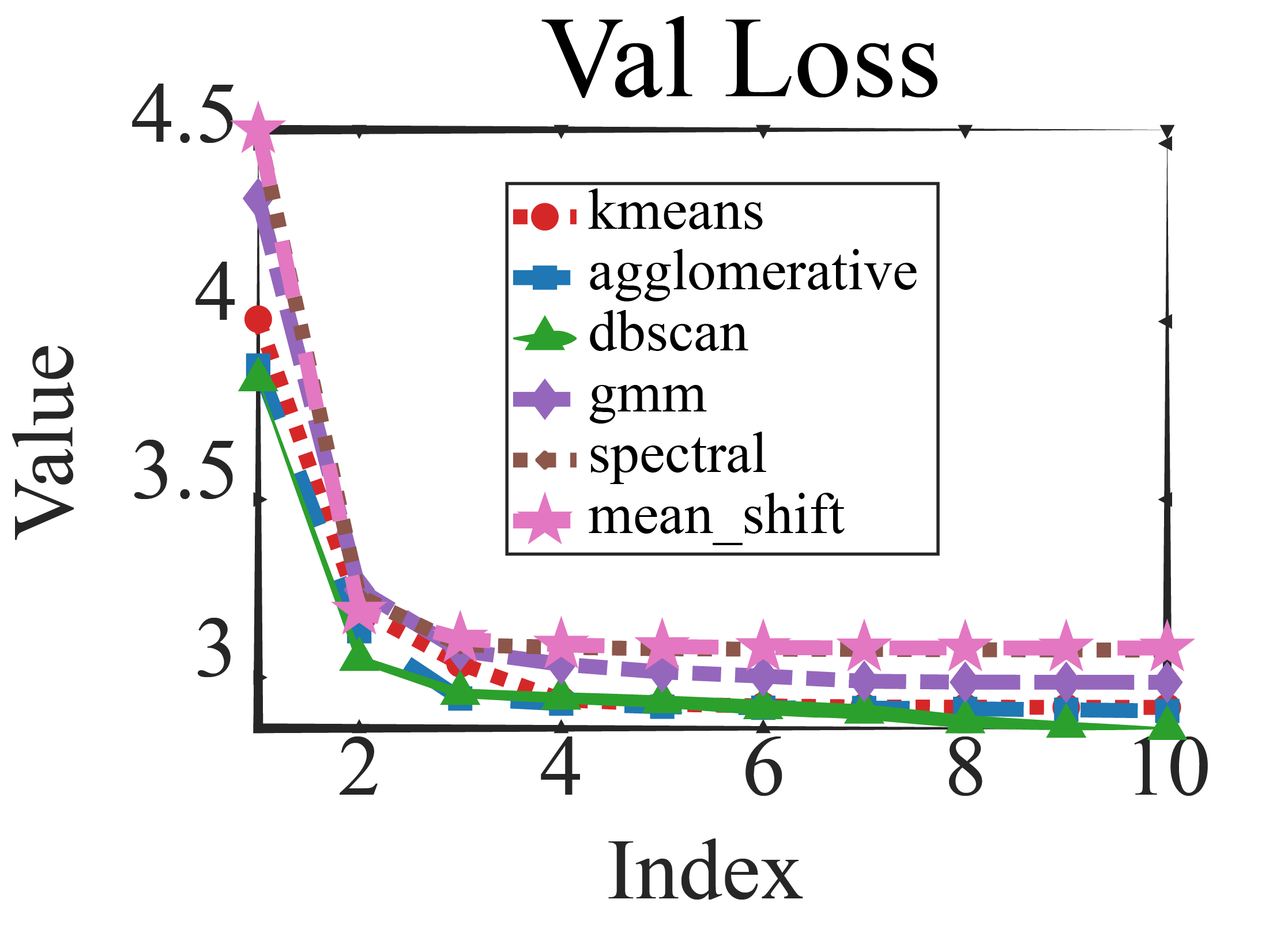}
    \caption{Comparison: 10 devices, 50 maxiter per device optimizer, 10 communication rounds}
    \label{fig:train_performance_nclass6}
\end{figure}

\begin{figure}[!htbh]
    \centering
    \begin{subfigure}[]{0.45\columnwidth}
    \centering
    \includegraphics[width=\columnwidth]{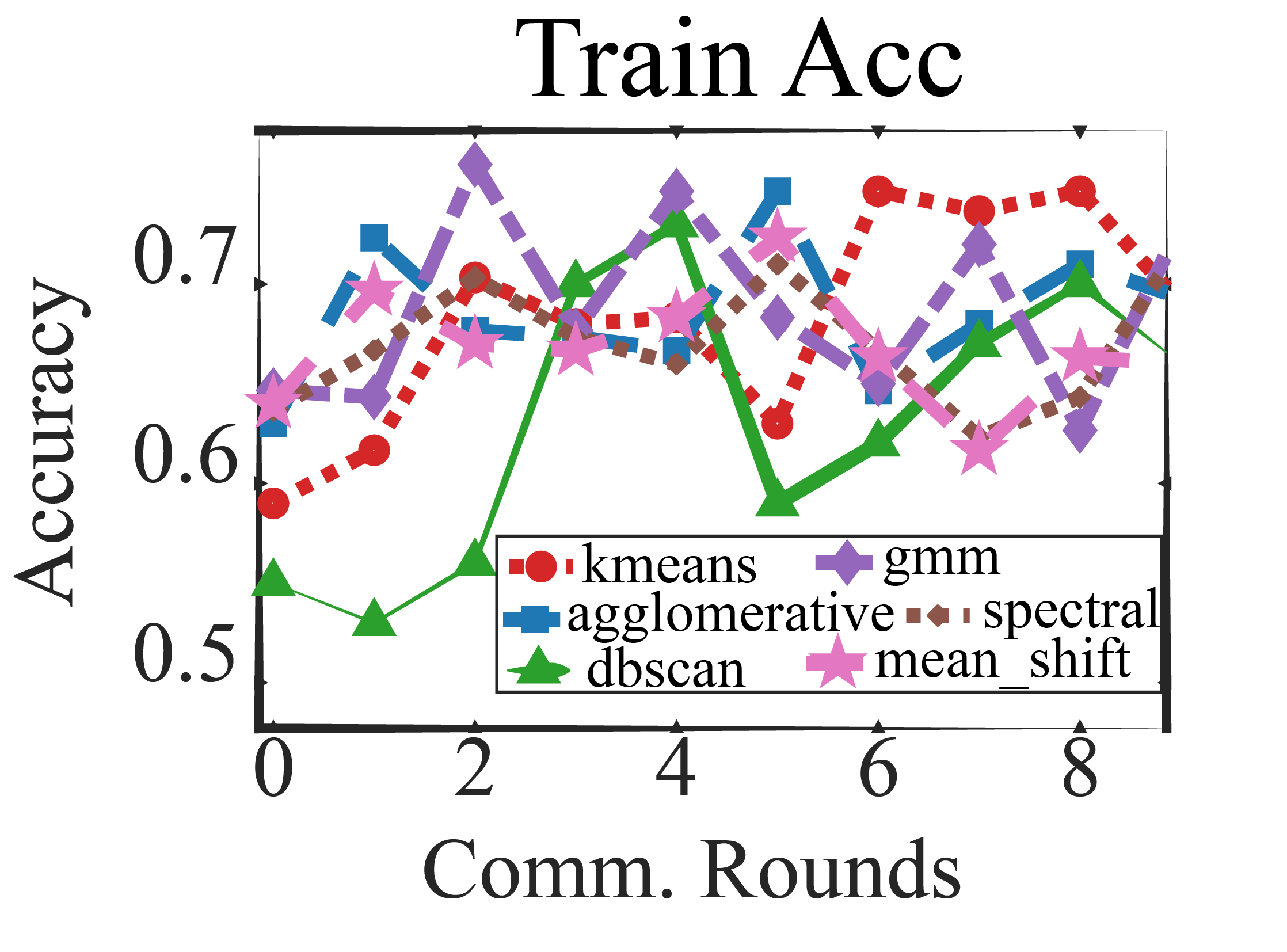}
    \caption{Train Accuracy}
    \label{fig:devices_train_methods}
    \end{subfigure}
    % \hfill
    \begin{subfigure}[]{0.45\columnwidth}
    \centering
    \includegraphics[width=\columnwidth]{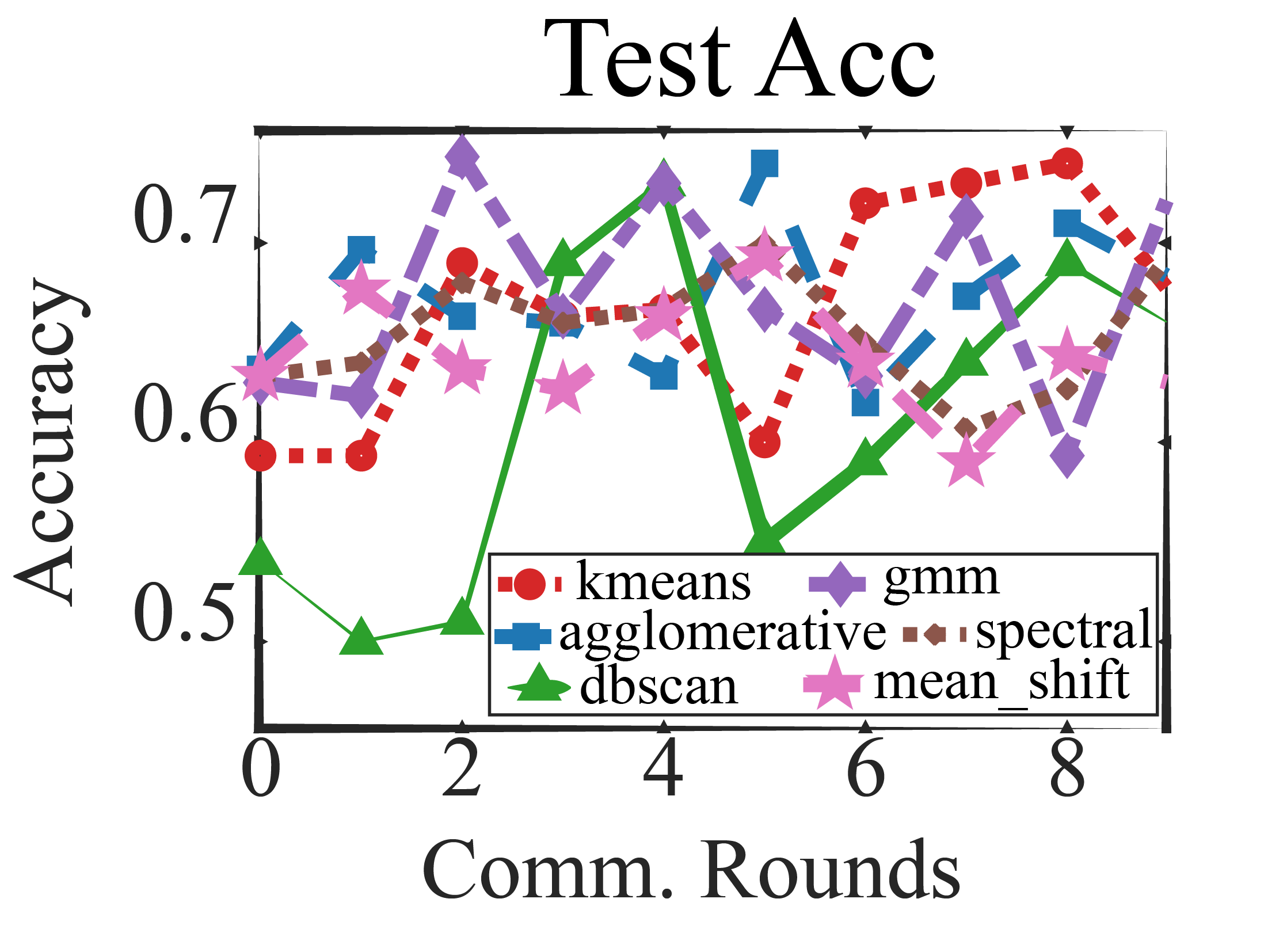}
    \caption{Test Accuracy}
    \label{fig:devices_test_methods}
    \end{subfigure}
     \centering
    \caption{Devices Performance}
    \label{fig:devices_performance_methods}
\end{figure}

We further outline exploratory results in this section.
The impact of various clustering algorithms is clearly visible in the following sets of experiments shown in Figures \ref{fig:server_performance_methods} and \ref{fig:devices_performance_methods}.
First, with all the clustering algorithms, we see the convergence guarantee. 
From the observation for server performance as seen in Figure \ref{fig:server_performance_methods}, the better performing clustering methods are DBSCAN, Kmeans, and agglomerative approaches.
Similar results are reflected in Figure \ref{fig:train_performance_nclass6} with validation loss.


Typical device performance indicates that KMeans generally outperforms other similar clustering algorithms. Nonetheless, the choice of algorithm significantly affects performance, as illustrated in Figure \ref{fig:devices_performance_methods}.

\section{Theoretical Analysis}
\subsection{Proof of Lemma 1}
Let \( F \) be a continuous Lipschitz function with Lipschitz constant \( L \). The regret \( R(T) \) of the COBYLA optimizer after \( T \) iterations is bounded by:
\[
   R_F(T) = \sum_{k=1}^{T} [ F(\theta_t) - F(\theta^*) ] \leq L \sum_{t=1}^{T} \Delta_t
\]
where, \( \theta_t \) is the point at iteration \( t \), \( \theta^* \) is the optimal solution and \( \Delta_t \) is the radius of the trust region at iteration \( t \).

\begin{proof}
With Lipschitz continuity, Since \( F \) is Lipschitz continuous with constant \( L \), for any \( \theta \) and \( \theta' \),
    \[
        |F(\theta) - F(\theta')| \leq L \|\theta - \theta'\|
    \]
COBYLA constructs a trust region around the current point \( \theta_t \) and solves a subproblem within this region:
    \[
        \min_d \quad F(\theta_t + d)
    \]
    subject to:
    \[
        \|d\| \leq \Delta_t
    \]
    Here, \( \Delta_t \) is the radius of the trust region.
In each iteration \( t \), COBYLA updates the point \( \theta_t \) to \( \theta_{t+1} \) by solving the sub-problem, ensuring that \( \|\theta_{t+1} - \theta_t\| \leq \Delta_t \).
The regret \( R(T) \) is the sum of differences between the function values at \( \theta_t \) and \( \theta^* \):
    \[
        R_F(T) = \sum_{t=1}^{T} [ F(\theta_t) - F(\theta^*) ]
    \]
    Using the Lipschitz continuity of \( F \),
    \[
        F(\theta_t) - F(\theta^*) \leq L \|\theta_t - \theta^*\|
    \]
Given that \( \|\theta_t - \theta^*\| \leq \Delta_t \),
    \[
        R_F(T) \leq L \sum_{t=1}^{T} \Delta_t
    \]
\end{proof}

\subsection{Proof of Theorem 1}
Similar to \cite{liConvergenceFedAvgNonIID2019}, we can extend the bound inequality to qFedAvg with a slight adaptation to different factors evident in the quantum setting. Let $\kappa = \frac{L}{\mu}$, $\gamma = \max\{8\kappa, E\}$, and the learning rate $\eta_t = \frac{2}{\mu(\gamma + t)}$. The bound is given by:

\begin{align}
\mathbb{E}[F(\theta_T)] - F^* \leq 
&\frac{2\kappa}{\gamma + T} \left( \frac{B^{\ket{\psi}}}{\mu} + 2L \|\theta_0 - \theta^*\|^2 \right) \notag \\
&+ L \sum_{t=1}^{T} \Delta_t \notag
\end{align}
From \cite{liConvergenceFedAvgNonIID2019}, we have
\[
    \mathbb{E}[F(\theta_T)] - F^* \leq \frac{2\kappa}{\gamma + T} \left( \frac{B^{\ket{\psi}}}{\mu} + 2L \|\theta_0 - \theta^*\|^2 \right)
\]
From Lemma 2, 
\[    R_F(T) \leq L \sum_{t=1}^{T} \Delta_t
\]
Incorporating respective factors, 
we have, 
\begin{align}
\mathbb{E}[F(\theta_T)] - F^* \leq 
&\frac{2\kappa}{\gamma + T} \left( \frac{B^{\ket{\psi}}}{\mu} + 2L \|\theta_0 - \theta^*\|^2 \right) \notag \\
&+ L \sum_{t=1}^{T} \Delta_t
\end{align}

\subsection{Convergence Rate}
To analyze the convergence rate, we begin by bounding the sum that involves the trust region radius \( \Delta_t \).
Assuming that the trust region radius decreases over iterations as
\[
\Delta_t = \frac{\Delta_0}{t^\alpha},
\]
with \( \alpha = 1 \) and initial radius \( \Delta_0 \). Then,
\[
L \sum_{t=1}^{T} \Delta_t = L \Delta_0 \sum_{t=1}^{T} \frac{1}{t} \leq L \Delta_0 (1 + \ln T).
\]
Substituting back, we get
\[
\mathbb{E}[F(\theta_T)] - F^* \leq \frac{2\kappa}{\gamma + T} \left( \frac{B}{\mu} + 2L \| \theta_0 - \theta^* \|^2 \right) + L \Delta_0 (1 + \ln T).
\]
As \( T \) increases, the dominant term is \( O(1/T) \), which confirms the complexity of time.
By setting \( \Delta_t = \frac{\Delta_0}{t} \), the cumulative error remains controlled
\[
L \sum_{t=1}^{T} \Delta_t \leq L \Delta_0 (1 + \ln T).
\]
The expected error after \( T \) iterations:
\[
\mathbb{E}[F(\theta_T)] - F^* \leq \underbrace{\frac{C'}{\gamma + T}}_{O(1/T)} + \underbrace{L \Delta_0 (1 + \ln T)}_{O(\ln T)},
\]
where \( C' = 2\kappa \left( \dfrac{B}{\mu} + 2L \| \theta_0 - \theta^* \|^2 \right) \).
Since \( \ln T \) grows slower than \(1 / T \), the convergence rate is dominated by \( O(1/T) \).

\end{document}